\documentclass[twoside,11pt]{article}
\usepackage{amsmath,amsthm,amssymb,bm,mathrsfs,}
\usepackage{graphicx}
\usepackage[ruled,linesnumbered]{algorithm2e}
\usepackage{caption,enumerate,natbib,listings,setspace}
\newcommand{\f}{f_{\mathcal{F}}^{\star}}
\newcommand{\tf}{\widetilde{f}_{\mathcal{F}}^{\star}}

\newcommand{\tbx}{\widetilde{\bm{X}}}
\newcommand{\ty}{\widetilde{Y}}

\usepackage[title]{appendix}
\usepackage{bbm}
\usepackage{booktabs}
\usepackage{subcaption}
\usepackage{comment}
\usepackage{enumitem}
\usepackage{multirow}
\usepackage{blindtext}
\newcommand{\argmin}{\mathop{\mathrm{argmin}}}

\newcommand{\sign}{\mathop{\mathrm{sign}}}
\newtheorem{Assumption}{{\bf Assumption}}
\newtheorem{theorem}{Theorem}
\newtheorem{definition}{Definition}
\newtheorem{lemma}{Lemma}
\newtheorem{corollary}{Corollary}

\usepackage{tikz}

\usepackage{jmlr2e}

\ShortHeadings{Utility Theory of Synthetic Data Generation}{Xu, Sun, and Cheng}
\firstpageno{1}
\usepackage{cellspace} % Add this package
\newcolumntype{C}[1]{>{\centering\arraybackslash}m{#1}}
\begin{document}

\title{\Large \bf Utility Theory of Synthetic Data Generation}

\author{\name Shirong Xu \email shirong@stat.ucla.edu. \\
       \addr Department of Statistics, University of California, Los Angeles\\
       \AND
       \name Will Wei Sun \email sun244@purdue.edu \\
       \addr Mitchell E. Daniels, Jr. School of Business, Purdue University.
       \AND Guang Cheng \email guangcheng@ucla.edu\\
        \addr Department of Statistics, University of California, Los Angeles.
       }

\editor{My editor}

%\editor{}
\date{}
\maketitle
\begin{abstract}
Synthetic data algorithms are widely employed in industries to generate artificial data for downstream learning tasks. While existing research primarily focuses on empirically evaluating utility of synthetic data, its theoretical understanding is largely lacking. This paper bridges the practice-theory gap by establishing relevant utility theory in a statistical learning framework. It considers two utility metrics: generalization and ranking of models trained on synthetic data. The former is defined as the generalization difference between models trained on synthetic and on real data. By deriving analytical bounds for this utility metric, we demonstrate that the synthetic feature distribution does not need to be similar as that of real data for ensuring comparable generalization of synthetic models, provided proper model specifications in downstream learning tasks. The latter utility metric studies the relative performance of models trained on synthetic data. In particular, we discover that the distribution of synthetic data is not necessarily similar as the real one to ensure consistent model comparison. Interestingly, consistent model comparison is still achievable even when synthetic responses are not well generated, as long as downstream models are separable by a generalization gap. Finally, extensive experiments on non-parametric models and deep neural networks have been conducted to validate these theoretical findings.
\end{abstract}

\section{Introduction}
In recent years, there has been a growing interest in synthetic data generation due to its versatility in a wide range of applications, including financial data \citep{dogariu2022generation,altman2024realistic} and medical data \citep{chen2021synthetic,qian2024synthcity,rafiei2024improving}. The core idea of data synthesis is to generate a synthetic surrogate of the real dataset that can either replace real data to protect privacy or supplement real data for training. For example, synthetic data has enabled financial institutions to share their data while ensuring compliance with data sharing restrictions \citep{assefa2020generating}. It has also been used to augment financial datasets for fraud detection purposes \citep{charitou2021synthetic}. Similarly, in the medical field, synthetic data have been used to improve data privacy and the performance of predictive models for disease diagnosis \citep{chen2021synthetic}. In the literature, significant progress has been made in developing synthetic data algorithms, including classical marginal-based methods \citep{bi2022distribution,li2023statistical,qian2024synthcity}, generative adversarial networks \citep{arjovsky2017wasserstein,goodfellow2020generative,zhou2022deep}, and diffusion models \citep{song2021maximum,ouyang2023improving,nguyen2024dataset}.

An important aspect of evaluating the quality of synthetic data is assessing the generalization ability of downstream tasks trained from synthetic data, i.e., learning utility. Specifically, \citet{jordon2022synthetic} posed two research questions concerning the utility of synthetic data when used for downstream tasks:
\begin{itemize}[leftmargin=1.5cm, itemindent=1cm]
\item[\textbf{Question 1}]: When do models trained on synthetic data perform comparably to those trained on real data?
\item[\textbf{Question 2}]: When is the model comparison of models trained on synthetic data consistent with that of models trained on real data?
\end{itemize}

\begin{figure}[ht]
\centering
            \begin{subfigure}[b]{0.49\textwidth}
\includegraphics[scale=0.46]{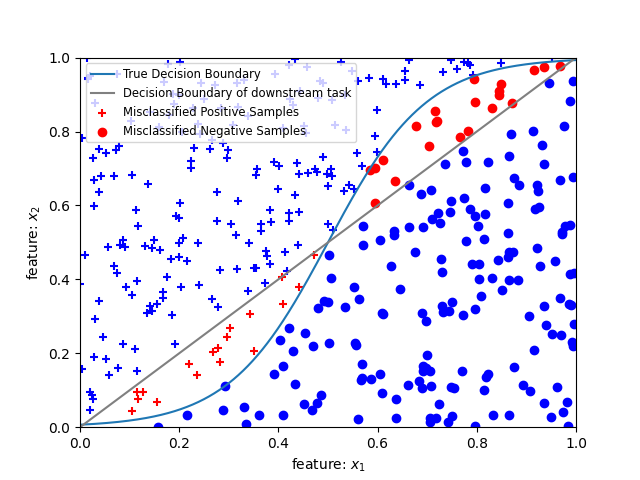}
            \end{subfigure}
            \begin{subfigure}[b]{0.49\textwidth}
\includegraphics[scale=0.46]{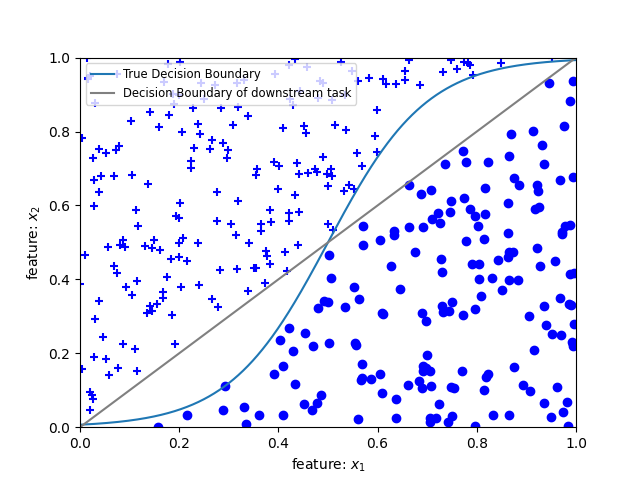}
    \end{subfigure}
    \caption{In this example, the true decision boundary is nonlinear, whereas the decision boundary for the downstream task is linear. Notably, if the misclassified samples are removed, the downstream linear decision boundary in the right plot is identical to that in the left plot, as it achieves zero prediction error.
}
    \label{fig:UF}
    \end{figure}

In this context, the learning utility of synthetic data lies in two key aspects. First, models trained on synthetic data exhibit similar generalization performance as those trained on real data, ensuring \textit{consistent generalization}. Second, the performance ranking of models trained on synthetic data align with those trained on real data, providing \textit{consistent model comparison}. Intuitively, both consistent generalization and consistent model comparison are less stringent requirements for synthetic data compared to requiring synthetic data to approximate real data in distribution. In Figure \ref{fig:UF}, we present a classification example as an illustration. This example shows that classifiers trained on two datasets with different underlying feature distributions can still be the same, ensuring consistent generalization. This motivates us to theoretically understand how a synthetic data algorithm affects the generalization of models trained from synthetic data and what properties a synthetic data algorithm should possess to ensure two types of learning utility.

To address \textbf{Question 1}, we consider a general utility metric \citep{hittmeir2019utility,el2021evaluating}, which is defined as the absolute difference between the generalization performances on real test data between the models learned from the real and synthetic datasets. The rationale behind this utility metric is that a good synthetic dataset should produce a model with comparable performance to the real dataset. In theory, we establish explicit bounds for the utility metric in the classification setting (Theorem \ref{Thm:UTBmore}). Specifically, the analytic bound is decomposed into three major components, including the quality of the synthetic features (fidelity to the feature), the estimation of the regression function (estimation of the relationship), and the specification of the model in the downstream learning task. This decomposition enables us to identify key aspects that synthetic data algorithms should prioritize when generating high-quality synthetic datasets for downstream learning tasks. 

To address \textbf{Question 2}, we consider the utility of maintaining consistent relative performances for downstream tasks trained on synthetic data \citep{jordon2019pate,jordon2022synthetic}. Ideally, the relative performance of models trained on synthetic data should align with that of models trained on real data. This aspect of utility is of critical importance when data holders are unable to release their real datasets for public competitions (e.g., Kaggle) due to privacy concerns. To establish sufficient conditions for this ``ranking" consistency, we propose a new metric called $(V,d)$-fidelity level, which measures the distributional difference between real and synthetic features. Specifically, we show that the results of the model comparison based on synthetic data (Theorem \ref{Thm:Consis}) depend on the generalization gap between two model classes and the distribution discrepancy between real and synthetic data. We demonstrate that even in cases where synthetic data is not well generated, consistent model comparison is still achievable. To our knowledge, our work provides the first analytic solution to the question posed in \citet{jordon2022synthetic} about when the relative performance of synthetic models is consistent with that of models trained on real data.

There are several theoretical findings regarding the utility metrics. Through our analysis, we conclude the following hierarchy regarding the requirements on synthetic data:
\begin{align*}
    \text{Distribution Approximation}
    \succ
    \text{Consistent Generalization}
    \succ
    \text{Consistent Model Comparison}.
\end{align*}
Here, $\succ$ represents a stronger requirement. Specifically, we first show that as long as synthetic features have perfect fidelity and the relationship between features and responses can be well approximated by synthetic data algorithms, consistent generalization is guaranteed. The first result is expected, since it is widely perceived that statistical results from synthetic data will be valid if the distribution of synthetic data is not far from that of real data. The learning utility of synthetic data is achieved when synthetic data achieves \textit{distribution approximation} to real data. Second, we prove that for achieving consistent generalization, the distribution approximation is actually not necessary. In particular, our analytic bound (Theorem \ref{Thm:UTBmore}) reveals an interesting phenomenon that if the model specification in the downstream learning task is correct in the sense that it can capture the relationship between features and responses, good fidelity of the feature is not needed for the convergence of utility metrics. This result shows that when the model specification is correct, the synthetic distribution is not required to be the same as the real distribution to produce a model with comparable generalization performance. This discovery contributes to our understanding of a surprising but frequently observed phenomenon: training models using synthetic data can improve generalization \citep{tremblay2018training,azizi2023synthetic}. For example, \citet{azizi2023synthetic} shows that the use of synthetic data generated from diffusion models can significantly improve the performance of ImageNet classification. Our result suggests that the reason behind this phenomenon is due to the high approximation powers of the function classes of downstream tasks. Third, we show that consistent model comparison imposes less stringent requirements on synthetic data. Specifically, consistent model comparison is still achievable even if consistent generalization fails, as long as the generalization gap between downstream tasks is large enough to neutralize the distributional difference between real and synthetic data.

\subsection{Related Works}

We review the literature on comparing the statistical results derived from synthetic data with those from real data. \citet{karr2006framework} calculated the averaged overlap between the confidence intervals of coefficients estimated from synthetic and real data, while their standardized difference are adopted by \citet{woo2015generalised} and \citet{nowok2016synthpop} as a utility measure in linear regression. These utility metrics focus on the degree to which the statistical results of synthetic data align with those of real data. Similar task-specific utility metrics have also been proposed for supervised learning, aiming to compare the generalization performance of models trained on synthetic and real data \citep{beaulieu2019privacy,hittmeir2019utility,rankin2020reliability,el2021evaluating}. Although numerous studies have introduced and evaluated utility metrics for synthetic data, the theoretical relationship between these metrics and the synthetic data distributions remains underexplored.

\subsection{Paper Organization}
The remainder of the paper is organized as follows. In Section \ref{Sec:Pre}, we introduce the necessary notations and present background information and basic concepts related to binary classification. In Section \ref{Sec:DataSyn}, we introduce a framework for analyzing the utility of synthetic data in the context of classification and present several examples where consistent generalization is guaranteed under imperfect synthetic data distribution. Section \ref{Sec:UB} establishes an analytic bound for the utility metric for classification and provides explicit sufficient conditions for ensuring consistent generalization. Additionally, we study the asymptotic behavior of the worst-case utility in a specific example where a two-step data generation process is employed. In Section \ref{Sec:MC}, we examine sufficient conditions on synthetic data to achieve consistent model comparison. In Section \ref{Sec:Experiments}, we conduct extensive simulations and real-world applications to support our theoretical results. In Appendix, we present detailed proofs for all lemmas, theorems, and corollaries.

\section{Preliminaries}
\label{Sec:Pre}
For a vector $\bm{x} \in \mathbb{R}^p$, we denote its $l_1$-norm and $l_2$-norm as $\Vert \bm{x} \Vert_1=\sum_{i=1}^p |x_i|$ and $\Vert \bm{x} \Vert_2=\big(\sum_{i=1}^p |x_i|^2\big)^{1/2}$, respectively. We let $\phi(\bm{x})=1/(1+\exp(-\sum_{i=1}^p x_i))$. We let $\text{Bern}(p)$ denote the Bernoulli random variable parametrized by $p$. For a function $f:\mathcal{X}\rightarrow \mathbb{R}$, we denote its $L_p$-norm with respect to the probability measure $\mu$ as $\Vert f\Vert_{L^p(\mu)}=\big(\int_{\mathcal{X}} |f(\bm{x})|^p d\mu(\bm{x})\big)^{1/p}$ and $\Vert f\Vert_{L^p}=\big(\int_{\mathcal{X}} |f(\bm{x})|^p d\bm{x}\big)^{1/p}$. For two given sequences $\{A_n\}_{n \in \mathbb{N}}$ and $\{ B_n\}_{n\in \mathbb{N}}$, we write $A_n \gtrsim B_n$ if there exists a constant $C>0$ such that $A_n \geq C B_n$ for any $n \in \mathbb{N}$. Additionally, we write $A_n \asymp B_n$ if $A_n \gtrsim B_n$ and $A_n \lesssim B_n$. For a continuous random variable $X$, we let $P_X(x)$ denote its probability density function at $x$ and $\mathbb{P}_{X}$ denote the associated probability measure. Let $\mathbb{E}_{X}$ denote the expectation taken with respect to the randomness of $X$. For a sequence of random variables $\{X_n\}_{n\in \mathbb{N}}$, we let $X_n =o_p(1)$ represent that $X_n$ converges to zero in probability. We denote a fully-connected neural network with Rectified Linear Unit (ReLU) activation function as
\begin{align*}
    L_{\bm{\omega}}(\bm{x}) = q_{L+1} \circ \sigma \circ q_{L} \circ \cdots \circ \sigma \circ q_2 \circ \sigma \circ q_1(\bm{x}),
\end{align*}
where $\sigma(x) = \max\{x, 0\}$ denotes the ReLU activation function, $\circ$ represents function composition, $q_l(\bm{x}) = \bm{A}_l \bm{x} + \bm{b}_l$ with $\bm{A}_l \in \mathbb{R}^{N_{l+1} \times N_l}$ and $\bm{b}_l \in \mathbb{R}^{N_{l+1}}$, and $\bm{\omega} = \{(\bm{A}_l, \bm{b}_l) \mid l = 1, \ldots, L+1\}$ denotes all parameters. Let $W = \max\{N_1, \ldots, N_{L+1}\}$ represent the maximum width and $L$ denote the depth of the neural network. We further let $\mathcal{N}(W, L)$ represent all $L$-layer fully-connected neural networks with maximum width $W$.

 Let $\mathcal{D}=\{ (\bm{x}_i,y_i) \}_{i=1}^n$ denote the dataset in binary classification, where $y_i$  takes values in $\{-1,1\}$ and
\begin{align*}
y_i = \begin{cases}
1 & \mbox{ with probability } \eta(\bm{x}_i), \\
-1&\mbox{ with probability } 1-\eta(\bm{x}_i), \\
\end{cases}
\end{align*}
where $\eta(\bm{x}_i) = \mathbb{P}(Y=1|\bm{X}=\bm{x}_i)$ and $Y$ denotes the response variable in classification. Denote the classification risk for a classifier $f:\mathcal{X} \rightarrow \{-1,1\}$ under the $0$-1 loss by $R(f) = \mathbb{E}\Big[ I\big(f(\bm{X}) \neq Y\big)\Big]$, where $I(\cdot)$ is an indicator function. The optimal classifier minimizing $R(f)$ is the Bayes classifier, which is given as $f^\star(\bm{X}) = \sign\big(\eta(\bm{X}) - 1/2\big)$. The excess risk in classification is defined as
\begin{align*}
\Phi(f) =R(f) -  R(f^\star) 
=  \mathbb{E}\Big[
I\big(f(\bm{X}) \neq f^\star(\bm{X}) \big)\big|
2\eta(\bm{X})-1
\big|
\Big],
\end{align*} 
where the expectation is taken with respect to $\bm{X}$. For a specific class of classifier $F$, we let $f_{\mathcal{F}}^\star = \argmin_{f\in \mathcal{F}}R(f)$ denote the optimal classifier in $\mathcal{F}$ to minimize $R(f)$.

For any generative model used for synthetic data generation, we let $\mathbb{P}_{\widetilde{\bm{X}},\widetilde{Y}}$ denote the underlying distribution. Correspondingly, we define the classification risk of a classifier $f$ under the synthetic distribution as $\widetilde{R}(f) =\mathbb{E}\left[
I\big(f(\widetilde{\bm{X}})\neq  \widetilde{Y}\big) 
\right]$. Additionally, we denote the conditional distribution of responses as $\widetilde{\eta}(\bm{x}) = \mathbb{P}(\widetilde{Y}=1|\widetilde{\bm{X}}=\bm{x})$. Let $\widetilde{f}_{\mathcal{F}}^\star = \argmin_{f \in \mathcal{F}} \widetilde{R}(f)$ denote the optimal function in $\mathcal{F}$ under the synthetic distribution $\mathbb{P}_{\widetilde{\bm{X}},\widetilde{Y}}$ and $\widetilde{f}^\star(\bm{x})=\sign(\widetilde{\eta}(\bm{x})-1/2)$ denote the Bayes classifier under the synthetic distribution. In this paper, we assume $\widetilde{\eta}(\bm{x}) \neq 1/2$ and $\eta(\bm{x}) \neq 1/2$ almost surely.

\section{Data Synthesis in Supervised Learning}
\label{Sec:DataSyn}

\begin{figure*}
 \centering
 \includegraphics[scale=0.32]{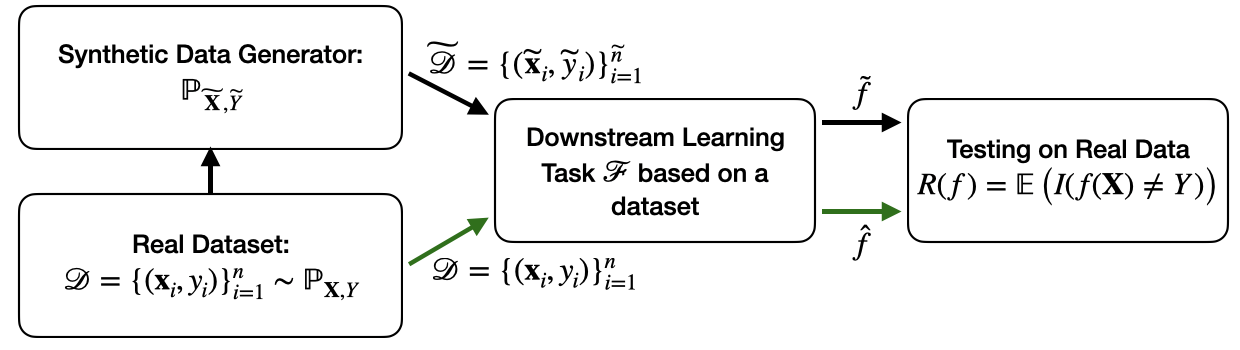}
 \caption{The architecture for generating and evaluating synthetic data in supervised learning. Red arrows indicate possible privacy breaches.}
 \label{Archi}
 \end{figure*}

In this paper, we consider a common framework for generating synthetic data for downstream learning tasks \citep{shoshan2023synthetic,van2023synthetic}. Specifically, a real dataset $\mathcal{D}=\{ (\bm{x}_i, y_i) \}_{i=1}^n$ is used to train a generative model, and a resulting synthetic dataset $\widetilde{\mathcal{D}} =\{(\widetilde{\bm{x}}_i, \widetilde{y}_i) \}_{i=1}^{\widetilde{n}}$ is released for downstream learning tasks, where $\widetilde{n}$ is the size of the synthetic dataset. As illustrated in Figure \ref{Archi}, a synthetic dataset $\widetilde{\mathcal{D}}$ is typically generated to replace $\mathcal{D}$ for a downstream training task $\mathcal{F}$. The objective of the downstream task is to obtain a classifier based on the released synthetic dataset $\widetilde{\mathcal{D}}$. Various frameworks exist for obtaining a classifier, such as plug-in methods \citep{audibert2007fast} and the empirical risk minimization framework \citep{bartlett2006convexity}.

Here we let $\widehat{f}$ and $\widetilde{f}$ be the resulting classifiers obtained from $\mathcal{D}$ and $\widetilde{\mathcal{D}}$, respectively. The central problem of data synthesis in supervised learning is whether $\widetilde{f}$ performs similarly to $\widehat{f}$ in terms of their generalization capabilities \citep{hittmeir2019utility,rankin2020reliability}. Hence, we consider the following utility metric:
\begin{align}
\label{UtiR}
U(\widetilde{f},\widehat{f}) = 
\Big|
R(\widetilde{f}) - R(\widehat{f})\Big|,
\end{align}
where $R(\widetilde{f})$ and $R(\widehat{f})$ represent the generalization performance of $\widetilde{f}$ and $\widehat{f}$ on unseen {\em real data}, respectively. In the literature, $R(\widetilde{f})$ is referred to as the ``train on synthetic data, test on real data" (TSTR) setting \citep{jordon2018measuring,jordon2022synthetic,van2023synthetic}.

The rationale behind $(\ref{UtiR})$ is that the classifier trained on a synthetic dataset should perform similarly in predicting unobserved samples (from the real distribution) as it would when trained on the real dataset. Therefore, an ideal synthetic data generative model should ensure that $U(\widetilde{f}, \widehat{f})$ converges to zero in probability. It is worth noting that $U(\widetilde{f}, \widehat{f})$ consists of two sources of randomness: one from the real dataset and another from the synthetic dataset. If $\mathcal{D}$ and $\widetilde{\mathcal{D}}$ are independent and generated from the same distribution, the convergence of $U(\widetilde{f}, \widehat{f})$ towards zero is straightforward and guaranteed by the convergence of the upper-bounding term, defined as follows:
\begin{align*}
U(\widetilde{f},\widehat{f})  \leq 
\Big|
R(\widetilde{f}) -R(f^\star_{\mathcal{F}})\Big|+
\Big| R(\widehat{f})-R(f^\star_{\mathcal{F}})
\Big|,
\end{align*}
where $f^\star_{\mathcal{F}} = \argmin_{f \in \mathcal{F}}R(f)$ denotes the optimal classifier in $\mathcal{F}$ to approximate the Bayes decision rule. The right-hand side can be seen as twice the estimation errors in statistical learning theory, of which the convergence in probability is supported by a vast body of literature \citep{vapnik1999overview,bartlett2006convexity}. To understand the scenarios in which we can ensure the convergence in probability of \( U(\widetilde{f}, \widehat{f}) \) towards zero, it is essential to understand the value to which it converges. Theoretically, as $n$ and $\widetilde{n}$ converge to infinity, $\widetilde{f}$ and $\widehat{f}$ intend to have the same generalization performance as $\tf$ and $\f$, respectively.

\subsection{Convergence of Utility Metric}

Let $U(\tf,\f)=R(\tf)-R(\f)$ denote the population version of $U(\widetilde{f},\widehat{f})$. It is evident that $U(\widetilde{f}_{\mathcal{F}}^{\star},f_{\mathcal{F}}^{\star})$ remains non-negative, a consequence of the optimality of $f_{\mathcal{F}}^{\star}$. A smaller value of $U(\widetilde{f}_{\mathcal{F}}^{\star},f_{\mathcal{F}}^{\star})$ implies a higher utility of $\mathbb{P}_{\tbx,\ty}$ for downstream classification tasks, indicating that the optimal classifier $\tf$ under the synthetic distribution is close to $\f$, which is obtained under the real distribution in terms of generalization.

\begin{theorem}
\label{UboundC}
 Let $\widehat{f}$ and $\widetilde{f}$ be classifiers trained from $\mathcal{D}$ and $\widetilde{\mathcal{D}}$, respectively. It holds that
\begin{align}
\label{UUUC}
    U(\tf,\f)-\Delta_1 - \Delta_2 \leq 
    U(\widehat{f},\widetilde{f}) \leq 
    U(\tf,\f)+\Delta_1 + \Delta_2,
\end{align}
where $\Delta_1 = \mathbb{E}_{\bm{X}}\left[I(\widehat{f}(\bm{X}) \neq \f(\bm{X}))\right]$ and $\Delta_2 = \mathbb{E}_{\bm{X}}\left[I(\widetilde{f}(\bm{X}) \neq \tf(\bm{X}))\right]$.
 \end{theorem}

In Theorem \ref{UboundC}, we establish a range for the value of \( U(\widehat{f},\widetilde{f}) \). In this result, $\Delta_1$ and $\Delta_2$ represent the probabilities of $\widehat{f}$ and $\widetilde{f}$ failing to have the same predictions as $\f$ and $\tf$ on future real samples, respectively. From a statistical viewpoint, as \( n \) and \( \widetilde{n} \) increase, both \( \Delta_1 \) and \( \Delta_2 \) will converge to zero in probability. Consequently, this makes \((\ref{UUUC})\) a tighter bound for \( U(\widehat{f}, \widetilde{f}) \). 

\begin{figure}[ht]
\centering
\begin{subfigure}[b]{0.475\textwidth}
        \includegraphics[scale=0.475]{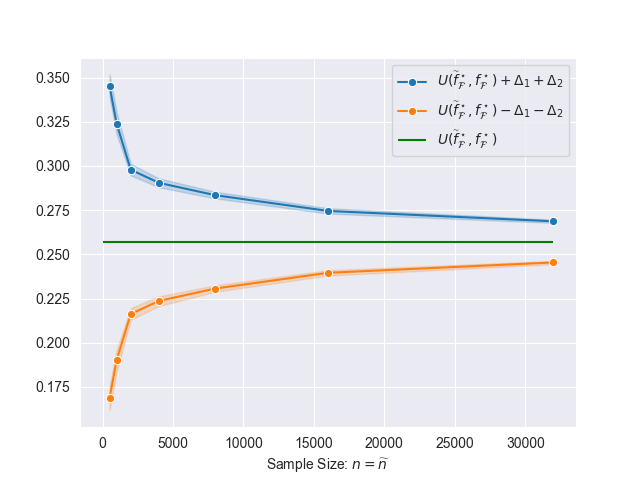}
        \caption{Copula GAN}
    \end{subfigure}
    \begin{subfigure}[b]{0.475\textwidth}
\includegraphics[scale=0.475]{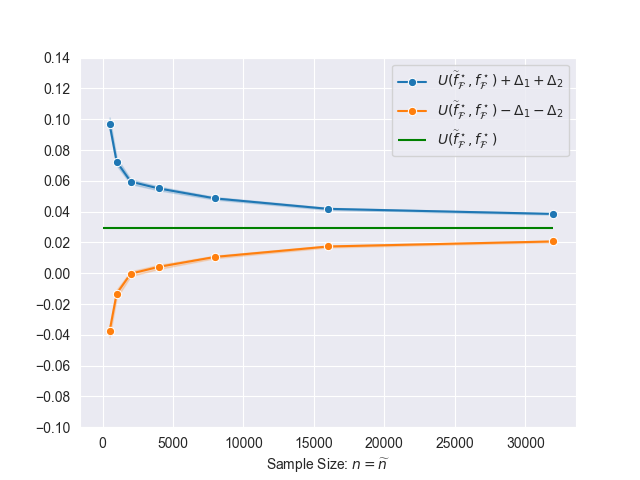}
        \caption{Gaussian Copula}
    \end{subfigure}
    \begin{subfigure}[b]{0.475\textwidth}
\includegraphics[scale=0.475]{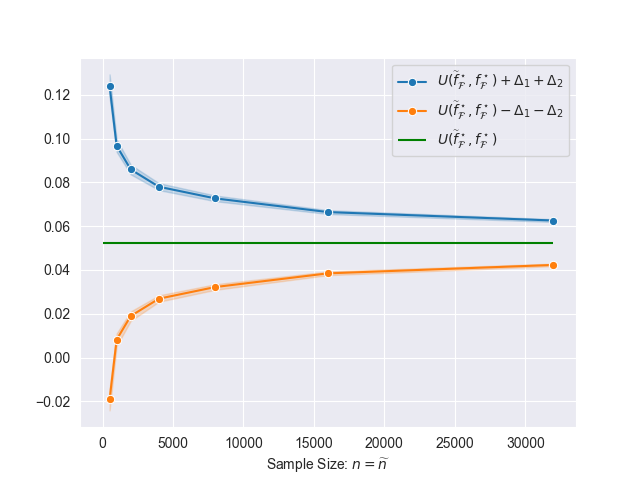}
        \caption{VAE}
    \end{subfigure}
    \begin{subfigure}[b]{0.475\textwidth}
\includegraphics[scale=0.475]{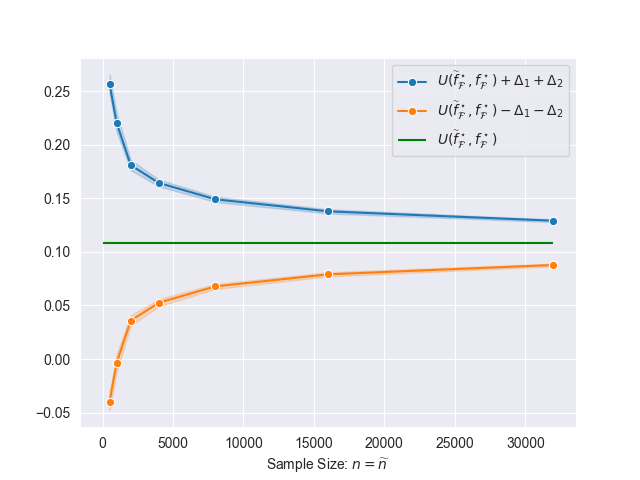}
        \caption{CTGAN}
    \end{subfigure}
\caption{The behavior of \( U(\widehat{f}, \widetilde{f}) \) under varying training sample sizes and different generative models.}
\label{fig:Bound_Con}
\end{figure}

To demonstrate the convergence of $\Delta_1$ and $\Delta_2$, we examine a simplified scenario in which the generative model is trained on a fixed-size dataset independent of $\mathcal{D}$. For real samples, we generate 4-dimensional real features \(\bm{x} \in \mathbb{R}^4\) from a mixture of Gaussians and generate the associated response via \(\text{Bern}(\phi(\bm{x}))\) with $\phi(\bm{x})=1/(1+\exp(-\sum_{i=1}^p x_i))$. In this scenario, we consider four generative models: Copula GAN \citep{patki2016synthetic}, Gaussian Copula \citep{Gaussiancopula}, CTGAN \citep{xu2019modeling}, and Variational Autoencoder (TVAE; \citealt{kingma2013auto,alain2014regularized}). For the downstream task, we use a linear SVM with sample sizes $n = \widetilde{n} \in \{500 \times 2^i \mid i = 0, \ldots, 6\}$ to estimate $\widehat{f}$ and $\widetilde{f}$. . To estimate \(\f\) and \(\tf\), we use $10^5$ real and synthetic samples for training. For each case, we replicate the experiment 100 times and report the averaged values along with their 95\% confidence intervals in Figure \ref{fig:Bound_Con}.

There are two conclusions from the results shown in Figure \ref{fig:Bound_Con}. First, as $n$ and $\widetilde{n}$ increase, both $\Delta_1$ and $\Delta_2$ approach zero for any synthetic data generative model. This result aligns with statistical learning theory \citep{bartlett2006convexity} that the empirical classifier will converge to its population counterpart as more data is used for training. Second, $U(\tf,\f)$ depends solely on the generative model and the dataset used for the generative model. As shown in Figure \ref{fig:Bound_Con}, the four generative models exhibit varying levels of utility for the downstream linear SVM task, given the same dataset. Among the evaluated methods, the Gaussian Copula modeling demonstrates the highest utility preservation, as indicated by the smallest value of $U(\tf,\f)$ in this simulation setting. This may be because Gaussian copula modeling captures the correlation relationship between variables effectively in this example. Theoretically, $U(\widetilde{f}, \widehat{f})$ will converge to $U(\tf,\f)$ as $n$ and $\widetilde{n}$ simultaneously approach infinity as proved in Theorem \ref{UboundC}. Evidently, $U(\widetilde{f}_{\mathcal{F}}^{\star},f_{\mathcal{F}}^{\star})$ remains non-negative, a consequence of the optimality of $f_{\mathcal{F}}^{\star}$. A lower value of $U(\widetilde{f}_{\mathcal{F}}^{\star},f_{\mathcal{F}}^{\star})$ implies a higher utility of $\mathbb{P}_{\tbx,\ty}$ for the downstream classification task. Therefore, if $U(\tf,\f) > 0 $, $U(\widetilde{f}, \widehat{f})$ will be bounded away from zero with high probability due to the concentration behavior.

%Based on the observations in Figure \ref{fig:Bound_Con}, we make the mild assumption regarding the convergence of $\Delta_1$ and $\Delta_2$.

%\begin{Assumption}
%\label{Ass1}
%    Assume that there exist two positive sequences $\{a_n\}_{n=1}^{\infty}$ and $\{b_{\widetilde{n}}\}_{\widetilde{n}=1}^{\infty}$, both converging to zero, such that $\Delta_1 = O(a_n)$ and $\Delta_2 = O(b_{\widetilde{n}})$.
%\end{Assumption}

%Assumption \ref{Ass1} states that as more data becomes available for downstream training, the resulting model approaches the corresponding optimal classifier within $\mathcal{F}$ in classifying new data.  

\begin{figure}[ht]
\centering
\begin{subfigure}[b]{0.475\textwidth}
\includegraphics[scale=0.475]{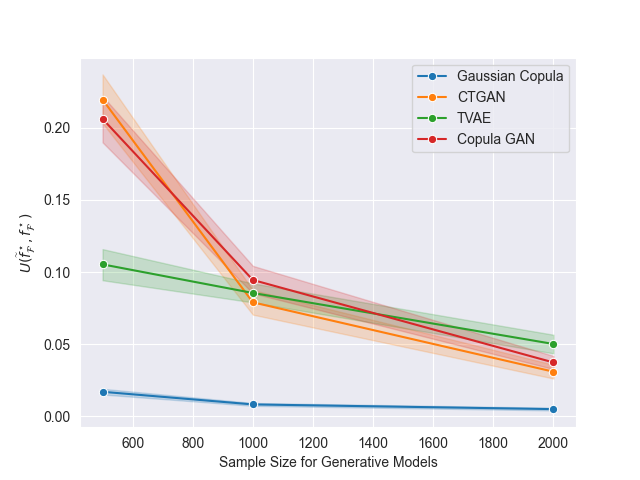}
      %  \caption{Linear SVM}
        \label{fig:subfig1}
    \end{subfigure}
    \begin{subfigure}[b]{0.475\textwidth}
\includegraphics[scale=0.475]{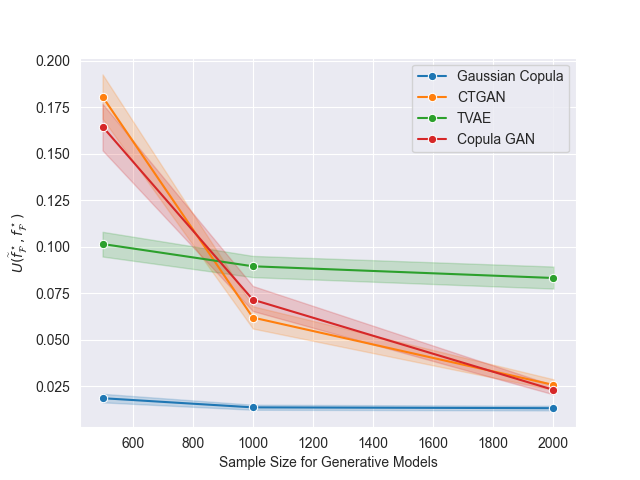}
        %\caption{Decision Tree}
        \label{fig:subfig2}
                    \end{subfigure}
                                \begin{subfigure}[b]{0.475\textwidth}
\includegraphics[scale=0.475]{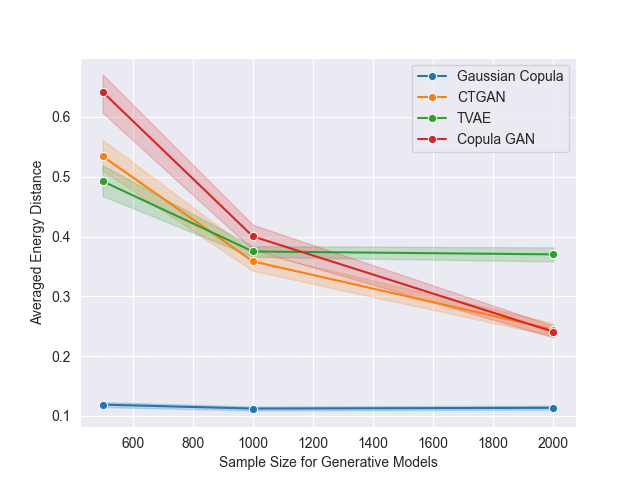}
       % \caption{Energy Distance}
        \label{fig:subfig3}
            \end{subfigure}
            \begin{subfigure}[b]{0.475\textwidth}
\includegraphics[scale=0.475]{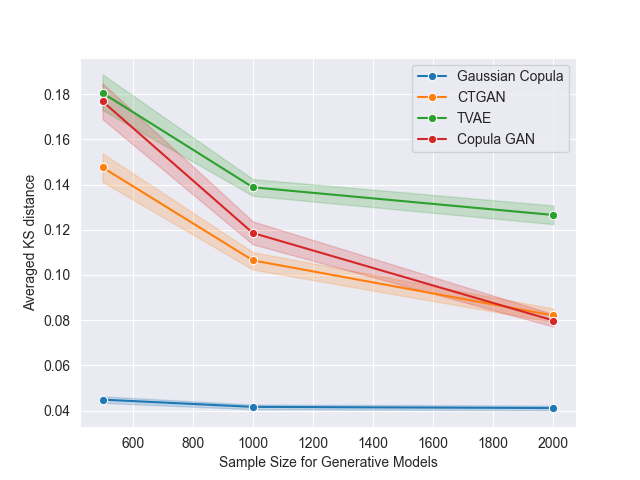}
      %  \caption{KS distance}
        \label{fig:subfig4}
    \end{subfigure}
    \caption{The real samples are generated similarly to those shown in Figure \ref{fig:Bound_Con}, and generative models are trained with sample sizes of \(\{500 \times 2^i, i=0,1,2\}\). We evaluate two downstream tasks: a linear SVM (top left) and a decision tree (top right). The distributional difference between real and synthetic data is measured by the energy distance (bottom left) and the KS distance (bottom right).
}
    \label{fig:UF1}
    \end{figure}

To achieve high utility of synthetic data, the analysis of the utility of synthetic data reduces to specifying the cases $R(\tf) = R(\f)$, i.e., $U(\tf,\f)=0$. A common approach is to ensure that $\mathbb{P}_{\widetilde{\bm{X}},\widetilde{Y}} = \mathbb{P}_{\bm{X},Y}$, signifying perfect alignment between the synthetic data distribution and the real data distribution. In this case, the utility metric $U(\widetilde{f}_{\mathcal{F}}^{\star}, f_{\mathcal{F}}^{\star})$ is exactly zero. This aligns with the objectives of various synthetic data generation algorithms, such as GAN \citep{goodfellow2014generative} and CTGAN \citep{xu2019modeling}, which aim to produce data that closely resembles the real data distribution. Theoretically, as the size of the training data for a generative model increases, the resulting data distribution is expected to more closely resemble the true distribution, leading to a smaller value of $U(\widetilde{f}_{\mathcal{F}}^{\star}, f_{\mathcal{F}}^{\star})$. To validate this hypothesis, we vary the sample sizes used to train different generative models and report the averaged $U(\widetilde{f}_{\mathcal{F}}^{\star}, f_{\mathcal{F}}^{\star})$ of a linear SVM and a decision tree across 100 replications, as shown in Figure \ref{fig:UF1}. Here the decision tree has a maximum depth of 5 and a minimum of 31 sampels per leaf. Additionally, we calculated the average Kolmogorov–Smirnov distance (KS; \citealt{stephens1974edf}) and the energy distance \citep{szekely2013energy} across all dimensions to evaluate the distributional difference between real and synthetic samples.

As illustrated in Figure \ref{fig:UF1}, increasing the amount of training data for the generative models tends to improve the utility of the generated data for the downstream linear SVM and the decision tree classifier. Notably, the patterns of distributional differences, as measured by the energy distance and KS distance, closely align with those of utility. This result shows that the improvement in utility arises from better distribution approximation of synthetic data. In particular, the Gaussian copula model not only achieves the highest utility but also exhibits the smallest distributional difference from the real data. From these results, we can conclude that the primary reason synthetic data generated by the Gaussian copula preserves the greatest utility is that its underlying distribution most closely resembles the real one.

\subsection{Achieving Optimal Utility with Imperfect Synthetic Data}
\label{SubSec:Example}

Even though existing generative models are optimized to approximate the distribution of real data, a key message we deliver here is that perfect alignment between real and synthetic distributions is not a necessary condition for achieving high learning utility. In what follows, we provide two exquisite examples to demonstrate that achieving perfect learning utility ($U(\widetilde{f}_{\mathcal{F}}^{\star},f_{\mathcal{F}}^{\star}) = 0$) is attainable even when the distributions of synthetic and real data differs significantly. These examples and the result in Figure \ref{fig:UF} highlight a significant implication: accurately approximating the real distribution may not be essential to ensure the learning utility of synthetic data.

\begin{example}
\label{Exam:1}
Assume that $\mathcal{X}=[-1,1]$ and $\widetilde{\eta}(x)=\eta(x)=4/5$ for any $x\in (0,1]$ and $\widetilde{\eta}(x)=\eta(x)=1/5$ otherwise. Let the probability density functions of $X$ and $\widetilde{X}$ be
\begin{align*}
P_{X}(x) =
\begin{cases}
1-\alpha, &\mbox{ if } x\in [0,1],\\
\alpha, &\mbox{ if } x \in [-1,0).
\end{cases}
\mbox{ and  }
P_{\widetilde{X}}(x) =
\begin{cases}
\alpha, &\mbox{ if } x\in [0,1],\\
1-\alpha,&\mbox{ if } x \in [-1,0).
\end{cases}
\end{align*}
Consider $\mathcal{F}=\{f(x)=\sign(\beta x): \beta \in \mathbb{R} \}$. We can easily see that $R(\f)=R(\tf)=1/5$ even though $\mathbb{P}_{X} \neq\mathbb{P}_{\widetilde{X}}$.
\end{example}

Example \ref{Exam:1} explores the scenario where synthetic labels are generated perfectly, i.e., $\widetilde{\eta}=\eta$, but the synthetic feature distribution differs from the real one. Furthermore, the downstream learning task involves a flexible function class $\mathcal{F}$ such that $R(f_{\mathcal{F}}^{\star})=0$. This example illustrates that consistent generalization remains attainable, provided that $\mathcal{F}$ is flexible enough, even in the presence of dissimilar synthetic feature distribution.

\begin{example}
\label{Exam:2}
Let $X$ and $\widetilde{X}$ follow the same distribution $P_{\widetilde{X}}(x)=P_{X}(x) = 1/2$ for $x \in [-1/2,1]$. Assume that $\widetilde{\eta}(x)=1-\beta$ and $\eta(x)=1$ for any $x\in (0,1]$, and $\widetilde{\eta}(x)=2\beta$ and $\eta(x)=0$ otherwise. Given that $\mathcal{F}=\{f(x)=\sign(a |x|): a \in\mathbb{R} \}$ and any $\beta \in [0,1/2]$, it holds that $\tf=\f$, which implies $R(\tf)=R(\f)$.
\end{example}

Example \ref{Exam:2} illustrates an intriguing scenario where consistent generalization remains achievable, even in the case where the generation of synthetic labels is imperfect and the downstream function class is not flexible enough. In this context, $\mathbb{P}_{\widetilde{Y}|\widetilde{\bm{X}}}$ fails to accurately approximate $\mathbb{P}_{Y|\bm{X}}$. This result implies that approximating the underlying regression function is not a prerequisite for generating utility-preserving synthetic data. In this example, it only requires $\widetilde{\eta}$ and $\eta$ to align in their tendency to generate labels for ensuring $R(\tf)=R(\f)$, that is $\sign(\widetilde{\eta}(\bm{x})-1/2)=\sign(\eta(\bm{x})-1/2)$ for all $x$.

From the above discussion, we can draw two main conclusions. First, achieving high learning utility does not necessarily require generative models to approximate the real data distribution. This finding is particularly useful since distribution approximation is often intractable for high-dimensional data, especially when real data is limited. Second, learning utility also depends on the choice of the downstream learning task.

\section{Analysis of Utility Metric}
\label{Sec:UB}

In this section, we establish an analytic upper bound for the utility metric $U(\tf,\f)$, aiming to quantify the influence of the model specification in the downstream learning task $\mathcal{F}$, and the synthetic data distribution $\mathbb{P}_{\widetilde{\bm{X}},\widetilde{Y}}$ on the behavior of $U(\tf,\f)$. Through our analytic bounds, we show that the impacts of $\mathcal{F}$ and $\mathbb{P}_{\widetilde{\bm{X}},\widetilde{Y}}$ are interactive. Specifically, it is unnecessary for $\mathbb{P}_{\widetilde{\bm{X}},\widetilde{Y}}$ to resemble $\mathbb{P}_{\bm{X},Y}$ under some specific model specifications for achieving the convergence of the proposed utility metric. %In this paper, we consider scenario that $\bm{X}$ and $\widetilde{\bm{X}}$ have the same supports. %Similar results for the classification setting can be found in supplementary file.
\begin{figure}[ht]
\centering
\includegraphics[scale=0.23]{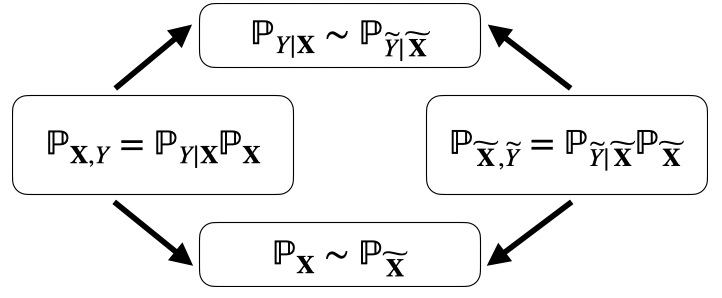}
\caption{The similarity between $\mathbb{P}_{\bm{X},Y}$ and $\mathbb{P}_{\widetilde{\bm{X}},\widetilde{Y}}$ can be decomposed into two key aspects: (1) \textbf{feature fidelity}: quantifying the proximity between $\mathbb{P}_{\bm{X}}$ and $\mathbb{P}_{\widetilde{\bm{X}}}$; and (2) \textbf{functional relationship estimation}: assessing the effectiveness of $\mathbb{P}_{\widetilde{Y}|\widetilde{\bm{X}}}$ in capturing the underlying functional relationship inherent in $\mathbb{P}_{Y|\bm{X}}$.}
\label{Fig:Decomposition}
\end{figure}

As depicted in Figure \ref{Fig:Decomposition}, rather than directly assessing the similarity between $\mathbb{P}_{\bm{X},Y}$ and $\mathbb{P}_{\widetilde{\bm{X}},\widetilde{Y}}$, the examination of the impact of the synthetic data distribution $\mathbb{P}_{\widetilde{\bm{X}},\widetilde{Y}}$ is decomposed into two distinct components: feature fidelity and functional relationship estimation. In this study, feature fidelity is defined as the similarity between real and synthetic features, while functional relationship estimation pertains to how well the relationship between features and responses within $\mathbb{P}_{\widetilde{\bm{X}},\widetilde{Y}}$ approximates that of $\mathbb{P}_{\bm{X},Y}$.

To assess the similarity between $\mathbb{P}_{\bm{X}}$ and $\mathbb{P}_{\widetilde{\bm{X}}}$, we employ the Integral Probability Metric (IPM), a widely used measure in synthetic data generation \citep{bousquet2020synthetic}. This metric is defined as follows.
\begin{definition}[]
The IPM between two probability measures $\mathbb{P}_{\bm{X}}$ and $\mathbb{P}_{\widetilde{\bm{X}}}$ with respect to a function class $\mathcal{H}$ is defined as
\begin{align}
\label{Metric:IPM}
D_{\mathcal{H}}(\mathbb{P}_{\bm{X}},\mathbb{P}_{\widetilde{\bm{X}}})=
\sup_{h \in \mathcal{H}}
\left|
\mathbb{E}_{\bm{X}\sim \mathbb{P}_{\bm{X}}}\Big[
h(\bm{X})
\Big]-
\mathbb{E}_{\widetilde{\bm{X}} \sim \mathbb{P}_{\widetilde{\bm{X}}}}\Big[
h(\widetilde{\bm{X}})
\Big]
\right|,
\end{align}
where $\mathcal{H}$ is a set of real-valued measurable functions on $\mathcal{X}$ such that $\sup_{h \in \mathcal{H}}\mathbb{E}_{\bm{X}\sim \mathbb{P}_{\bm{X}}}\Big[
h(\bm{X})
\Big]<\infty$ and $\sup_{h \in \mathcal{H}} \mathbb{E}_{\bm{X}\sim \mathbb{P}_{\widetilde{\bm{X}}}}\Big[
h(\bm{X})
\Big]<\infty$.
\end{definition}

As for the impact of downstream task $\mathcal{F}$, we consider the following metric:
\begin{align}
\Lambda(\mathcal{F}) = \max\left\{  \mathbb{P}\Big( \f(\bm{X}) \neq f^\star(\bm{X}) \Big),
      \mathbb{P}\Big( \tf(\bm{X}) \neq \widetilde{f}^\star(\bm{X}) \Big)\right\}.
\end{align}
Here, $\Lambda(\mathcal{F})$ measures the probability that the optimal classifier within $\mathcal{F}$ fails to make the same prediction on real unobserved data as the Bayes classifier under both the real and synthetic distributions. In other words, $\Lambda(\mathcal{F})$ decreases as the flexibility of $\mathcal{F}$ increases, and $\Lambda(\mathcal{F}) = 0$ if $f^\star \in \mathcal{F}$, which indicates correct model specification in the downstream task.

For evaluating the quality of relationship estimation, we consider the following two metrics:
\begin{align}
\label{Metric:Fun}
\Upsilon(\widetilde{\eta})= \mathbb{E}
\left\{
I \left(
f^{\star}(\bm{X}) \neq \widetilde{f}^{\star}(\bm{X})
\right)
\left|\widetilde{\eta}(\bm{X})-\eta(\bm{X})\right|
\right\}   \mbox{ and }
 \Vert \widetilde{\eta}-\eta\Vert_{L^2(\mathbb{P}_{\bm X})}.
\end{align}
Here, $\Upsilon(\widetilde{\eta})$ evaluates whether $\widetilde{\eta}$ has the same tendency as $\eta$ in generating labels, whereas $\Vert \widetilde{\eta} - \eta \Vert_{L^2(\mathbb{P}_{\bm{X}})}$ measures the $L_2$-error of $\widetilde{\eta}$ in approximating $\eta$. It is worth noting that $\Upsilon(\widetilde{\eta}) = 0$ only entails $f^\star(\bm{x}) = \widetilde{f}^\star(\bm{x})$ almost surely. In other words, $\Vert \widetilde{\eta} - \eta \Vert_{L^2(\mathbb{P}_{\bm{X}})} = 0$ is a stronger requirement for relationship estimation, which will result in $\Upsilon(\widetilde{\eta}) = 0$ automatically.

\subsection{An Analytic Utility Bound}
\label{SubSec:ANABound}
Two examples in Section \ref{SubSec:Example} show that the utility metric is inherently influenced by several factors: the degree of feature fidelity, the disparity between \(\mathbb{P}_{\widetilde{Y}|\widetilde{\bm{X}}}\) and \(\mathbb{P}_{Y|\bm{X}}\), and the choice of \(\mathcal{F}\). A noteworthy observation is that these components do not affect the utility independently. Therefore, we aim to develop a general utility bound based on (\ref{Metric:IPM})-(\ref{Metric:Fun}), which correspond to the impacts of feature fidelity, the downstream task \(\mathcal{F}\), and functional relationship estimation, respectively.

\begin{theorem}
For any class of classifiers $\mathcal{F}$ and $\mathbb{P}_{\widetilde{\bm{X}},\widetilde{Y}}$, it holds that
\label{Thm:UTBmore}
\begin{align}
\label{UUUCC}
U(\tf,\f) 
\leq &2D_{\mathcal{K}}(\mathbb{P}_{\bm{X}},\mathbb{P}_{\widetilde{\bm{X}}})
+ 4 \Lambda(\mathcal{F}) \Vert \widetilde{\eta}-\eta\Vert_{L^2(\mathbb{P}_{X})}+
4 \Upsilon(\widetilde{\eta}),
\end{align}
where $\mathcal{K} = \left\{k(\bm{x})=|2\widetilde{\eta}(\bm{x})-1|I\big(f(\bm{x}) \neq \widetilde{f}^\star(\bm{x})\big), f \in \{f_{\mathcal{F}}^\star,\widetilde{f}^\star_{\mathcal{F}}\} \right\}$.
\end{theorem}

Theorem \ref{Thm:UTBmore} establishes an upper bound for the proposed utility metric based on (\ref{Metric:IPM})-(\ref{Metric:Fun}). This upper bound encompasses the impacts of feature fidelity, the downstream task, and the estimation of the functional relationship. In Table \ref{tab:utility}, all corresponding components are specified to illustrate their roles in affecting $U(\tf, \f)$.

\begin{table}[ht]

\centering
\caption{Impact of Different Factors on the Utility Metric $U(\tf,\f)$.}
\label{tab:utility}
\begin{tabular}{lc|>{\raggedright\arraybackslash}p{7.4cm}}
\toprule[2pt]
\textbf{Aspects} & \textbf{Metric} & \textbf{Analysis}\\
\midrule
Downstream Task $\mathcal{F}$ & $\Lambda(\mathcal{F})$ &
(1) $\Lambda(\mathcal{F})=0$ if $f^\star \in \mathcal{F}$ and $f^\star=\widetilde{f}^\star$
\\
\hline
\multirow{2}{*}{Relationship Estimation} & $ \Vert \widetilde{\eta}-\eta\Vert_{L^2(\mathbb{P}_{\bm X})}$ & (1) $\Vert \widetilde{\eta}-\eta\Vert_{L^2(\mathbb{P}_{\bm X})}=0$ if $\widetilde{\eta}(\bm{x}) =\eta(\bm{x})$ a.s.  \\
& $ \Upsilon(\widetilde{\eta})$  & $ (1)\Upsilon(\widetilde{\eta})=0$ if $\widetilde{f}^\star(\bm{x})=f^\star(\bm{x})$ a.s.\\
\hline
\multirow{2}{*}{Feature Fidelity} & \multirow{2}{*}{$D_{\mathcal{K}}(\mathbb{P}_{\bm{X}},\mathbb{P}_{\widetilde{\bm{X}}})$} & 
(1) $D_{\mathcal{K}}(\mathbb{P}_{\bm{X}},\mathbb{P}_{\widetilde{\bm{X}}})=0$ if $\mathbb{P}_{\bm{X}} = \mathbb{P}_{\widetilde{\bm{X}}}$ \\
& & (2)
$D_{\mathcal{K}}(\mathbb{P}_{\bm{X}},\mathbb{P}_{\widetilde{\bm{X}}})=0$ if $f^\star \in \mathcal{F}$ and $f^\star=\widetilde{f}^\star$
\\
\bottomrule[2pt]
\end{tabular}
\end{table}

Table \ref{tab:utility} examines the influence of three key aspects on the utility metric and summarizes the scenarios in which they equal zero. First, the impact of $\mathcal{F}$ on utility can be categorized into two cases based on the flexibility of $\mathcal{F}$. If $f^\star \in \mathcal{F}$ and $f^\star = \widetilde{f}^\star$, then $\Lambda(\mathcal{F})$ is zero. Otherwise, $\Lambda(\mathcal{F})$ is bounded away from zero for any less flexible $\mathcal{F}$. Second, the functional relationship estimation is characterized by two metrics in (\ref{UUUC}): $\Vert \widetilde{\eta} - \eta \Vert_{L^2(\mathbb{P}_{X})}$ and $\Upsilon(\widetilde{\eta})$. From the utility bound in (\ref{UUUC}), it can be seen that if the downstream task satisfies $\Lambda(\mathcal{F})=0$, we only require $\Upsilon(\widetilde{\eta}) = 0$ to achieve $U(\widetilde{f}^\star, f^\star) = 0$. Otherwise, we must ensure $\Vert \widetilde{\eta} - \eta \Vert_{L^2(\mathbb{P}_{X})} = 0$. Third, the feature fidelity is characterized by $D_{\mathcal{K}}(\mathbb{P}_{\bm{X}}, \mathbb{P}_{\widetilde{\bm{X}}})$, which measures the closeness between $\mathbb{P}_{\bm{X}}$ and $\mathbb{P}_{\widetilde{\bm{X}}}$ over a specific function class $\mathcal{K}$ that comprises only two functions.
 Consider a simple scenario where $f^\star = \widetilde{f}^\star$, then the convergence of $D_{\mathcal{K}}(\mathbb{P}_{\bm{X}}, \mathbb{P}_{\widetilde{\bm{X}}})$ can be analyzed in the following two cases:
\begin{enumerate}
\item \textbf{Perfect Feature Fidelity}: If $\mathbb{P}_{\bm{X}} = \mathbb{P}_{\widetilde{\bm{X}}}$, we have $D_{\mathcal{K}}(\mathbb{P}_{\bm{X}},\mathbb{P}_{\widetilde{\bm{X}}})=0$ regardless of $\mathcal{F}$.
\item \textbf{Correct Model Specification}: If the model specification in the downstream task is correct such that $f^\star \in \mathcal{F}$, then we have $D_{\mathcal{K}}(\mathbb{P}_{\bm{X}},\mathbb{P}_{\widetilde{\bm{X}}}) =0$ for any $\mathbb{P}_{\widetilde{\bm{X}}}$.
\end{enumerate}

The second result is intriguing because $D_{\mathcal{K}}(\mathbb{P}_{\bm{X}}, \mathbb{P}_{\widetilde{\bm{X}}}) = 0$ does not imply that $\mathbb{P}_{\bm{X}} = \mathbb{P}_{\widetilde{\bm{X}}}$. In other words, the utility metric $U(\tf,\f)$ is also influenced by $\mathcal{F}$. Below, we present another scenario where perfect utility is achieved even when the synthetic distribution is not identical to the real one.

\begin{corollary}
\label{Cro:Down}
If $\mathcal{F}$ is appropriately chosen such that $\f = f^\star$ and $\sign(\widetilde{\eta}(\bm{x})-1/2)=\sign(\eta(\bm{x})-1/2)$ for any $\bm{x}\in \mathcal{X}$, it then follows that $U(\tf,\f)=0$ for any synthetic feature distribution $\mathbb{P}_{\widetilde{\bm{X}}}$.
\end{corollary}

\begin{figure}[ht]
\centering
            \begin{subfigure}[b]{0.49\textwidth}
\includegraphics[scale=0.3]{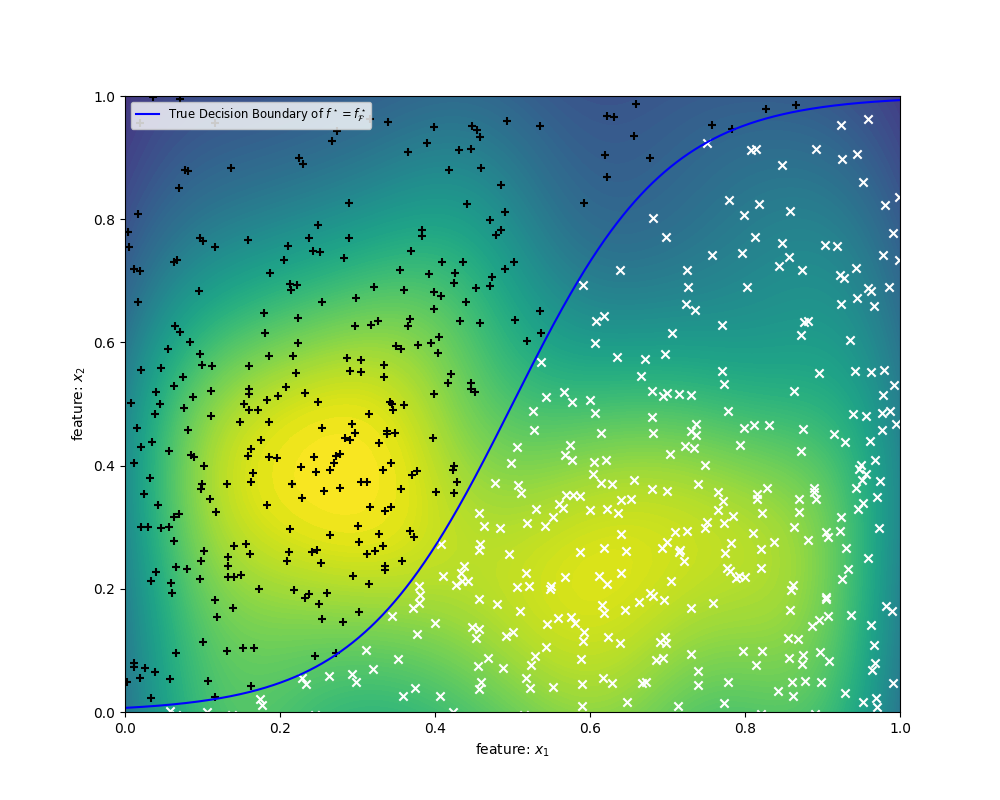}
            \end{subfigure}
            \begin{subfigure}[b]{0.49\textwidth}
\includegraphics[scale=0.3]{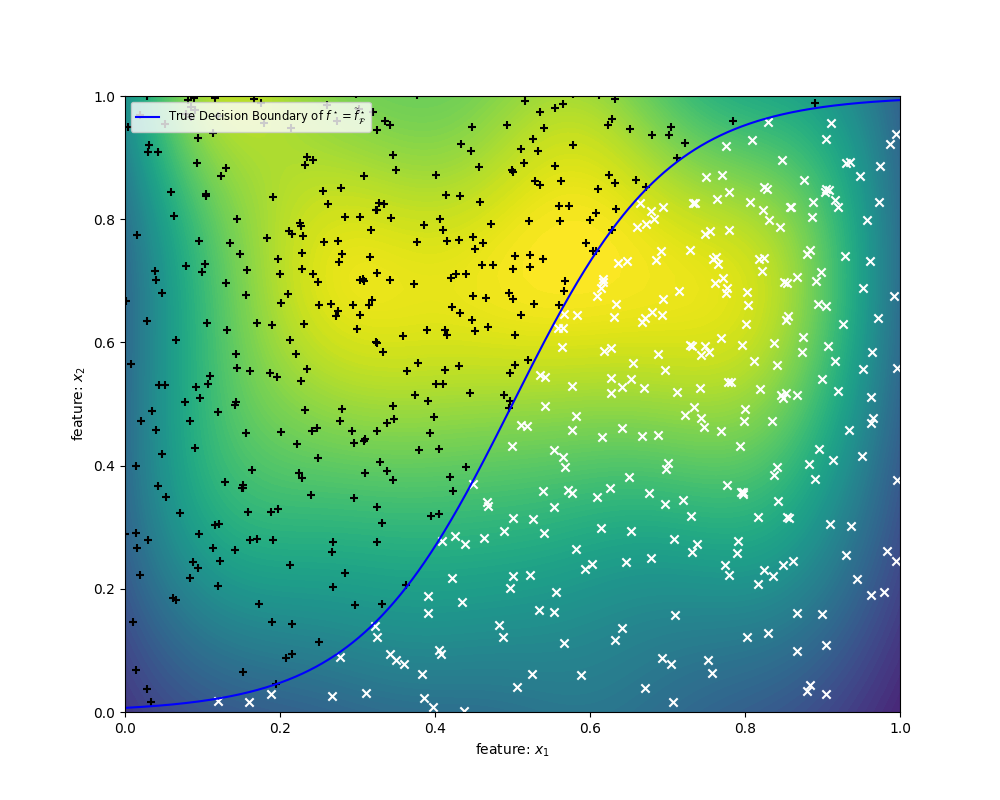}
    \end{subfigure}
    \caption{An example of Corollary \ref{Cro:Down}. In this example, the downstream task accurately specifies the true decision boundary ($f^\star=\f$). Given that $f^\star = \widetilde{f}^\star$, the estimated decision boundary of $\widetilde{f}$ under the synthetic distribution is identical to that of $\f$. This result remains invariant to changes in the density of $\widetilde{\bm{X}}$.
}
    \label{fig:UF2}
    \end{figure}

Corollary \ref{Cro:Down} is a direct result of Theorem \ref{Thm:UTBmore}, illustrating that if the model specification is accurate, achieving perfect feature fidelity is {\em not} necessary for the downstream task when the downstream function class $\mathcal{F}$ is substantially large such that $f^\star \in \mathcal{F}$. Our theoretical findings explain a prevalent phenomenon \citep{tremblay2018training,azizi2023synthetic} wherein synthetic data can significantly enhance the performance of downstream tasks. This enhancement stems from the inherent flexibility of the downstream function class, facilitating improved generalization. For illustration, we present an example in Figure \ref{fig:UF2}. Clearly, as the downstream learning task is sufficiently flexible to specify the true decision boundary, consistent generalization performance, i.e., $U(\tf,\f)=0$, is achievable regardless of the density of $\widetilde{\bm{X}}$.

\subsection{Worst-Case Utility Bound under Low-Noise Condition}
Section \ref{SubSec:ANABound} analyzed the utility of synthetic data for a specific downstream task $\mathcal{F}$. In practical scenarios, data holders typically lack knowledge about the classification models that will be evaluated on the released synthetic data. For example, in Kaggle competitions, competitors might test various classification models on the provided dataset \citep{hittmeir2019utility}. Therefore, data holders should guarantee the utility for any classification model. Additionally, \citet{rankin2020reliability} assess the utility of synthetic medical data using various classification methods, demonstrating that different utility values are observed for different classification methods on the same synthetic dataset. This phenomenon underscores the importance of considering the worst-case utility of synthetic data. Consequently, we can define
\begin{equation}
    \label{Eqn:WorstCase}
    \text{Worst-Case Utility}: \overline{U}(\mathbb{P}_{\widetilde{\bm{X}},\widetilde{Y}})=\sup_{\mathcal{F}} U(\tf,\f).
\end{equation}
The worst-case utility is defined as the supremum of $U(\tf, \f)$ across all downstream tasks. In this context, the metric should depend solely on the synthetic distribution $\mathbb{P}_{\widetilde{\bm{X}},\widetilde{Y}}$.

Next, a natural question arises: How does $\mathbb{P}_{\widetilde{\bm{X}}, \widetilde{Y}}$ affect $\overline{U}(\mathbb{P}_{\widetilde{\bm{X}}, \widetilde{Y}})$? To address this question, we start with the following common assumption in classification \citep{AnnalsTyspart, bartlett2006convexity}, which facilitates a more precise analysis of the utility bound. 

\begin{Assumption}[Low-Noise Condition] 
\label{Ass:Low_noise}
There exist positive constants $C_0$, $\widetilde{C}_0$, $\gamma$, and $\widetilde{\gamma}$ such that for any $t > 0$,
\[
\mathbb{P}\left(|\eta(\bm{X}) - \frac{1}{2}| \leq t\right) \leq C_0 t^{\gamma} \quad \text{and} \quad \mathbb{P}\left(|\widetilde{\eta}(\bm{X}) - \frac{1}{2}| \leq t\right) \leq \widetilde{C}_0 t^{\widetilde{\gamma}},
\]
where the probability measure is $\mathbb{P}_{\bm{X}}$.
\end{Assumption}
In Assumption \ref{Ass:Low_noise}, the parameters $C_0$ and $\gamma$ are used to characterize the behavior of the regression function $\eta$ in the vicinity of the level $\eta(\bm{x}) = 1/2$. Specifically, a larger value of $\gamma$ or a smaller value of $C_0$ indicates lower noise in the labels. Theoretically, lower label noise facilitates the learning of the optimal classifier. In this paper, we use different pairs $(C_0, \gamma)$ and $(\widetilde{C}_0, \widetilde{\gamma})$ to represent the label noise in real and synthetic labels, respectively. Specifically, if $\widetilde{\eta} = \eta$, then $(C_0, \gamma) = (\widetilde{C}_0, \widetilde{\gamma})$. Under Assumption \ref{Ass:Low_noise}, we present the following theorem, which provides an upper bound for the worst-case utility.

\begin{theorem}[\textbf{Worst-Case Utility Bound}]
\label{Thm:WUB}
Under Assumption \ref{Ass:Low_noise}, for any $\mathbb{P}_{\widetilde{\bm{X}},\widetilde{Y}}$, there exists some positive constant $C_1$ such that
\label{Thm:UTBmoreFinal}
\begin{align}
\label{UTB:Sup}
\overline{U}(\mathbb{P}_{\widetilde{\bm{X}}, \widetilde{Y}})
\leq 2\mathrm{TV}(\mathbb{P}_{\bm{X}},\mathbb{P}_{\widetilde{\bm{X}}})
+ C_1 \left(\Vert \widetilde{\eta}-\eta\Vert_{L^2(\mathbb{P}_{X})}+\Vert \widetilde{\eta}-\eta  \Vert_{L^2(\mathbb{P}_{\bm{X}})}^{\frac{2\gamma^\prime+2}{\gamma^\prime+2}}\right),
    \end{align}
    where $\gamma^\prime=\max\{\gamma,\widetilde{\gamma}\}$ with $\gamma$ and $\widetilde{\gamma}$ being defined in Assumption \ref{Ass:Low_noise}.
\end{theorem}

Theorem \ref{Thm:UTBmoreFinal} shows that the worst-case utility can be bounded by the total variation distance between the real and synthetic feature distributions (feature fidelity) and the $L_2$ error of $\widetilde{\eta}$ in approximating $\eta$ (relationship estimation). It is worth noting that $\frac{2\gamma^\prime+2}{\gamma^\prime+2} \geq 1$ for any $\gamma' \geq 0$. Therefore, $\Vert \widetilde{\eta} - \eta \Vert_{L^2(\mathbb{P}_{X})}$ is the dominant term when evaluating the quality of relationship estimation.

\begin{figure}[ht]
    \centering
    \includegraphics[scale=0.3]{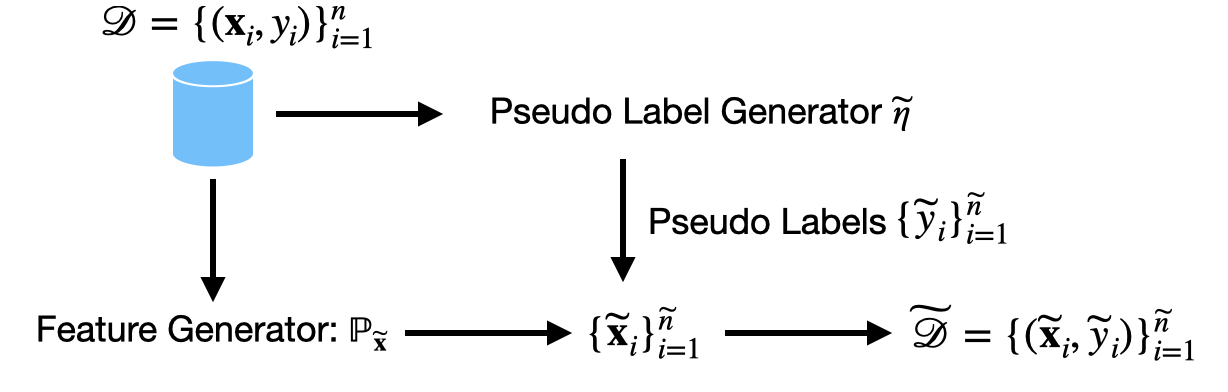}
    \caption{A two-stage generation framework of subsequent synthetic datasets for learning tasks \citep{carmon2019unlabeled, gowal2021improving}.}
    \label{fig:FrameSyn}
\end{figure}

Next, we use the utility bound in (\ref{UTB:Sup}) to analyze a popular framework for generating synthetic data for learning tasks \citep{carmon2019unlabeled, gowal2021improving}. This framework involves generating synthetic features using a generative model and training a classifier to assign labels. For illustration, we present the general process of this framework in Figure \ref{fig:FrameSyn}. 

In the following, we will analyze the asymptotic behavior of the worst-case utility of a specific example within the framework illustrated in Figure \ref{fig:FrameSyn}. Specifically, we assume that the generative adversarial framework is employed for feature generation. In this framework, a common practice is to use a neural network, denoted as $g_{\bm{\theta}}(\cdot)$ and parametrized by $\bm{\theta}$, which takes random noise $\bm{Z}$ from a prespecified distribution as input and outputs the resulting synthetic data. Specifically, for a random sample $\bm{z}_i$, $ g_{\bm{\theta}}(\bm{z}_i) $ produces a synthetic sample $\widetilde{\bm{x}}_i$, that is, $\widetilde{\bm{x}}_i = g_{\bm{\theta}}(\bm{z}_i)$. Within this framework, the neural generator $g_{\bm{\theta}}$ is optimized by minimizing the following minimax problem,
\begin{align}
\label{GAN:generator}
\text{Feature Generator: }
g_{\widehat{\bm{\theta}}}=
\argmin_{g_{\bm{\theta}} \in \mathcal{G}}
\max_{h \in \mathcal{H}_{dis}}
\left\{
    \underset{\widetilde{\bm{X}} \sim g_{\bm{\theta}}(\bm{Z})}{\mathbb{E}} h(\widetilde{\bm{X}})-
    \underset{\bm{X} \sim \widehat{\mathbb{P}}_{\bm{X}}}{\mathbb{E}}
    h(\bm{X})
    \right\},
\end{align}
where $\mathcal{G}$ denotes the class of generators, $\mathcal{H}_{dis}$ denotes the class of discriminators, and $\mathbb{P}_{\bm{Z}}$ denotes the distribution of some random noise $\bm{Z}$. Here, $\widehat{\mathbb{P}}_{\bm{X}}$ is an constructed empirical distribution of $n$ real features $\mathcal{D}_{\bm{X}}=\{\bm{x}_i\}_{i=1}^n$. For example, $\widehat{P}_{\bm{X}}(\bm{x})$ can be $\frac{1}{n}\sum_{i=1}^n \delta_{\bm{x}}(\mathcal{D}_{\bm{X}})$ with $\delta_{\bm{x}}(\cdot)$ being the Dirac measure or a smoothed version of the empirical distribution such as $\frac{1}{nh}\sum_{i=1}^n K_h( \bm{x}_i-\bm{x})$ considered in \citet{liang2021well}.

To facilitate the analysis of worst-case utility in this example, we impose the following assumption for the generative adversarial framework: the discriminator function class consists of all functions bounded by 1, while the generator is capable of capturing the underlying feature distribution. The latter assumption implies zero distribution approximation error for the neural generators, a claim that has been empirically validated in various applications.

\begin{Assumption}
\label{Ass:Gen}
Assume that $\mathcal{X}=[0,1]^p$ and $\mathcal{H}_{dis}=\{h(\bm{x}):\Vert h \Vert_{\infty} \leq 1\}$ is the class of all measurable functions whose sup norm is bounded by 1. Furthermore, $\mathcal{G}$ is chosen such that $\min_{g_{\bm{\theta}} \in \mathcal{G}}
    \mathrm{TV}(\mathbb{P}_{\bm{X}},\mathbb{P}_{g_{\bm{\theta}}(\bm{Z})})=0$.
\end{Assumption}

For pseudo label generation, we consider a class of fully-connected neural network aimed at approximating the regression function $\eta$. Specifically, the label generator is obtained through the following task:
\begin{align}
    \text{Pseudo Label Generator: } 
    L_{\widehat{\bm{\omega}}}= \argmin_{L_{\bm{\omega}} \in \mathcal{N}(W,L)}\frac{1}{n}\sum_{i=1}^n 
    \left( y_i^\prime - L_{\bm \omega}(\bm{x}_i)\right)^2,
\end{align}
where $y_i^\prime = (y_i+1)/2$ and $L_{\bm{\omega}}$ represents a neural network parametrized by $\bm{\omega}$. After obtaining $L_{\widehat{\bm{\omega}}}$, we can employ a clipped estimator $\widetilde{\eta}(\bm{x})=\max\{1,\min\{L_{\widehat{\bm{\omega}}}(\bm{x}),0\}\}$ as in \citet{nakada2020adaptive}.

Next, we impose an assumption regarding the smoothness of the true feature density function $ P_{\bm{X}} $ and the regression function $ \eta $. Specifically, we assume that the weak derivatives of order $\alpha$ for $ P_{\bm{X}} $ and order $\beta$ for $ \eta $ are squared integrable. This assumption is mild and generally imposed in the domain of learning theory for neural networks \citep{yang2024optimal,liang2021well,siegel2023optimal}.

\begin{Assumption}[Feature and Label Smootheness]
\label{Ass:FeaSmoo}
    Assume that there exists smoothness parameters $\alpha,\beta \in \mathbb{N}$ and a radius $r$ such that 
    \begin{align*}
        \max\left\{
\left(\sum_{\bm{\gamma}:\Vert \bm{\gamma}\Vert_1 \leq \alpha} \Vert D^{(\bm{\gamma})}P_{\bm{X}} \Vert_{L^2}^2\right)^{\frac{1}{2}},
\left(\sum_{\bm{\gamma}:\Vert \bm{\gamma}\Vert_1 \leq \beta} \Vert D^{(\bm{\gamma})}\eta \Vert_{L^2}^2\right)^{\frac{1}{2}}
        \right\} \leq r,
    \end{align*}
where $\bm{\gamma}=(\gamma_1,\ldots,\gamma_p)$ is a multi-index and $D^{(\bm{\gamma})}P_{\bm{X}}$ represents the $\bm{\gamma}$-weak derivative of the real density function $P_{\bm{X}}$.
\end{Assumption}

\begin{theorem}
    \label{Thm:ConditGene}
    Under Assumptions \ref{Ass:Low_noise}-\ref{Ass:FeaSmoo}, there exists $\mathcal{N}(W,L)$ with $L \asymp n^{\frac{p}{4\beta+2p}}$ and fixed $W$ such that the following holds:
    \begin{align}
\mathbb{E}_{\mathcal{D}}
    \left[
  \overline{U}(\mathbb{P}_{\widetilde{\bm{X}}, \widetilde{Y}})
  \right]
\lesssim 
n^{-\frac{\alpha}{2\alpha+p}} \vee n^{-\frac{\beta}{2\beta+p}},
\end{align}
where $\mathbb{E}_{\mathcal{D}}(\cdot)$ denotes the expectation taken with respect to the real dataset $\mathcal{D}=\{(\bm{x}_i,y_i)\}_{i=1}^n$ and $\lesssim$ omits the logarithmic term.
\end{theorem}

Theorem \ref{Thm:ConditGene} shows that the expected worst-case utility converges to zero at a rate of $O(n^{-\frac{\alpha}{2\alpha+p}} \vee n^{-\frac{\beta}{2\beta+p}})$. This convergence rate is determined by the lower smoothness between $P_{\bm{X}}$ and $\eta$. This result shows that as more training samples are used for generating synthetic data under the framework illustrated in Figure \ref{fig:FrameSyn}, the resulting model for any downstream task will converge to the optimal model trained on the real data distribution. This provides a theoretical explanation for why synthetic data significantly improves downstream training, as observed in \citet{gowal2021improving}.

\section{Model Comparison Based on Synthetic Data}
\label{Sec:MC}
In addition to consistent generalization, another use of synthetic data involves ranking model performances solely based on synthetic data \citep{jordon2018measuring, jordon2022synthetic}. This necessitates consistency in the ranking of model performances between synthetic and real data, which is especially crucial when synthetic data is employed in competitions, such as Kaggle. In this section, we study sufficient conditions on the synthetic data distribution that ensure that the relative generalization performance of models trained from synthetic data remains consistent with that of models trained from real data. We first present a definition for consistent model comparison based on synthetic data. 
\begin{definition}
(Consistent Model Comparison) Let $(\mathcal{F}_1,\mathcal{F}_2)$ be a pair classes of classification models for comparison in the downstream learning task. We say the synthetic distribution $\mathbb{P}_{\widetilde{\bm{X}},\widetilde{Y}}$ preserves the utility of consistent model comparison if
\begin{align*}
\Big(R(f_{\mathcal{F}_1}^\star)-R(f_{\mathcal{F}_2}^\star)\Big)
\Big(R(\widetilde{f}_{\mathcal{F}_1}^\star)-R(\widetilde{f}_{\mathcal{F}_2}^\star)\Big)>0,
\end{align*}
where $\widetilde{f}_{\mathcal{F}_i}^\star =\argmin_{f\in \mathcal{F}_i}\widetilde{R}(f)$ for $i=1,2$.
\end{definition}

\begin{figure}[ht]
\centering
\includegraphics[scale=0.24]{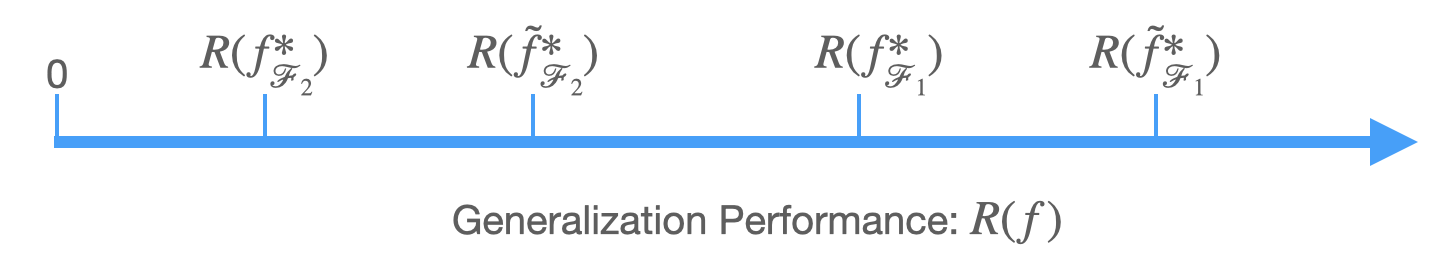}
\caption{An illustrative example for consistent model comparison when the synthetic and real distributions are not the same.}
\label{fig-MC}
\end{figure}

Particularly, when the synthetic and real distributions are identical, the optimal models will necessarily be the same. As a result, it is straightforward to derive a consistent model comparison based on synthetic data. However, more interestingly, consistent model comparison can still be achieved using synthetic data even if the synthetic and real distributions are not identical. To illustrate this phenomenon, we provide an intuitive example in Figure \ref{fig-MC}. In this example, we assume that $f_{\mathcal{F}_2}^\star$ outperforms $f_{\mathcal{F}_1}^\star$ in generalization performance, i.e., $R(f_{\mathcal{F}_2}^\star)<R(f_{\mathcal{F}_1}^\star)$. According to the analytic bound established in Theorem \ref{Thm:UTBmore}, the gap between $R(f_{\mathcal{F}_2}^\star)$ and $R(\widetilde{f}_{\mathcal{F}_2}^\star)$ resulting from the difference between the synthetic and real distributions will vanish as the synthetic distribution becomes more similar to the real one. 

This example demonstrates that consistent model comparison is possible if the synthetic distribution is constructed such that $R(\widetilde{f}_{\mathcal{F}_2}^\star)$ falls in the interval $[R(f_{\mathcal{F}_2}^\star),R(f_{\mathcal{F}_1}^\star)]$, providing a sufficient condition for consistent comparison on synthetic data. In other words, consistent model comparison between $\mathcal{F}_1$ and $\mathcal{F}_2$ can be guaranteed if
\begin{align}
 U(\widetilde{f}_{\mathcal{F}_2}^\star,f_{\mathcal{F}_2}^\star)  \leq R(f_{\mathcal{F}_1}^\star)-R(f_{\mathcal{F}_2}^\star)
 \Leftrightarrow   R(\widetilde{f}_{\mathcal{F}_2}^\star)\leq R(f_{\mathcal{F}_1}^\star).
\end{align}
Therefore, consistent model comparison is a less stringent requirement on synthetic data and does not necessarily imply consistent generalization. Motivated by this example, we aim to explicate the sufficient conditions for synthetic data to yield identical model comparison as that of real data.

\subsection{An Example for Inconsistent Model Comparison}
In this section, we present an illustrative example in classification to demonstrate that the dissimilarity between the marginal distributions of the synthetic and real features sometimes might lead to distinctive optimal models under the same model specification, which results in inconsistent model comparison. Here we present an illustrative example for classification to demonstrate that the dissimilarity between the marginal distributions of the synthetic and real features sometimes might lead to distinctive optimal models under the same model specification, which results in inconsistent model comparison. We denote the support of synthetic and real features as $\mathcal{X}=[-1,1]$. Let the decision rule function in classification be $\eta(x)=1$ for $x\in [0,1]$ and $-1$ otherwise. The true feature $X$ and the synthetic feature $\widetilde{X}$ are further assumed to follow the corresponding distributions,
\begin{align*}
P_{X}(x) =\begin{cases}
1-\alpha, &\mbox{ if } x\in [0,1],\\
\alpha,&\mbox{ if } x \in [-1,0).
\end{cases}
\mbox{ and }
P_{\widetilde{X}}(x) =
\begin{cases}
\alpha, &\mbox{ if } x\in [0,1],\\
1-\alpha,&\mbox{ if } x \in [-1,0).
\end{cases}
\end{align*}
We further assume that the estimation model $\widetilde{\eta}$ and $\widetilde{\eta}$ are the same as $\eta$ and $\eta$, respectively.

For the classification task, we also suppose there are two model specifications $\mathcal{F}_1= \{f(x)=\sign(|x|-\beta): \beta \in [0,1/2]\}$ and $\mathcal{F}_2= \{f(x)=\sign(|x|-\beta): \beta \in [1/4,1/3]\}$, the risks of the true and synthetic distributions are, respectively, given as
\begin{align*}
R(f) =
1-\alpha+(2\alpha-1)\beta \mbox{ and }
\widetilde{R}(f) = 
\alpha-(2\alpha-1)\beta.
\end{align*}
For any $\alpha>1/2$, we have $f_{\mathcal{F}_1}^\star(x) = \sign(|x|)$ and $f_{\mathcal{F}_2}^\star(x)=\sign(|x|-1/4)$, while $\widetilde{f}_{\mathcal{F}_1}^\star(x) = \sign(|x|-1/2)$ and $\widetilde{f}_{\mathcal{F}_2}^\star(x)=\sign(|x|-1/3)$. Clearly, the model comparison is inconsistent since $R(\widetilde{f}_{\mathcal{F}_1}^\star)<R(\widetilde{f}_{\mathcal{F}_2}^\star)$. The example provided illustrates how changes in the marginal distribution of features can influence the optimal model for minimizing risk under a given model specification. While a more flexible model specification can offer better generalization performance by capturing complex relationships between features and the target variable, it also increases the possibility of estimating an inferior model based on synthetic data with low feature fidelity.

\subsection{Consistent Model Comparison}

In this section, we examine how the quality of synthetic data influences the results of model comparisons. Our objective is to shed light on the conditions necessary for ensuring that the relative performance of optimal models derived from synthetic data aligns with those derived from real data. As seen in Figure \ref{Fig:Decomposition}, our analysis of these sufficient conditions will be grounded in the relationships between $\mathbb{P}_{Y|\bm{X}}$ and $\mathbb{P}_{\widetilde{Y}|\widetilde{\bm{X}}}$, as well as between $\mathbb{P}_{\bm{X}}$ and $\mathbb{P}_{\widetilde{\bm{X}}}$.

Although a sufficient condition for consistent model comparison can be derived using the utility bound in (\ref{UUUC}), the IPM used to measure the fidelity of the features depends on the downstream models for comparison. Therefore, a more intricate tool for analyzing feature fidelity is needed. The problem of evaluating feature fidelity bears resemblance to assessing the degree of covariate shift in transfer learning, especially when we view real and synthetic distributions as target and source distributions, respectively. Following this logic, the analysis of feature fidelity can focus on the density ratio \( r(\bm{x}) \), as explored in transfer learning literature \citep{cortes2010learning, stojanov2019low, ma2023optimally}:
\[
r(\bm{x}) = \frac{P_{\bm{X}}(\bm{x})}{P_{\widetilde{\bm{X}}}(\bm{x})}.
\]
This ratio can be utilized to characterize the distributional difference between real and synthetic feature distributions. In the literature of transfer learning, different assumptions are imposed on $r(\bm{x})$ to analyze its impact on the downstream generalization capability, such as the boundedness of $r(\bm{x})$ \citep{kpotufe2017lipschitz,simchowitz2023statistical} and the boundedness of the second moment of $r(\bm{x})$ \citep{cortes2010learning,stojanov2019low,ma2023optimally}, where the former is known as the bounded density ratio assumption and the latter equivalently indicates the boundedness of $\chi^2$-divergence between the distributions of $\bm{X}$ and $\widetilde{\bm{X}}$. Even though the boundedness of $\chi^2$-divergence between the distributions of $\bm{X}$ and $\widetilde{\bm{X}}$ is more general by allowing for the unboundedness of $r(\bm{x})$, it still fails to capture various other differences between the real and synthetic distributions. In the following, we provide a top example to validate this argument.

\begin{example}
\label{Exam:Inf}
Suppose $P_{X}(x)=2x$ and $P_{\widetilde{X}}(x)=2-2x$ for $x \in [0,1]$, then $\chi^2(\mathbb{P}_{X} \Vert \mathbb{P}_{\widetilde{X}})=\chi^2(\mathbb{P}_{\widetilde{X}}\Vert \mathbb{P}_{X} )=\infty$. %However, $\mathbb{P}_{X}$ and $\mathbb{P}_{\widetilde{X}}$ satisfy $(2,1)$-fidelity level.
\end{example}

Example \ref{Exam:Inf} demonstrates that $\chi^2$-divergence may fail to capture the difference between two distributions when their density ratio is unbounded, resulting in an infinite $\chi^2$-divergence. Therefore, in this example, $\chi^2$-divergence as a metric for assessing feature fidelity is broken when analyzing models trained on synthetic data.

Therefore, we present a novel metric called $(V,d)$-fidelity level that evaluates the differences between real and synthetic feature distributions. Our proposed metric resembles $f$-divergences \citep{renyi1961measures} in essence, as both rely on density ratios for characterizing distributional difference. The proposed $(V,d)$-fidelity level offers an alternative means of gauging the disparity between two distributions. In essence, by utilizing the $(V,d)$-fidelity level, we can carry out a more intricate analysis of how the deviation of the synthetic distribution from the real distribution influences downstream model comparison.

\begin{definition}[$(V,d)$-fidelity level]
Let $\mathcal{X}$ be the support of $\mathbb{P}_{\bm{X}}$ and $\mathbb{P}_{\widetilde{\bm{X}}}$. We say the distributions $\mathbb{P}_{\bm{X}}$ and $\mathbb{P}_{\widetilde{\bm{X}}}$ satisfy $(V,d)$-fidelity level if there exist $V \geq 0$ and $0<d \leq \infty$ such that for any $C \geq 1$, 
\begin{align}
\label{NewDis}
\max\left\{\int_{\mathcal{X}_C} P_{\bm{X}}(\bm{x})d\bm{x},
\int_{\mathcal{X}_C'} P_{\widetilde{\bm{X}}}(\bm{x})d\bm{x}
\right\} \leq V \cdot C^{-d},
\end{align}
where $\mathcal{X}_C = \left\{\bm{x} \in \mathcal{X}: r(\bm{x}) \geq C
\right\}$ and $\mathcal{X}_C' = \left\{\bm{x} \in \mathcal{X}:  r(\bm{x})\leq 1/C
\right\}$.
\end{definition}
It should be noted that the proposed metric in (\ref{NewDis}) is symmetric and utilizes two parameters $V$ and $d$ to describe the decay rates of the probabilities of areas with high density ratio values. A greater value of $d$ or a smaller value of $V$ signifies a higher degree of similarity between the two distributions. A notable attribute of $(V, d)$-fidelity levels is their capacity to accommodate scenarios where many $f$-divergence metrics are unbounded as in Example \ref{Exam:Inf}. Additionally, it is possible for a pair of distributions to meet the $(V,d)$-fidelity level for a set of values of $V$ and $d$. For instance, if $\mathbb{P}_{\bm{X}}$ and $\mathbb{P}_{\widetilde{\bm{X}}}$ conform to the $(1,1)$-fidelity level, then they must also meet $(V,d)$-fidelity level for any $V>1$.

\begin{figure}[ht]
\centering
\includegraphics[scale=0.45]{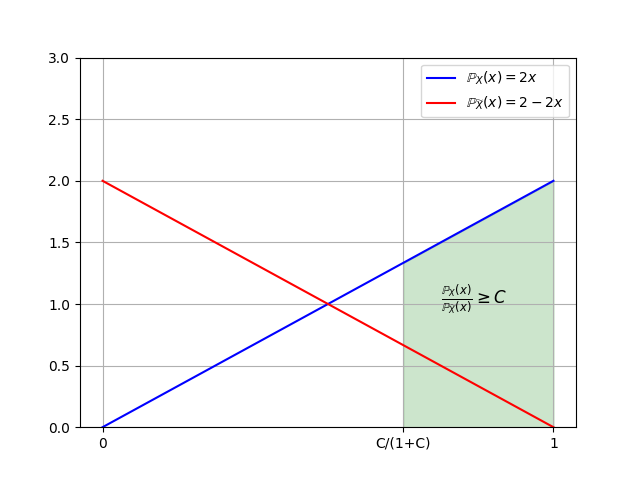}
\includegraphics[scale=0.45]{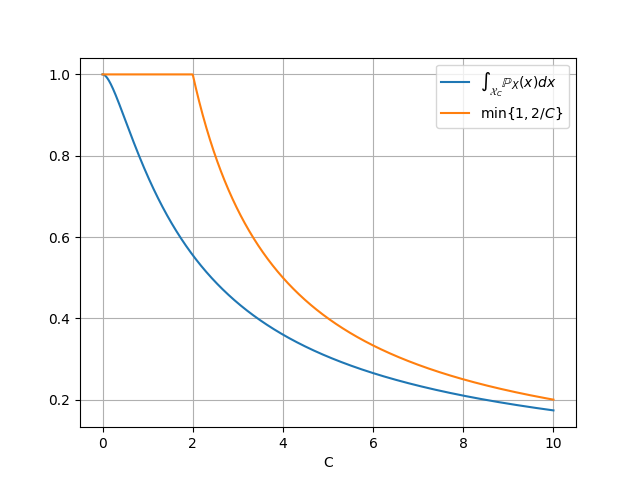}
\caption{Left: The density functions $P_{X}(x)$ and $P_{\widetilde{X}}(x)$ for $x\in [0,1]$, while the shaded area denotes the probability of $\mathcal{X}_C$. Right: the values of $\int_{\mathcal{X}_C} P_{X}(x)d x$ for $C  \geq 0$ and its decreasing pattern is captured by the upper bound function $\frac{2}{C}$ for large $C$.}
\label{fig:Inf}
\end{figure}

To study the effectiveness of $(V,d)$-fidelity level, we use it to analyze Example \ref{Exam:Inf} in Figure \ref{fig:Inf}. Clearly, the deviation between the distributions $\mathbb{P}_{X}$ and $\mathbb{P}_{\widetilde{X}}$ is characterized by the proposed metric with $(V,d)=(2,1)$. Furthermore, we specify the values of $(V,d)$ for various scenarios of real and synthetic feature distributions in Lemma \ref{Lemma:SM}.

\begin{lemma}
\label{Lemma:SM}
The following are some cases of $(V,d)$-fidelity level. 
\begin{itemize}
\item[(1)] If $\mathbb{P}_{\bm{X}}=\mathbb{P}_{\widetilde{\bm{X}}}$, then $\mathbb{P}_{\bm{X}}$ and $\mathbb{P}_{\widetilde{\bm{X}}}$ satisfy $(1,\infty)$-fidelity level.
\item[(2)] If the density ratio $M_1 \leq r(\bm{x}) \leq M_2$ for some positive constants $M_1 \in [0,1]$ and $M_2 \in [1,\infty]$, then $\mathbb{P}_{\bm{X}}$ and $\mathbb{P}_{\widetilde{\bm{X}}}$ satisfy $(\max\{M_2^d,M_1^{-d}\},d)$-fidelity level for any $d>0$.
\item[(3)] If $\chi^2(\mathbb{P}_{\bm{X}} \Vert \mathbb{P}_{\widetilde{\bm{X}}})$ and $\chi^2(\mathbb{P}_{\widetilde{\bm{X}}} \Vert \mathbb{P}_{\bm{X}})$ are finite, then $\mathbb{P}_{X}$ and $\mathbb{P}_{\widetilde{X}}$ satisfy $(M_3,1)$-fidelity level, where $M_{3} =\max\{\chi^2(\mathbb{P}_{\bm{X}} \Vert \mathbb{P}_{\widetilde{\bm{X}}}),\chi^2(\mathbb{P}_{\widetilde{\bm{X}}} \Vert \mathbb{P}_{\bm{X}})\}+1$. Conversely, if $\mathbb{P}_{\bm X}$ and $\mathbb{P}_{\widetilde{\bm X}}$ satisfy $(V,d)$-fidelity level with $d >1$, then
\begin{align*}
\max\{
\chi^2(\mathbb{P}_{\bm{X}} \Vert \mathbb{P}_{\widetilde{\bm{X}}}),
\chi^2(\mathbb{P}_{\widetilde{\bm{X}}} \Vert \mathbb{P}_{\bm{X}})
\} \leq 
(C-1)^2+\frac{V \cdot 2^d}{C^{d-1}} \cdot (2^{d-1}-1)^{-1},
\end{align*}
for any $C \geq 1$.
\item[(4)] If $X \sim \text{Exp}(K\lambda)$ and $\widetilde{X} \sim \text{Exp}(\lambda)$ with $K>1$, then $\mathbb{P}_{X}$ and $\mathbb{P}_{\widetilde{X}}$ satisfy $(1,\frac{1}{K-1})$-fidelity level.
\item[(5)] If $X \sim  N(0,\sigma^2)$ and $\widetilde{X}\sim N(\mu,\sigma^2)$, then $\mathbb{P}_{X}$ and $\mathbb{P}_{\widetilde{X}}$ satisfy $(e^{\lambda \mu/2+\sigma^2\lambda^2/2},\lambda \sigma^2 \mu^{-1}/2)$-fidelity level for any $\lambda>0$. Particularly, choosing $\lambda=\sqrt{\mu}$, $\mathbb{P}_{X}$ and $\mathbb{P}_{\widetilde{X}}$ satisfy $(1,\infty)$-fidelity level as $\mu \rightarrow 0$.
\end{itemize}
\end{lemma}
In Lemma \ref{Lemma:SM}, case (1) considers the case of perfect feature fidelity, i.e., $\mathbb{P}_{\bm{X}} = \mathbb{P}_{\widetilde{\bm{X}}}$. This case is equivalently characterized by $(V,d) = (1, \infty)$. Case (2) refers to the case of bounded density ratio, which can be characterized by various values of $(V,d)$. This is because the proposed $(V,d)$-fidelity metric mainly characterizes the similarity based on the tail probability of the values of the density ratio. Given the boundedness of the density ratio, various types of tail bounds for the density ratio may exist. Case (3) establishes the relationship between the $\chi^2$-divergence and the $(V,d)$-fidelity level. An interesting result is that if $\mathbb{P}_{\bm{X}}$ and $\mathbb{P}_{\widetilde{\bm{X}}}$ satisfy the $(V,d)$-fidelity level with $d > 1$, then their $\chi^2$-divergence must be finite. This result implies that there exist some scenarios, like Example \ref{Exam:Inf}, where the difference between real and synthetic feature distributions cannot be characterized by $\chi^2$-divergence, but can be captured for some values of $(V,d)$ with $d \leq 1$. Cases (4) and (5) specifies the values of $(V,d)$ for exponential and Gaussian distributions, respectively.

\begin{theorem}
\label{Thm:Consis}
Let $(\mathcal{F}_1,\mathcal{F}_2)$ be a pair of model classes with $R(f_{\mathcal{F}_2}^\star)<R(f_{\mathcal{F}_1}^\star)$. For a synthetic data distribution $\mathbb{P}_{\widetilde{\bm{X}}}$ satisfying $(V,d)$-fidelity level, we define
\begin{align*}
&\mathcal{W}(\mathcal{F}_1,\mathcal{F}_2)=\Phi(f_{\mathcal{F}_1}^*)-
K_{d,V}\left(
\Phi(f_{\mathcal{F}_2}^*)
\right)^{\frac{d^2}{(d+1)^2}}, \\
&
\mathcal{B}(\widetilde{\eta},\eta)=4K_{d,V}
\Vert \widetilde{\eta} - \eta \Vert_{L^2(\mathbb{P}_{\bm{X}})}^{\frac{d^2}{(d+1)^2}}+
4
\Vert \widetilde{\eta} - \eta \Vert_{L^2(\mathbb{P}_{\bm{X}})},
\end{align*}
 where $K_{d,V} = ( d^{\frac{1}{d+1}}+d^{-\frac{d}{d+1}})^{\frac{2d+1}{d+1}} 
V^{\frac{2d+1}{(d+1)^2}}$. If $\mathcal{B}(\widetilde{\eta},\eta) \leq \mathcal{W}(\mathcal{F}_1,\mathcal{F}_2)$, then we have consistent model comparison based on synthetic data, i.e., 
     $
     \Big(R(f_{\mathcal{F}_2}^\star)-R(f_{\mathcal{F}_1}^\star)\Big)\Big(R(\widetilde{f}_{\mathcal{F}_2}^\star)-R(\widetilde{f}_{\mathcal{F}_1}^\star)\Big)>0.
     $
\end{theorem}

Theorem \ref{Thm:Consis} sheds light on the sufficient conditions that $\mathbb{P}_{\widetilde{\bm{X}},\widetilde{Y}}$ must fulfill to guarantee consistent model comparison in the downstream task. In this theorem, $\mathcal{B}(\widetilde{\eta},\eta)$ characterizes the approximation error of $\widetilde{\eta}$ to $\eta$, while $\mathcal{W}(\mathcal{F}_1,\mathcal{F}_2)$ characterizes the generalization performance gap between the optimal models of $\mathcal{F}_1$ and $\mathcal{F}_2$. An intriguing observation is that achieving consistent model comparison only entails $\mathcal{B}(\widetilde{\eta},\eta)$ to be smaller than $\mathcal{W}(\mathcal{F}_1,\mathcal{F}_2)$. This result implies that even in a situation where synthetic responses are not well generated ($\Vert \widetilde{\eta}-\eta\Vert_{L^2(\mathbb{P}_{\bm{X}})}$ is large), consistent model comparison remains attainable as long as $\mathcal{W}(\mathcal{F}_1,\mathcal{F}_2)$ is large enough. Additionally, $(V,d)$-fidelity level affects the difficulty to meet the condition $\mathcal{B}(\widetilde{\eta},\eta) \leq \mathcal{W}(\mathcal{F}_1,\mathcal{F}_2)$. Specifically, the quantity $C_{d,V,U}$ converges to $1$ as $d$ approaches infinity (larger similarity between $\mathbb{P}_{\bm{X}}$ and $\mathbb{P}_{\widetilde{\bm{X}}}$) irrespective of the specific values of $V$ and $U$.

%\wei{[Wei: we need to discuss examples where the condition $\mathcal{B}(\widetilde{\eta},\eta) \leq \mathcal{W}(\mathcal{F}_1,\mathcal{F}_2)$ is satisfied, right? It is different from the condition $\mathcal{W}(\mathcal{F}_1,\mathcal{F}_2)>0$.]}

In what follows, we provide several examples explicating the condition $\mathcal{W}(\mathcal{F}_1,\mathcal{F}_2)>\mathcal{B}(\widetilde{\eta},\eta) $ under varying $(V,d)$-fidelity levels and the condition that synthetic responses are generated well ($\widetilde{\eta} = \eta$).

\begin{example}
\label{Exam:4}
If $\Phi(f_{\mathcal{F}_2}^\star)=0$ and $\Phi(f_{\mathcal{F}_1}^\star)>0$, then $\mathcal{W}(\mathcal{F}_1,\mathcal{F}_2)=\sqrt{\Phi(f_{\mathcal{F}_1}^\star)}>0$ automatically holds true.
\end{example}

\begin{example}
If $\mathbb{P}_{\bm{X}}=\mathbb{P}_{\widetilde{\bm{X}}}$, then $\mathcal{W}(\mathcal{F}_1,\mathcal{F}_2) = \sqrt{\Phi(f_{\mathcal{F}_1}^\star)}-\sqrt{\Phi(f_{\mathcal{F}_2}^\star)}>0$, where $\mathcal{W}(\mathcal{F}_1,\mathcal{F}_2)>0$ automatically holds true by the assumption that $R(f_{\mathcal{F}_2}^\star)<R(f_{\mathcal{F}_1}^\star)$.
\end{example}

\begin{theorem}
\label{Thm:Consis_2}
For any $v>0$, let $\mathcal{H}(v)=\left\{(\mathcal{F}_1,\mathcal{F}_2):\mathcal{W}(\mathcal{F}_1,\mathcal{F}_2) \geq v\right\}$ denote a class of function pairs  $(\mathcal{F}_1,\mathcal{F}_2)$. Assume that (1) $\mathbb{P}_{\widetilde{\bm{X}}}$ and $\mathbb{P}_{\bm{X}}$ satisfy $(V,d)$-fidelity level; (2) $\widetilde{\eta}$ is a consistent estimator of $\eta$ such that for some positive sequence $\{a_n\}_{n=1}^{\infty}$ tending towards infinity, for any $n \geq 1$ and any $\delta>0$,
$
\mathbb{P}\left(
\Vert
\widetilde{\eta}-\eta \Vert_{L^2(\mathbb{P}_{\bm{X}})}
>\delta
\right) \leq C_1 \exp(-C_2 a_n\delta),
$
for some positive constants $C_1$ and $C_2$. Then, for any $v>0$, 
\begin{align*}
&\sup_{(\mathcal{F}_1,\mathcal{F}_2) \in \mathcal{H}(v)}
\mathbb{P}
\left(
\big(R(f_{\mathcal{F}_2}^\star)-R(f_{\mathcal{F}_1}^\star)\big)\big(R(\widetilde{f}_{\mathcal{F}_2}^\star)-R(\widetilde{f}_{\mathcal{F}_1}^\star)\big)<0
\right) \\ \leq &  C_1\exp\left(-C_2 a_n v^{\frac{(d+1)^2}{d^2}}(4K_{d,V}+4)^{-\frac{(d+1)^2}{d^2}}\right),
\end{align*}
where the randomness results from the training dataset $\mathcal{D}$ used for estimating $\eta$.
\end{theorem}
Theorem \ref{Thm:Consis_2} shows that consistent model comparison is guaranteed with a high probability for any pair of model classes $(\mathcal{F}_1,\mathcal{F}_2)$ within $\mathcal{H}(v)$ under suitable conditions. Notably, assumption (2) in Theorem \ref{Thm:Consis_2} indicates as $n$ approaches infinity, the synthetic conditional distribution $\mathbb{P}_{\widetilde{Y}|\widetilde{\bm{X}}}$ accurately captures the relationship between features and response in $\mathbb{P}_{Y|\bm{X}}$. This assumption is mild and is generally attainable through some non-parametric regression techniques \citep{fan1992variable,fan1997local,zhao2019analysis} or neural networks \citep{nakada2020adaptive,kohler2021rate}.

\begin{corollary}
\label{Coro:Consis}
Under the assumptions of Theorem \ref{Thm:Consis_2}, if $\mathcal{F}_2$ is chosen appropriately such that $\Phi(f_{\mathcal{F}_2}^\star)=0$, then we have
\begin{align}
\label{Coro_Ineq}
&\mathbb{P}
\left(
\big(R(f_{\mathcal{F}_2}^\star)-R(f_{\mathcal{F}_1}^\star)\big)\big(R(\widetilde{f}_{\mathcal{F}_2}^\star)-R(\widetilde{f}_{\mathcal{F}_1}^\star)\big)<0
\right) \notag \\
\leq & 
C_1\exp\left(-C_2 a_n \left(\Phi(f_{\mathcal{F}_1}^\star)\right)^{\frac{(d+1)^2}{d^2}}(4K_{d,V}+4)^{-\frac{(d+1)^2}{d^2}}\right).
\end{align}
\end{corollary}
Corollary \ref{Coro:Consis} considers a specific scenario where $\mathcal{F}_2$ is the optimal function class with $\Phi(f_{\mathcal{F}_2}^\star)=0$. In this special case, consistent model comparison holds true for any $\mathcal{F}_1$ with $\Phi(f_{\mathcal{F}_1}^\star)>0$ under different $(V,d)$-fidelity levels, provided that $\Vert \widetilde{\eta}-\eta\Vert_{L^2(\mathbb{P}_{\bm{X}})}=o_p(1)$. Furthermore, the inequality in (\ref{Coro_Ineq}) aligns with the intuition that a larger value of $\Phi(f_{\mathcal{F}_1}^\star)$ corresponds to a greater probability of ensuring consistent model comparison under varying $(V,d)$-fidelity levels. This result unveils an intriguing observation: the optimal model from the correct model specification under synthetic data consistently outperforms other optimal models derived from inferior model specifications, irrespective of the distribution of synthetic features. Moreover, Corollary \ref{Coro:Consis} demonstrates that consistent model comparison can be achievable across varying degrees of dissimilarity between real and synthetic feature distributions, provided that one downstream model class exhibits the best generalization performance, i.e. $\Phi(f_{\mathcal{F}_2}^\star)=0$.

\section{Experiments}
\label{Sec:Experiments}

In this section, we present a series of simulations and a real-world application to validate our theoretical findings. In Sections \ref{Sec:CG} and \ref{Sec:CMC}, we conduct simulations to demonstrate our conclusions about consistent generalization and model comparison, respectively. In Section \ref{Sec:RealAPp}, we use the MNIST dataset as an example to illustrate that a more flexible function class in the downstream task facilitates achieving consistent generalization.

\subsection{Simulation: Consistent Generalization}
\label{Sec:CG}
Our first simulation focuses on verifying the convergence of the utility metric as described in (\ref{UtiR}) and investigating how the convergence of the utility metric is affected by various factors such as model specification, feature fidelity, and the estimation models used. Our experimental results are in line with our theoretical findings, thereby confirming the tightness and effectiveness of the analytic bounds established in (\ref{UUUC}).

\textbf{Simulation Setting:} We primarily consider the generation framework illustrated in Figure \ref{fig:FrameSyn}. Additionally, we explore practical scenarios where non-parametric methods are employed as the estimation model under various model specifications. The generation of a real dataset $\{(\bm{x}_i, y_i)\}_{i=1}^n$ for classification is structured as follows. First, we generate the real features $\bm{x}_i$ via a truncated multivariate normal distribution \citep{botev2017normal} with lower bounds $\bm{l} = (-2, -2)$, upper bounds $\bm{u} = (2, 2)$, and a mean vector $\bm{m} = (1, 1)$. The true responses are generated via $y_i = I(2\exp(x_{i1}) + x_{i1}^3 + 2\exp(x_{i2}) - x_{i2}^3 > 2)$. We employ random forests (RF; \citealp{breiman2001random}) and deep neural networks \citep{farrell2021deep} for generating synthetic responses due to their capability for approximation. For the hyperparameters of the random forest, we set the number of trees and the maximum depth of each tree to $3/2n^{1/2}$ and $\log(n)$, respectively. For the deep neural network, we use a 3-layer perceptron with 10 hidden units and ReLU activation functions. We consider the following feature transformations for a linear SVM, corresponding to three model specifications: (1) $\psi_1(\bm{x}) = (x_1, x_2)$; (2) $\psi_2(\bm{x}) = (x_1, x_2, x_1^2, x_2^2)$; (3) $\psi_3(\bm{x}) = (\exp(x_1), \exp(x_2), x_1^3, x_2^3)$. Clearly, $\psi_3$ represents the correct model specification that achieves the optimal generalization error. We generate a testing dataset of 50,000 samples to estimate the risks of the estimated models.

\begin{figure}[ht]
        \centering
                \begin{subfigure}[b]{0.46\textwidth}
            \centering
            \includegraphics[width=\textwidth]{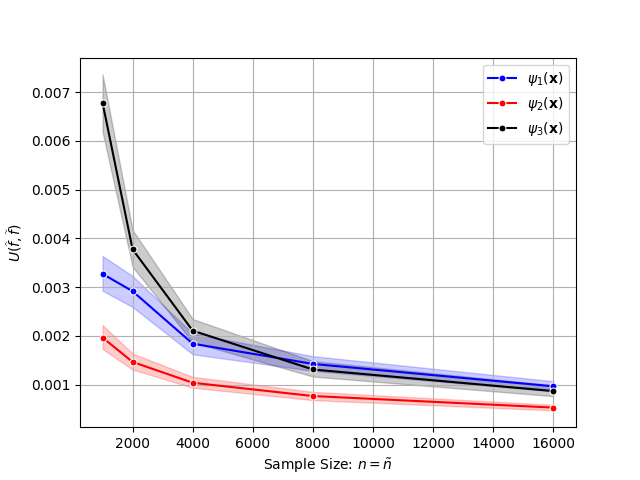}
            \caption[]%
            {$\widetilde{\eta}$: RF with perfect features} 
                        \label{Pera}   
        \end{subfigure}
        \begin{subfigure}[b]{0.46\textwidth}  
            \centering 
            \includegraphics[width=\textwidth]{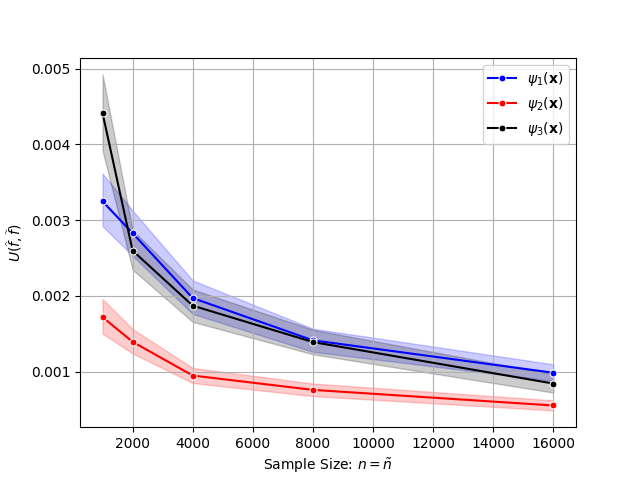}
            \caption[]%
            {$\widetilde{\eta}$: DNN with perfect features}    
        \end{subfigure}

        \begin{subfigure}[b]{0.46\textwidth}\label{123}
            \centering 
            \includegraphics[width=\textwidth]{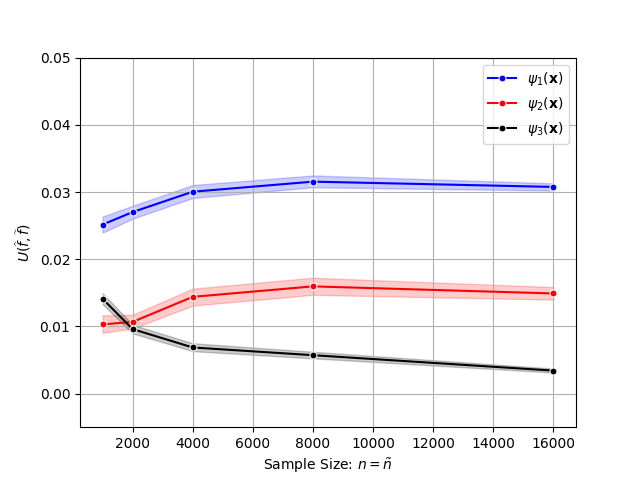}
            \caption[]%
            {$\widetilde{\eta}$: RF with imperfect features}    
        \end{subfigure}
        \begin{subfigure}[b]{0.46\textwidth}
            \centering
            \includegraphics[width=\textwidth]{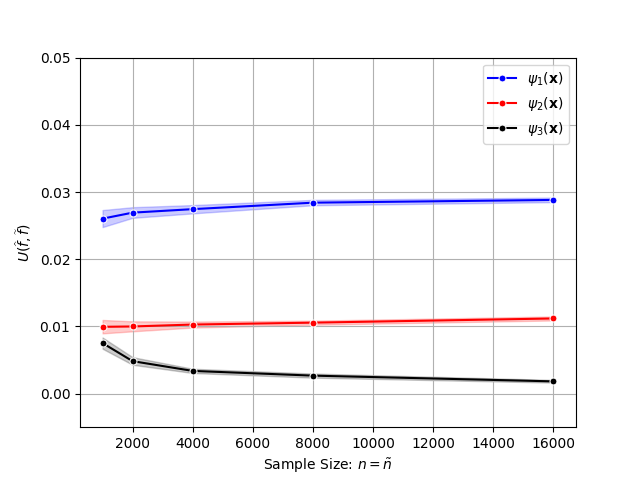}
            \caption[]%
            {$\widetilde{\eta}$: DNN with imperfect features}    
        \end{subfigure} 
        \caption{The estimated utility metrics with perfect features (a)-(b) and imperfect features (c)-(d) for Scenarios I and II, respectively.
        }
        \label{fig:Three}
    \end{figure}

To investigate the effect of feature fidelity on the utility metrics, we consider two scenarios. In \textbf{Scenario I}, the synthetic and real feature distributions are assumed to be the same, and the convergence of utility metrics is analyzed under various model assumptions and estimation methods. In \textbf{Scenario II}, the synthetic feature distribution is a uniform distribution over the same support, and the behavior of utility metrics is studied under three different model specifications. The sample size is set as $n\in \{1000 \times 2^i,i=1,2,3,4,5\}$, and each setting is replicated 100 times to evaluate the convergence of utility metrics. We reports the average estimated utility metrics and their 95\% confidence intervals for both scenarios in Figure \ref{fig:Three}.

    As can be seen in Figure \ref{fig:Three}, with perfect feature fidelity, the utility metric $U(\widehat{f},\widetilde{f})$ with different model specifications and estimation methods decrease to zero simultaneously as expected. In stark contrast, the utility metrics of wrong model specifications $\psi_1(\bm{x})$ and $\psi_2(\bm{x})$ converge to some fixed positive constants as $n$ increases and finally stay unchanged. Conversely, for the correct model specification $\psi_3(\bm{x})$, the utility metrics with different label generators converge to zero as $n$ increases. These results empirically support our conclusions about Theorem \ref{Thm:UTBmore}.

    \subsection{Simulation: Consistent Model Comparison}
    \label{Sec:CMC}
    In this part, we conduct an experiment to validate two phenomenon of model comparison based on synthetic data. Specifically, consistency of model comparison based on synthetic data is achievable when one of the models accurately specifies the regression function, or when the generalization gap between two model specifications is wide enough to offset any differences in distribution between synthetic and real features.
    
    \textbf{Simulation Setting:} We mainly consider four feature transformations for comparisons under linear SVM: (1) $\psi_1 = (x_1, x_2)$; (2) $\psi_2(\bm{x}) = (x_1, x_2, x_1^2)$; (3) $\psi_3(\bm{x}) = (x_1, x_2, x_1^3)$; (4) $\psi_4(\bm{x}) = (\exp(x_1), \exp(x_2), x_1^3, x_2^3)$. We first run these four model specifications using 50,000 real samples generated from the identical truncated normal distribution in Section \ref{Sec:CG} for training and evaluate them on 50,000 testing samples. The true performance ranking is $\psi_1 \prec \psi_3 \prec \psi_2 \prec \psi_4$, indicating that $\psi_4$ has the smallest testing error. It is also worth noting that the generalization gap between $\psi_1$ and $\psi_2$ is larger than that between $\psi_1$ and $\psi_3$. To evaluate the effect of feature fidelity, we consider a mixed dataset that contains $\alpha \times 100$ percent real samples and $(1-\alpha) \times 100$ percent synthetic samples from a uniform distribution over the same support. Here, $\alpha$ controls the discrepancy between real and synthetic feature distributions, and $\alpha=1$ corresponds to perfect feature fidelity. We consider $\alpha \in \{0.2, 0.4, 0.6, 0.8, 1\}$ and run these four model specifications on $100,000$ synthetic samples in 100 replications. We then measure their performance on the same real testing dataset and report the percentage of achieving consistent model comparison over 100 replications in Table \ref{tab:Scen_2_CMC}.

\begin{table}[h!]
\centering
\caption{The percentage of achieving consistent model comparison between different model specifications and feature fidelity over 100 replications.}
\begin{tabular}{c|c|c|c|c|c|c|c}
\toprule[2pt]
$\widetilde{\eta}$&Fidelity & $\psi_1$ vs $\psi_2$  & $\psi_1$ vs $\psi_3$  & $\psi_1$ vs $\psi_4$  & $\psi_2$ vs $\psi_3$ & $\psi_2$ vs $\psi_4$ & $\psi_3$ vs $\psi_4$ \\
\midrule
\multirow{5}{*}{RF}
&$\alpha=0.2$  & 60\%  &  32\%  & 100\%    & 84\%   &  100\%   & 100\%    \\
&$\alpha=0.4$  &  81\% &  39\%  & 100\%    &  95\%  &  100\%   & 100\%    \\
&$\alpha=0.6$  &  97\% &  72\%  & 100\%    &  100\%  &  100\%   & 100\%    \\
&$\alpha=0.8$  &  100\% &  96\%  &  100\%   &  100\%  &  100\%   & 100\%    \\
&$\alpha=1$  & 100\% &  100\%  &  100\%  & 100\%   &  100\%  &  100\%   \\
\bottomrule
\multirow{5}{*}{DNN}
&$\alpha=0.2$  & 53\%  &  26\%  & 100\%    & 88\%   &  100\%   & 100\%    \\
&$\alpha=0.4$  &  79\% &  39\%  & 100\%    &  95\%  &  100\%   & 100\%    \\
&$\alpha=0.6$  &  97\% &  71\%  & 100\%    &  100\%  &  100\%   & 100\%    \\
&$\alpha=0.8$  &  100\% &  97\%  &  100\%   &  100\%  &  100\%   & 100\%    \\
&$\alpha=1$  & 100\% &  100\%  &  100\%  & 100\%   &  100\%  &  100\%         \\
\bottomrule[2pt]
\end{tabular}
\label{tab:Scen_2_CMC}
\end{table}

It can be seen in Table \ref{tab:Scen_2_CMC} that as feature fidelity improves (i.e., as $\alpha$ increases), achieving consistent model comparison becomes more attainable when using either RF and DNN to generate synthetic labels. Ultimately, 100\% consistent model comparison is achievable when $\alpha=1$ between these model specifications. Additionally, consistent model comparison between $\psi_4$ and other model specifications is always achievable even when feature fidelity is poor ($\alpha=0.2$). This is because $\psi_4$ is the correct model specification, and hence consistent model comparison is always attainable, which aligns with our Example \ref{Exam:4}. Furthermore, at the same level of feature fidelity, consistent model comparison between $\psi_1$ and $\psi_2$ is more achievable than between $\psi_1$ and $\psi_3$. This finding supports our conclusion from Theorem \ref{Thm:Consis} that the difficulty of model comparison diminishes as the generalization gap increases. The underlying reason for this phenomenon is that the generalization gap between $\psi_1$ and $\psi_2$ is greater than that between $\psi_1$ and $\psi_3$. Particularly, when $\alpha=0.8$, 100\% consistent model comparison is achieved between $\psi_1$ and $\psi_2$, indicating that achieving consistent model comparison imposes less stringent requirements on synthetic data compared to those for consistent generalization.

\subsection{Real Application: MNIST Dataset}\label{Sec:RealAPp}

\begin{figure}[ht]
        \centering
         \begin{subfigure}[b]{0.323\textwidth}
            \centering
            \includegraphics[width=\textwidth]{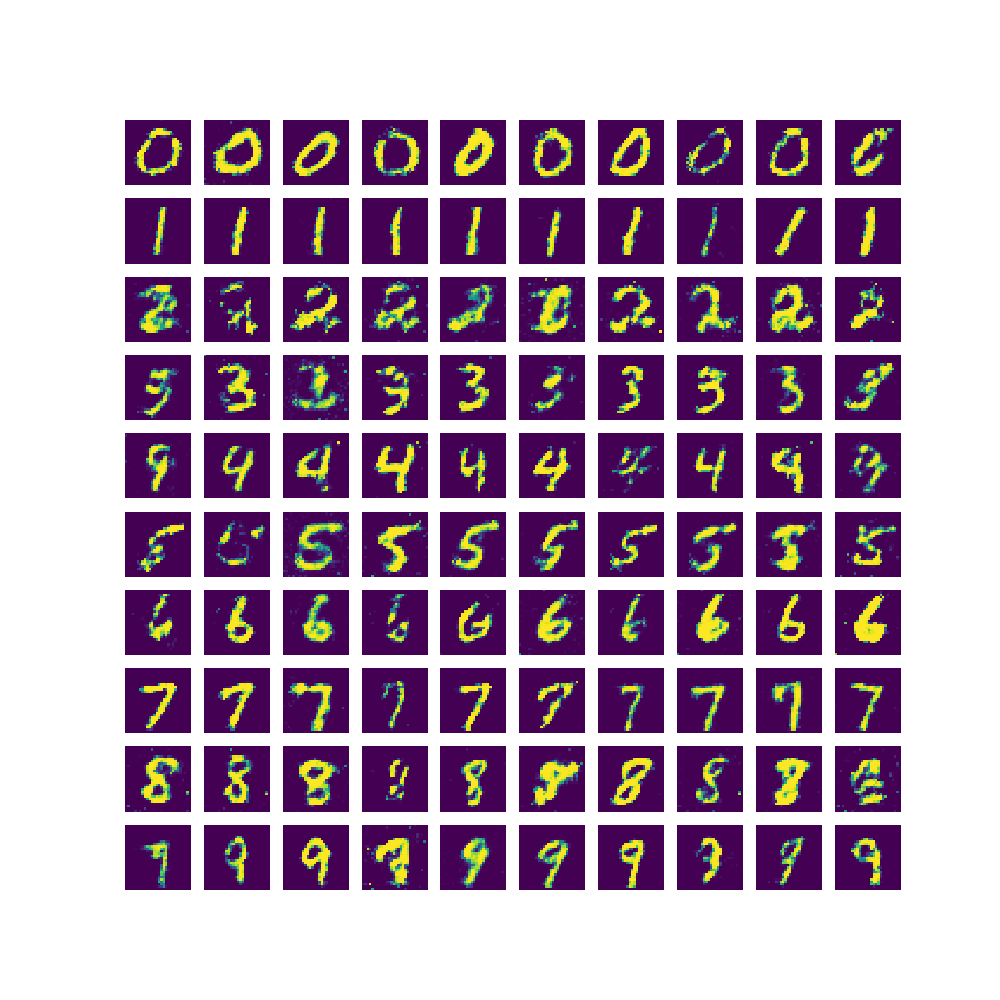}
            \caption[]%
            {30 epochs}    
        \end{subfigure}
          %       \begin{subfigure}[b]{0.24\textwidth}
       %     \centering
      %      \includegraphics[width=\textwidth]{Epoch200.png}
      %      \caption[]%
      %      {200 epochs}    
     %   \end{subfigure}
                 \begin{subfigure}[b]{0.323\textwidth}
            \centering
            \includegraphics[width=\textwidth]{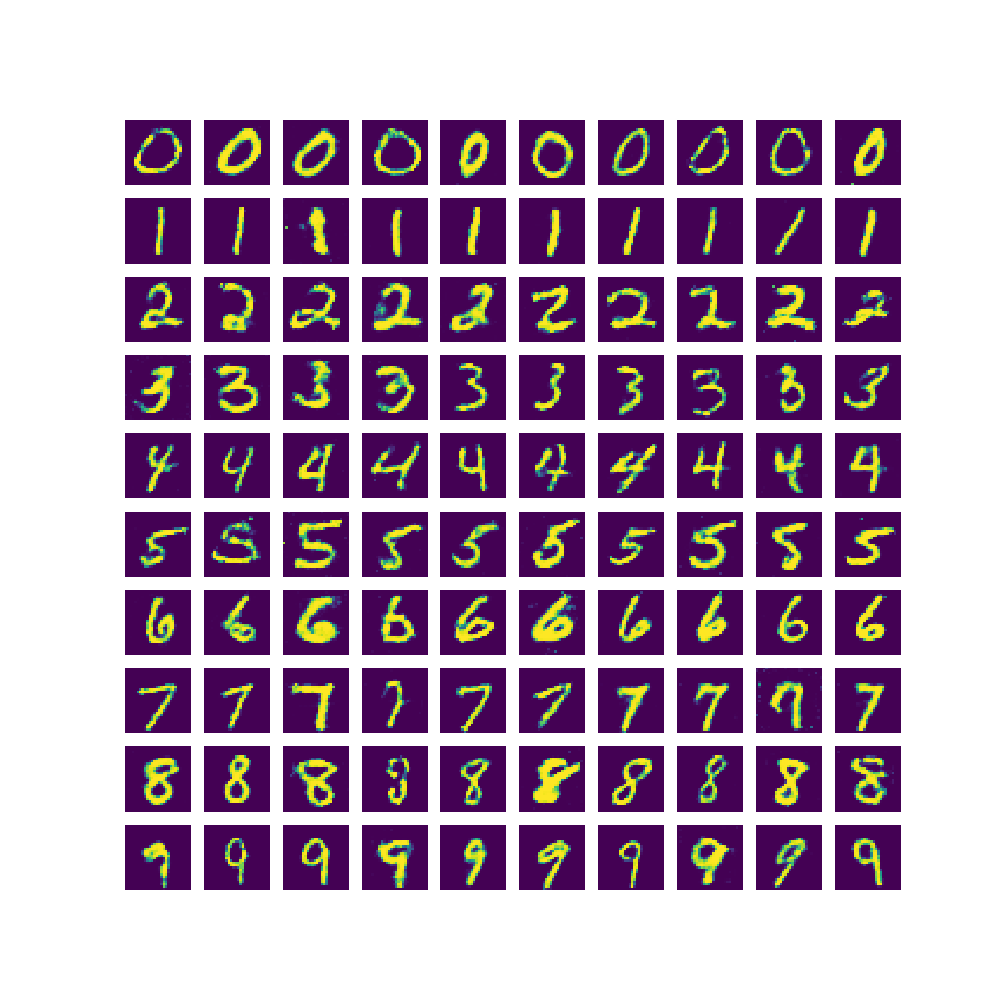}
            \caption[]%
            {90 epochs}    
        \end{subfigure}
                         \begin{subfigure}[b]{0.323\textwidth}
            \centering
            \includegraphics[width=\textwidth]{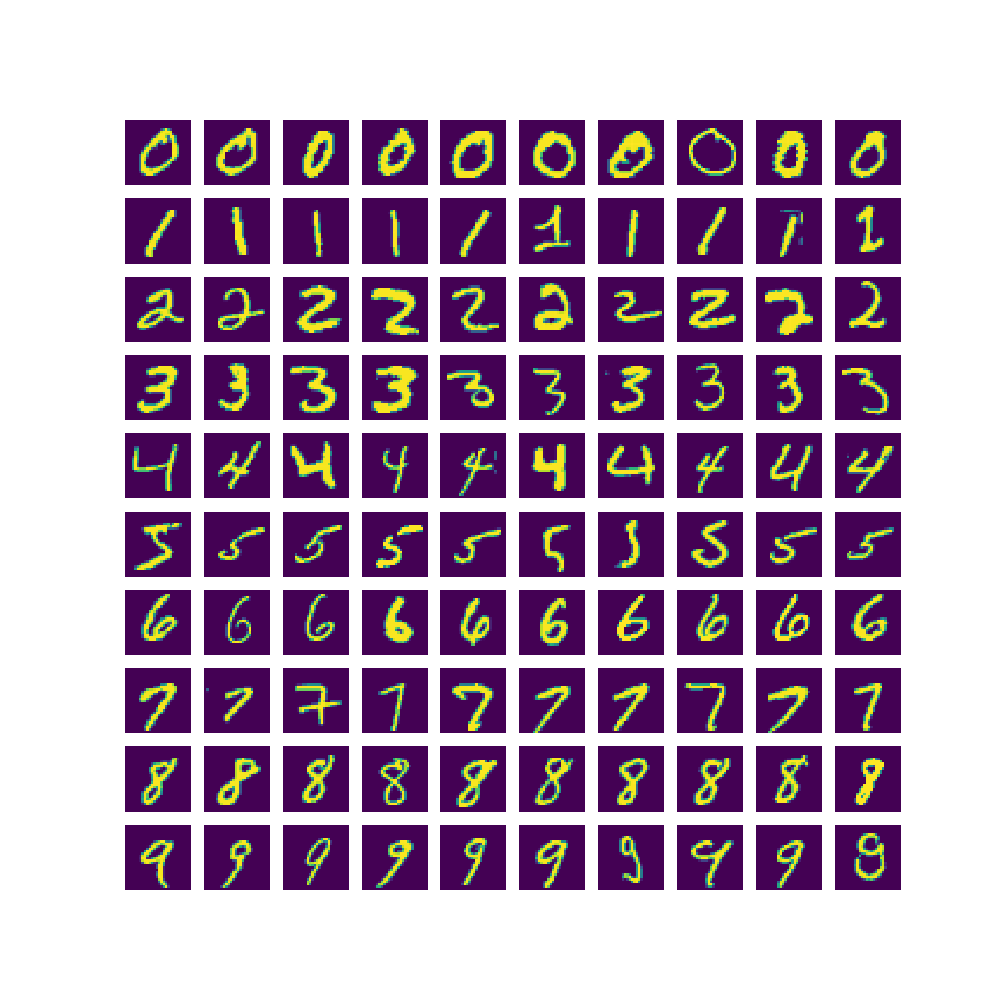}
            \caption[]%
            {Raw images}    
        \end{subfigure}
        \caption{Synthetic images generated by GANs at different epochs and real images. As the training epoch increases, synthetic images become clearer and resemble real images more.}
        \label{fig:synthe}
    \end{figure}

In this section, we use the MNIST dataset \citep{lecun1998mnist} as an example to demonstrate how models trained on synthetic data can generalize well on real data with appropriate model specification, even when the quality of the synthetic data is subpar. The MNIST dataset comprises 60,000 training images and 10,000 testing images, with each sample being a 28$\times$28 grey-scale pixel image of one of the 10 digits. We utilize a conditional generative adversarial network (GAN; \citealt{goodfellow2014generative,mirza2014conditional}) to generate synthetic images and labels. Both the generator and discriminator are 4-layer perceptrons with LeakyReLU activation functions. Each digit is assigned a 10-dimensional embedding vector, which is combined with 100-dimensional Gaussian samples to serve as input for the generator. Under the supervision of the discriminator, the generator is trained to map Gaussian samples and embedding vectors to corresponding image and label pairs. These are then reshaped into 28$\times$28 synthetic images and their corresponding synthetic labels. We employ the Adam optimizer with a learning rate of 0.0001 and use generators trained for 100 and 400 epochs to generate synthetic images representing different feature fidelity levels. As depicted in Figure \ref{fig:synthe}, as the training epoch increases, synthetic images become clearer and more closely resemble true images. Additionally, the two synthetic image datasets have the same size as the real dataset for each digit.

\begin{figure}[ht]
        \centering
         \begin{subfigure}[b]{0.31\textwidth}
            \centering
            \includegraphics[width=\textwidth]{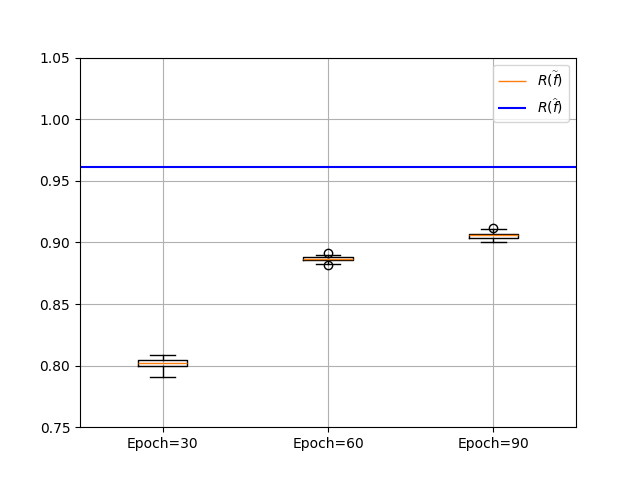}
            \caption[]%
            {Model 1}    
        \end{subfigure}
                 \begin{subfigure}[b]{0.31\textwidth}
            \centering
            \includegraphics[width=\textwidth]{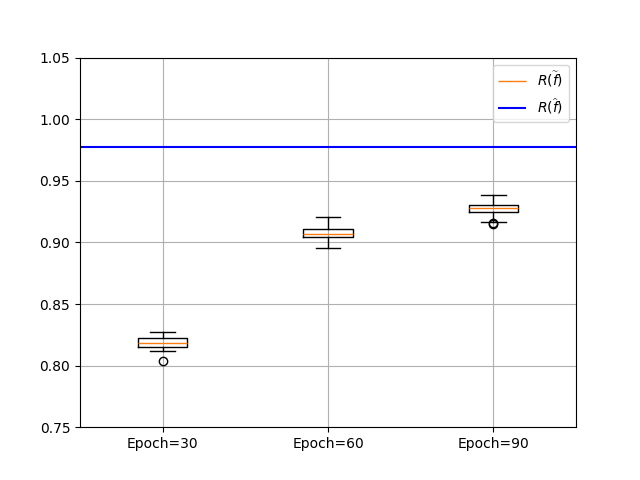}
            \caption[]%
            {Model 2}    
        \end{subfigure}
                         \begin{subfigure}[b]{0.31\textwidth}
            \centering
            \includegraphics[width=\textwidth]{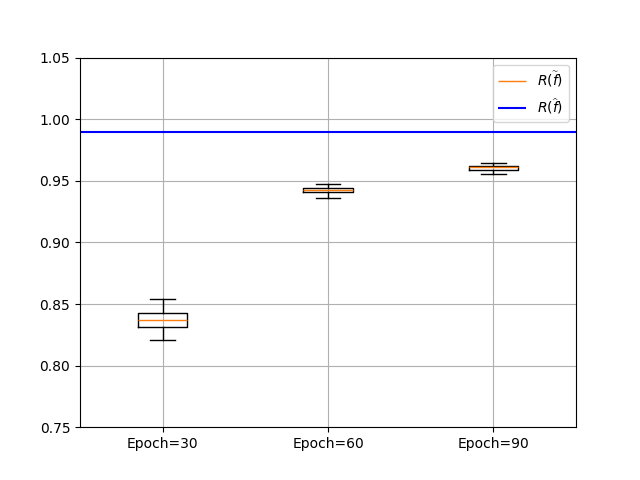}
            \caption[]%
            {Model 3}    
        \end{subfigure}
        \caption{
     The testing accuracies on 10,000 real images of different model specifications and two generators trained in 30, 60, and 90 epochs, respectively. The blue line denotes the testing accuracy of the model trained on the raw training dataset under the same model specification.
     }
        \label{fig:Mnist}
    \end{figure}

In this experiment, our aim is to emphasize the significance of model specification by considering three distinct downstream model architectures: (1) a two-layer perceptron with 50 hidden units that takes flattened images as input and outputs the probabilities for each class; (2) a neural network that integrates a convolutional layer with 8 output channels, followed by max pooling and two fully connected layers; and (3) a neural network comprising three convolutional layers with output channels of 32, 64, and 64, respectively, accompanied by max pooling and two fully connected layers. The neural networks are trained using the Adam optimizer with a learning rate of 0.0002. The three model specifications, trained on the real dataset, achieve prediction accuracies of 96.16\%, 97.71\%, and 98.97\% on the real testing dataset. Therefore, the third model specification exhibit the best generaliation performance. Then, we utilize synthetic data generated by a GAN trained for 30, 60, and 90 epochs to train these three models and evaluate these models on real testing samples. The results are presented in Figure \ref{fig:Mnist}, averaged over 30 replications.

Figure \ref{fig:Mnist} depicts the impact of increasing training epochs on the performance of all three model specifications, which demonstrates that higher quality synthetic images lead to improved model performance. Notably, compared with Model 3, Model 1 and Model 2 exhibit a significant drop in generalization accuracy on testing samples when the real dataset is substituted with synthetic data. Conversely, the drop in generalization accuracy is less severe for Model 3, which aligns with our theoretical findings that an accurate model specification is more important than the feature fidelity of synthetic data in yielding comparable performance to real data. Furthermore, consistent model comparison remains achievable even with suboptimal data quality. Notably, for synthetic data generated by a GAN trained for 30 epochs, the comparison among the three models aligns with that observed on real data. This finding supports our conclusion that consistent model comparison is a more straightforward task than consistent generalization.

\appendix
\clearpage

\setcounter{page}{1}
\setcounter{equation}{0}
\setcounter{section}{0}

\section{Proof of Lemmas}

In this Appendix, we provide the proof of Lemma \ref{Lemma:SM} and present Lemma \ref{Lemma:CompLemma} along with its proof for establishing Theorem \ref{Thm:WUB}. \\

\noindent
\textbf{Proof of Lemma \ref{Lemma:SM}.} We provide proofs for examples in Lemma \ref{Lemma:SM} sequentially.

\noindent
\textbf{Example 1.} In the first example, $P_{\bm{X}}(\bm{x})=P_{\widetilde{\bm{X}}}(\bm{x})$ for any $\bm{x} \in \mathcal{X}$. It follows that
\begin{align*}
\int_{\mathcal{X}_C'} P_{\widetilde{\bm{X}}}(\bm{x})d\bm{x}=
\int_{\mathcal{X}_C} P_{\bm{X}}(\bm{x})d\bm{x} = \begin{cases}
1, C \in [0,1], \\
0, C \in (1,+\infty).
\end{cases}
\end{align*}
The desired result immediately follows by setting $V=1$ and $d=\infty$.

\noindent
\textbf{Example 2.}  For the second example, the density ratio $P_{\bm{X}}(\bm{x})/P_{\widetilde{\bm{X}}}(\bm{x})$ is upper bounded by $M_2$ and lower bounded by $M_1$. Therefore, we can verify that 
\begin{align*}
\int_{\mathcal{X}_C} P_{\bm{X}}(\bm{x})d\bm{x} = \begin{cases}
1, C \in [0,M_1], \\
\text{unknown}, C \in (M_1,M_2), \\
0, C \in [M_2,+\infty).
\end{cases}
\end{align*}
Note that the value of $\int_{\mathcal{X}_C} P_{\bm{X}}(\bm{x})d\bm{x}$ is unknown on the interval $(M_1,M_2)$, depending on the explicit distributions of $\bm{X}$ and $\widetilde{\bm{X}}$. Therefore, we consider the worst case that $\int_{\mathcal{X}_C} P_{\bm{X}}(\bm{x})d\bm{x} \equiv 1$ for $C \in (M_1,M_2)$. Then by setting $V=M_2^d$, we can verify that 
\begin{align*}
\int_{\mathcal{X}_C} P_{\bm{X}}(\bm{x})d\bm{x} \leq M_2^d \cdot C^{-d}, \text{ for any } C >0. 
\end{align*} 
Applying a similar argument to $M_2^{-1} \leq P_{\widetilde{\bm{X}}}(\bm{x})/P_{\bm{X}}(\bm{x}) \leq M_1^{-1}$ yields that
\begin{align*}
\int_{\mathcal{X}_C'} P_{\widetilde{\bm{X}}}(\bm{x})d\bm{x} \leq (M_1C)^{-d}, \text{ for any } C >0. 
\end{align*}
To sum up, we get
\begin{align*}
\max
\left\{
\int_{\mathcal{X}_C} P_{\bm{X}}(\bm{x})d\bm{x} ,
\int_{\mathcal{X}_C'} P_{\widetilde{\bm{X}}}(\bm{x})d\bm{x}
\right\} \leq \max\{M_2^d ,M_1^{-d}\} \cdot C^{-d}, \text{ for any } C >0. 
\end{align*}
This completes the proof for the second example.

\noindent
\textbf{Example 3.} For the third example, we denote that $M_4 =\chi^2(\mathbb{P}_{\bm{X}} \Vert \mathbb{P}_{\widetilde{\bm{X}}})$ and $M_5 = \chi^2(\mathbb{P}_{\widetilde{\bm{X}}} \Vert \mathbb{P}_{\bm{X}})$. By the definition of chi-square divergence, we have
\begin{align*}
\chi^2(\mathbb{P}_{\bm{X}} \Vert \mathbb{P}_{\widetilde{\bm{X}}}) = 
\int_{\mathcal{X}}
P_{\widetilde{\bm{X}}}(\bm{x})
\frac{P_{\bm{X}}^2(\bm{x})}{P^2_{\widetilde{\bm{X}}}(\bm{x})}d\bm{x} -1
=
\int_{\mathcal{X}}
P_{\bm{X}}(\bm{x})
\frac{P_{\bm{X}}(\bm{x})}{P_{\widetilde{\bm{X}}}(\bm{x})}d\bm{x} -1=M_4.
\end{align*}
For the set $\mathcal{X}_C = \left\{\bm{x} \in \mathcal{X}: P_{\bm{X}}(\bm{x})/P_{\widetilde{\bm{X}}}(\bm{x}) \geq C
\right\}$, 
\begin{align}
\label{Ineq:M3}
\int_{\mathcal{X}_C} P_{\bm{X}}(\bm{x})d\bm{x} \leq \frac{1}{C}
\int_{\mathcal{X}_C} P_{\bm{X}}(\bm{x}) \frac{P_{\bm{X}}(\bm{x})}{P_{\widetilde{\bm{X}}}(\bm{x})}d\bm{x} \leq \frac{M_4+1}{C}.
\end{align}
Applying a similar argument to $\int_{\mathcal{X}_C'} \mathbb{P}_{\widetilde{\bm{X}}}(\bm{x})d\bm{x}$ yields that
\begin{align}
\label{Ineq:M4}
\int_{\mathcal{X}_C'} P_{\widetilde{\bm{X}}}(\bm{x})d\bm{x} =
\int_{\mathcal{X}_C'} P_{\bm{X}}(\bm{x})\frac{P_{\widetilde{\bm{X}}}(\bm{x})}{P_{\bm{X}}(\bm{x})}d\bm{x} \leq \frac{M_5+1}{C}.
\end{align}
Combine (\ref{Ineq:M3}) and (\ref{Ineq:M4}) together yields that
\begin{align*}
\max\left\{
\int_{\mathcal{X}_C} P_{\bm{X}}(\bm{x})d\bm{x},
\int_{\mathcal{X}_C'} P_{\widetilde{\bm{X}}}(\bm{x})d\bm{x}
\right\} \leq \frac{\max\{M_4,M_5\}+1}{C}.
\end{align*}
Therefore, $ \mathbb{P}_{\bm{X}}$ and $\mathbb{P}_{\widetilde{\bm{X}}}$ satisfy $(\max\{M_3,M_4\}+1,1)$-fidelity level.

Next, we suppose that $ \mathbb{P}_{\bm{X}}$ and $\mathbb{P}_{\widetilde{\bm{X}}}$ satisfy $(V,d)$-fidelity level, then the $\chi^2$-divergence can be bounded as
\begin{align*}
\chi^2(\mathbb{P}_{\bm{X}} \Vert \mathbb{P}_{\widetilde{\bm{X}}}) =&
\int_{\mathcal{X}}
P_{\bm{X}}(\bm{x})
\frac{P_{\widetilde{\bm{X}}}(\bm{x})}{P_{\bm{X}}(\bm{x})}
\left(
\frac{P_{\bm{X}}(\bm{x})}{P_{\widetilde{\bm{X}}}(\bm{x})}-1
\right)^2d\bm{x} 
 \\
 =& 
\sum_{i=0}^{\infty}
\int_{\mathcal{X}^{(i)}}
P_{\bm{X}}(\bm{x})
\frac{P_{\widetilde{\bm{X}}}(\bm{x})}{P_{\bm{X}}(\bm{x})}
\left(
\frac{P_{\bm{X}}(\bm{x})}{P_{\widetilde{\bm{X}}}(\bm{x})}-1
\right)^2d\bm{x},
\end{align*}
where $\mathcal{X}^{(0)} = \{\bm{x}\in \mathcal{X}:P_{\bm{X}}(\bm{x})/P_{\widetilde{\bm{X}}}(\bm{x}) \leq C \}$ and $\mathcal{X}^{(i)} = \{\bm{x}\in \mathcal{X}:2^{i-1}C \leq P_{\bm{X}}(\bm{x})/P_{\widetilde{\bm{X}}}(\bm{x})\leq 2^i C \}$ for $i \geq 1$ with $C>1$. Notice that $f(x)=(x-1)^2/x$ is an increasing function for $x \geq 1$, it follows that for each $i \geq 1$
\begin{align*}
\int_{\mathcal{X}^{(i)}}
P_{\bm{X}}(\bm{x})
\frac{P_{\widetilde{\bm{X}}}(\bm{x})}{P_{\bm{X}}(\bm{x})}
\left(
\frac{P_{\bm{X}}(\bm{x})}{P_{\widetilde{\bm{X}}}(\bm{x})}-1
\right)^2d\bm{x} \leq 
\int_{\mathcal{X}^{(i)}}
P_{\bm{X}}(\bm{x}) \frac{(2^i C-1)^2}{2^iC} d\bm{x}
\leq \frac{V \cdot 2^d}{2^{(d-1)i}C^{d-1}}.
\end{align*}
Given that $d >1$, it holds that
\begin{align}
\label{Exam3_1}
\sum_{i=1}^{\infty}
\int_{\mathcal{X}^{(i)}}
P_{\bm{X}}(\bm{x})
\frac{P_{\widetilde{\bm{X}}}(\bm{x})}{P_{\bm{X}}(\bm{x})}
\left(
\frac{P_{\bm{X}}(\bm{x})}{P_{\widetilde{\bm{X}}}(\bm{x})}-1
\right)^2d\bm{x} \leq 
\sum_{i=1}^{\infty}
\frac{V \cdot 2^d}{2^{(d-1)i}C^{d-1}} = 
\frac{V \cdot 2^d}{C^{d-1}} \cdot (2^{d-1}-1)^{-1}.
\end{align}
For $\mathcal{X}^{0}$, we have
\begin{align}
\label{Exam3_2}
\int_{\mathcal{X}^{(0)}}
P_{\widetilde{\bm{X}}}(\bm{x})
\left(
\frac{P_{\bm{X}}(\bm{x})}{P_{\widetilde{\bm{X}}}(\bm{x})}-1
\right)^2d\bm{x} \leq (C-1)^2.
\end{align}
Combining (\ref{Exam3_1}) and (\ref{Exam3_2}) yields that
\begin{align*}
\chi^2(\mathbb{P}_{\bm{X}} \Vert \mathbb{P}_{\widetilde{\bm{X}}})  \leq 
(C-1)^2+\frac{V \cdot 2^d}{C^{d-1}} \cdot (2^{d-1}-1)^{-1}.
\end{align*}
As $(V,d)$-fidelity level is symmetric for $\mathbb{P}_{\bm{X}}$ and $\mathbb{P}_{\widetilde{\bm{X}}}$, we also have
\begin{align*}
\chi^2(\mathbb{P}_{\widetilde{\bm{X}}} \Vert\mathbb{P}_{\bm{X}} ) \leq 
(C-1)^2+\frac{V \cdot 2^d}{C^{d-1}} \cdot (2^{d-1}-1)^{-1}.
\end{align*}
The desired result immediately follows by combining these two upper bounds together, and this completes the proof.

\noindent
\textbf{Example 4.} If $X \sim \text{Exp}(K\lambda)$ and $\widetilde{X} \sim \text{Exp}(\lambda)$, their density ratio is given as
\begin{align*}
   \rho(x) =\frac{P_X(x)}{P_{\widetilde{X}}(x)}= \frac{K e^{-K\lambda x}}{ e^{-\lambda x}} = Ke^{-(K-1)\lambda x}.
\end{align*}
Then $\rho(x) \geq C$ indicates $x
    \leq \frac{-1}{(K-1)\lambda}\log(C/K)$. Then the corresponding probability can be computed as
    \begin{align*}
       \int_{0}^{\max\left\{\frac{-1}{(K-1)\lambda}\log(C/K),0\right\}}
        K\lambda e^{-K\lambda x}dx=&
        \max\left\{
        1 - \exp\left(\frac{K}{K-1}\log(C/K)\right),0\right\}
         \\
         =&
        \max\left\{
        1 -\left(\frac{C}{K}\right)^{\frac{K}{K-1}}
        ,0\right\}
        .
    \end{align*}
    Next, we consider the case of $\rho(x) \leq C^{-1}$, which indicates $x \geq \frac{1}{(K-1)\lambda} \log(KC)$. The associated probability is given as
    \begin{align*}
        \int_{\frac{1}{(K-1)\lambda} \log(KC)}^{+\infty}
        \lambda e^{-\lambda x}dx
        =K^{-\frac{1}{K-1}} \left(\frac{1}{C}\right)^{\frac{1}{K-1}}.
    \end{align*}
Next, we turn to find values of $(V,d)$ such that
\begin{align}
\label{Ineq}
     \max\left\{
        1 -\left(\frac{C}{K}\right)^{\frac{K}{K-1}}
        ,K^{-\frac{1}{K-1}} \left(\frac{1}{C}\right)^{\frac{1}{K-1}}\right\} \leq V \cdot C^{-d},
\end{align}
for $C \geq 1$. Here we can choose $d = \frac{1}{K-1}$ and $V=1$, and (\ref{Ineq}) holds true for any $C \geq 1$. This completes the proof of the last example.

\noindent
\textbf{Example 5.} In the last example, the density ratio is given as
\begin{align*}
\frac{P_{\bm{X}}(x)}{P_{\widetilde{\bm X}}(x)}
=
\frac{\exp(-(x-\mu)^2/\sigma^2)}{\exp(-x^2/\sigma^2)}=
\exp(2\sigma^{-2}\mu x-\mu^2 \sigma^{-2}).
\end{align*}
Then $P_{\bm{X}}(x)/P_{\widetilde{\bm X}}(x) \geq C$ leads to $x \geq \frac{\sigma^2\log C+\mu^2}{2\mu}$. With this, for any positive $\lambda$, we have
\begin{align*}
&\int_{\frac{\sigma^2\log C+\mu^2}{2\mu}}^{\infty}
\frac{1}{\sqrt{2\pi}\sigma}
\exp\Big(-\frac{(x-\mu)^2}{2\sigma^2} \Big)dx =
\int_{\frac{\sigma^2\log C-\mu^2}{2\mu}}^{\infty}
\frac{1}{\sqrt{2\pi}\sigma}
\exp\Big(-\frac{x^2}{2\sigma^2} \Big)dx \\
\leq &
\exp(\lambda \mu/2) C^{-\frac{\lambda \sigma^2}{2\mu}}
\int_{\frac{\sigma^2\log C-\mu^2}{2\mu}}^{\infty}
\frac{1}{\sqrt{2\pi}\sigma}
\exp\Big(-\frac{x^2}{2\sigma^2} +\lambda x\Big)dx 
\leq 
\exp\Big(\frac{\lambda \mu+\sigma^2\lambda^2}{2}\Big) C^{-\frac{\lambda \sigma^2}{2\mu}}.
\end{align*}
Notice that $P_{\bm{X}}(x)$ and $P_{\widetilde{\bm{X}}}(x)$ are symmetric at $x=\mu/2$, therefore
\begin{align*}
\int^{\frac{\mu^2-\sigma^2\log C}{2\mu}}_{-\infty}
\frac{1}{\sqrt{2\pi}\sigma}
\exp\Big(-\frac{x^2}{2\sigma^2} \Big)dx \leq \exp\Big(\frac{\lambda \mu+\sigma^2\lambda^2}{2}\Big) C^{-\frac{\lambda \sigma^2}{2\mu}}.
\end{align*}
Finally, we have $P_{\bm X}(x)$ and $P_{\widetilde{\bm X}}(x)$ satisfy $(\exp(\lambda \mu/2+\sigma^2\lambda^2/2),\lambda \sigma^2 \mu^{-1}/2)$-fidelity level for any $\lambda>0$. Taking $\lambda=\sqrt{\mu}$, we have
\begin{align*}
\lim_{\mu\rightarrow 0} \exp(\mu^{3/2}/2+\sigma^2\mu/2)=1\mbox{ and }
\lim_{\mu\rightarrow0} \sigma^2 \mu^{-1/2}/2 =\infty.
\end{align*}
This completes of proof of Lemma \ref{Lemma:SM}. \qed \\

\begin{lemma}
\label{Lemma:CompLemma}
    For any $\mathbb{P}_{\bm{X},Y}$ and $\mathbb{P}_{\widetilde{\bm{X}},\widetilde{Y}}$ satisfying Assumption \ref{Ass:Low_noise}, it holds true that 
    \begin{align*}
\mathbb{E}
\left\{
I \left(
f^{\star}(\bm{X}) \neq \widetilde{f}^{\star}(\bm{X}) 
\right)
\left|\widetilde{\eta}(\bm{X})-\eta(\bm{X})\right|
\right\} \lesssim     \Vert \eta - \widetilde{\eta} \Vert_{L^2(\mathbb{P}_{\bm{X}})}^{\frac{2\gamma^\prime+2}{\gamma^\prime+2}},
    \end{align*}
    where $\gamma^\prime=\max\{\gamma,\widetilde{\gamma}\}$ with $\gamma$ and $\widetilde{\gamma}$ being defined in Assumption \ref{Ass:Low_noise}.
   
\end{lemma}

\noindent
\textbf{Proof of Lemma \ref{Lemma:CompLemma}}. We first consider the following decomposition
\begin{align*}
&\mathbb{E}
\left\{
I \left(
f^{\star}(\bm{X}) \neq \widetilde{f}^{\star}(\bm{X}) 
\right)
\left|\widetilde{\eta}(\bm{X})-\eta(\bm{X})\right|
\right\} \\
=&\mathbb{E}
\left\{
I \left(
f^{\star}(\bm{X}) \neq \widetilde{f}^{\star}(\bm{X}) 
\right)
\left|\widetilde{\eta}(\bm{X})-\eta(\bm{X})\right|
I(|\eta(\bm{X})-1/2| \leq t)
\right\} \\
+&\mathbb{E}
\left\{
I \left(
f^{\star}(\bm{X}) \neq \widetilde{f}^{\star}(\bm{X}) 
\right)
\left|\widetilde{\eta}(\bm{X})-\eta(\bm{X})\right|
I(|\eta(\bm{X})-1/2| \geq t)
\right\}\triangleq I_1+I_2,
\end{align*}
for any positive constant $t>0$.
Next, we turn to bound $I_1$ and $I_2$ separately. Following from the fact that $|\eta(\bm{x})-1/2|\leq |\eta(\bm{x})-\widetilde{\eta}(\bm{x})|$ when $\widetilde{f}(\bm{x}) \neq f^\star(\bm{x})$, we have
\begin{align}
\label{Bound_For_I_1}
    I_1 = &
    2\mathbb{E}
    \Big[
    I(\widetilde{f}^\star(\bm{X})\neq f^\star(\bm{X}))|\eta(\bm{X})-\widetilde{\eta}(\bm{X})|
    I(|\eta(\bm{X})-1/2| \leq t)
    \Big]\notag \\
    \leq & 2
    \sqrt{
    \mathbb{E}
    \Big[
    (\eta(\bm{X})-\widetilde{\eta}(\bm{X}))^2
    \Big]} \cdot \sqrt{\mathbb{P}(|\eta(\bm{X})-1/2| \leq t)}\notag \\
    \leq &2\Vert \eta - \widetilde{\eta}\Vert_{L^2(\mathbb{P}_{\bm{X}})}C_0^{1/2} t^{\gamma/2},
\end{align}
where the first inequality follows from the fact that $|\eta(\bm{X})-1/2| \leq |\eta(\bm{X})-\widetilde{\eta}(\bm{X})|$ if $\widetilde{f}^\star(\bm{X})\neq f^\star(\bm{X})$ and the last inequality follows from the Cauchy–Schwarz inequality.

Next, $I_2$ can be bounded as
\begin{align}
\label{Bound_For_I_2}
    I_2 =  & 2\mathbb{E}
    \Big[
    I(\widetilde{f}^\star(\bm{X})\neq f^\star(\bm{X}))
    |\eta(\bm{X})-\widetilde{\eta}(\bm{X})|
I(|\eta(\bm{X})-1/2| > t)
    \Big] \notag \\
    \leq &
    2\mathbb{E}
    \Big[
    \left(\eta(\bm{X})-\widetilde{\eta}(\bm{X})\right)^2
    \Big] t^{-1} = 2t^{-1} \Vert \eta - \widetilde{\eta} \Vert_{L^2(\mathbb{P}_{\bm{X}})}^2.
\end{align}
Combining (\ref{Bound_For_I_1}) and (\ref{Bound_For_I_2}) yields 
\begin{align*}
\mathbb{E}
\left\{
I \left(
f^{\star}(\bm{X}) \neq \widetilde{f}^{\star}(\bm{X}) 
\right)
\left|\widetilde{\eta}(\bm{X})-\eta(\bm{X})\right|
\right\} \leq &
    2\Vert \eta - \widetilde{\eta} \Vert_{L^2(\mathbb{P}_{\bm{X}})}C_0^{1/2} t^{\gamma/2}+
    2t^{-1} \Vert \eta - \widetilde{\eta} \Vert_{L^2(\mathbb{P}_{\bm{X}})}^2.
\end{align*}
Setting $t = C_0^{-\frac{1}{\gamma+2}}\Vert \eta - \widetilde{\eta} \Vert_{L^2(\mathbb{P}_{\bm{X}})}^{\frac{2}{\gamma+2}}$ yields that
\begin{align*}
\mathbb{E}
\left\{
I \left(
f^{\star}(\bm{X}) \neq \widetilde{f}^{\star}(\bm{X}) 
\right)
\left|\widetilde{\eta}(\bm{X})-\eta(\bm{X})\right|
\right\}\leq 
    4 C_0^{\frac{1}{\gamma+2}}\left(\Vert \eta - \widetilde{\eta} \Vert_{L^2(\mathbb{P}_{\bm{X}})}^2\right)^{\frac{\gamma+1}{\gamma+2}}.
\end{align*}

It is worth noting that $\mathbb{E}
\left\{
I \left(
f^{\star}(\bm{X}) \neq \widetilde{f}^{\star}(\bm{X}) 
\right)
\left|\widetilde{\eta}(\bm{X})-\eta(\bm{X})\right|
\right\}$ is symmetric about $\widetilde{\eta}$ and $\eta$. Applying a similar argument yields
\begin{align*}
    \mathbb{E}
\left\{
I \left(
f^{\star}(\bm{X}) \neq \widetilde{f}^{\star}(\bm{X}) 
\right)
\left|\widetilde{\eta}(\bm{X})-\eta(\bm{X})\right|
\right\}\leq 
    4\widetilde{C}_0^{\frac{1}{\widetilde{\gamma}+2}}\left(\Vert \eta - \widetilde{\eta} \Vert_{L^2(\mathbb{P}_{\bm{X}})}^2\right)^{\frac{\widetilde{\gamma}+1}{\widetilde{\gamma}+2}}.
\end{align*}
To sum up, it follows that
\begin{align*}
 &   \mathbb{E}
\left\{
I \left(
f^{\star}(\bm{X}) \neq \widetilde{f}^{\star}(\bm{X})
\right)
\left|\widetilde{\eta}(\bm{X})-\eta(\bm{X})\right|
\right\} \\
\leq & \max\left\{
4 C_0^{\frac{1}{\gamma+2}}\left(\Vert \eta - \widetilde{\eta} \Vert_{L^2(\mathbb{P}_{\bm{X}})}^2\right)^{\frac{\gamma+1}{\gamma+2}},
4\widetilde{C}_0^{\frac{1}{\widetilde{\gamma}+2}}\left(\Vert \eta - \widetilde{\eta} \Vert_{L^2(\mathbb{P}_{\bm{X}})}^2\right)^{\frac{\widetilde{\gamma}+1}{\widetilde{\gamma}+2}}
\right\} \\
\lesssim & \left(\Vert \eta - \widetilde{\eta} \Vert_{L^2(\mathbb{P}_{\bm{X}})}^2\right)^{\frac{\gamma^\prime+1}{\gamma^\prime+2}},
\end{align*}
where $\gamma^\prime=\max\{\gamma,\widetilde{\gamma}\}$ and this completes the proof. \qed \\

\section{Proof of Theorems}
In this Appendix, we provide proofs for Theorems \ref{UboundC}-\ref{Thm:Consis_2}. \\

\noindent
\textbf{Proof of Theorem \ref{UboundC}.} We first note that
\begin{align*}
R(f) =& \int_{\mathcal{X}} \Big[ I\big(f(\bm{x})<0\big)\eta(\bm{x})+
I\big(f(\bm{x})>0\big)\big(1-\eta(\bm{x})\big) \Big] P_{\bm{X}}(\bm{x})d\bm{x} \\
 =& 
\int_{\mathcal{X}} |2\eta(\bm{x})-1|I\big(f(\bm{x})(\eta(\bm{x})-1/2)<0\big)
P_{\bm{X}}(\bm{x})d\bm{x} 
+\int_{\mathcal{X}}\min\{ \eta(\bm{x}),1-\eta(\bm{x})\}
P_{\bm{X}}(\bm{x})d\bm{x} \\
=&\Phi(f) + \mathbb{E}_{\bm{X}}\big[
\min\{ \eta(\bm{X}),1-\eta(\bm{X})\}
\big] = \Phi(f) +R^\star,
\end{align*}
where $R^\star$ is the Bayes risk. Therefore, $\big| R(\widehat{f})-R(\widetilde{f}) \big|$ can be further bounded as
\begin{align*}
&\left|
    \big| R(\widehat{f})-R(\widetilde{f}) \big|
    -
    \big| R(\f)-R(\tf) \big|\right|
   \\
   \leq & 
    \big|  R(\widehat{f})-R(\f) \big|+
    \big|  R(\widetilde{f})-R(\tf) \big| \\
    = &
    \big|  \Phi(\widehat{f})-\Phi(\f) \big|+
    \big|  \Phi(\widetilde{f})-\Phi(\tf) \big|.
\end{align*}
Next, we note that
\begin{align*}
   \big|  \Phi(\widehat{f})-\Phi(\f) \big| =&
   \left|
   \int_{\mathcal{X}} |2\eta(\bm{x})-1|I\big(\widehat{f}(\bm{x})(\eta(\bm{x})-1/2)<0\big)
P_{\bm{X}}(\bm{x})d\bm{x} \right.\\
&\left.-
\int_{\mathcal{X}} |2\eta(\bm{x})-1|I\big(\f(\bm{x})(\eta(\bm{x})-1/2)<0\big)
P_{\bm{X}}(\bm{x})d\bm{x} \right| \\
\leq &
\int_{\mathcal{X}} |2\eta(\bm{x})-1|I\big(\f(\bm{x}) \neq \widehat{f}(\bm{x})\big)
P_{\bm{X}}(\bm{x})d\bm{x}  \\
\leq & \mathbb{E}_{\bm{X}}\left[I(\widehat{f}(\bm{X}) \neq \f(\bm{X}))\right].
\end{align*}
Applying a similar argument to $\big|  \Phi(\widetilde{f})-\Phi(\tf) \big|$, we have
\begin{align*}
\big|  \Phi(\widetilde{f})-\Phi(\tf) \big|\leq    
\mathbb{E}_{\bm{X}}\left[I(\widetilde{f}(\bm{X}) \neq \tf(\bm{X}))\right].
\end{align*}
Therefore, we have
\begin{align*}
    \left|
    U(\widehat{f},\widetilde{f})
    -
    U(\tf,\f)\right| \leq 
    \mathbb{E}_{\bm{X}}\left[I(\widehat{f}(\bm{X}) \neq \f(\bm{X}))\right]+
    \mathbb{E}_{\bm{X}}\left[I(\widetilde{f}(\bm{X}) \neq \tf(\bm{X}))\right].
\end{align*}
This completes the proof. \qed \\

\noindent
\textbf{Proof of Theorem \ref{Thm:UTBmore}.} We first notice that for any classifier $f \in \mathcal{F}$
\begin{align*}
R(f) =& \int_{\mathcal{X}} \Big[ I\big(f(\bm{x})<0\big)\eta(\bm{x})+
I\big(f(\bm{x})>0\big)\big(1-\eta(\bm{x})\big) \Big] P_{\bm{X}}(\bm{x})d\bm{x} \\
 =& 
\int_{\mathcal{X}} |2\eta(\bm{x})-1|I\big(f(\bm{x})(\eta(\bm{x})-1/2)<0\big)
P_{\bm{X}}(\bm{x})d\bm{x} 
+\int_{\mathcal{X}}\min\{ \eta(\bm{x}),1-\eta(\bm{x})\}
P_{\bm{X}}(\bm{x})d\bm{x} \\
=&\Phi(f) + \mathbb{E}_{\bm{X}}\big[
\min\{ \eta(\bm{X}),1-\eta(\bm{X})\}
\big] = \Phi(f) +R^\star,
\end{align*}
where $R^\star$ is the Bayes risk. Additionally, for any $f \in \mathcal{F}$, we have
\begin{align*}
& \left| \widetilde{\Phi}(f) - \Phi(f)\right|\\
=&\Big|
\int_{\mathcal{X}} |2\widetilde{\eta}(\bm{x})-1|I\big(f(\bm{x})(\widetilde{\eta}(\bm{x})-1/2)<0\big)
P_{\widetilde{\bm{X}}}(\bm{x})d\bm{x}- 
\int_{\mathcal{X}}
|2\eta(\bm{x})-1|I\big(f(\bm{x})(\eta(\bm{x})-1/2)<0\big)
P_{\bm{X}}(\bm{x})d\bm{x} \Big| \\
\leq &
\Big|
\int_{\mathcal{X}} |2\widetilde{\eta}(\bm{x})-1|I\big(f(\bm{x})(\widetilde{\eta}(\bm{x})-1/2)<0\big)
P_{\widetilde{\bm{X}}}(\bm{x})d\bm{x}- 
\int_{\mathcal{X}}
|2\widetilde{\eta}(\bm{x})-1|I\big(f(\bm{x})(\widetilde{\eta}(\bm{x})-1/2)<0\big)
P_{\bm{X}}(\bm{x})d\bm{x} \Big|\\
+
&
\Big|
\int_{\mathcal{X}} |2\widetilde{\eta}(\bm{x})-1|I\big(f(\bm{x})(\widetilde{\eta}(\bm{x})-1/2)<0\big)
P_{\bm{X}}(\bm{x})d\bm{x}- 
\int_{\mathcal{X}}
|2\eta(\bm{x})-1|I\big(f(\bm{x})(\eta(\bm{x})-1/2)<0\big)
P_{\bm{X}}(\bm{x})d\bm{x} \Big|\\
\triangleq &  A_1(g)+A_2(g).
\end{align*}
To bound $\big| \widetilde{\Phi}(g) - \Phi(g)\big|$, it remains to bound $A_1(g)$ and $A_2(g)$, respectively. $A_1(g)$ can be written as 
\begin{align*}
A_1(g) = &
\Big|
\int_{\mathcal{X}} |2\widetilde{\eta}(\bm{x})-1|I\big(f(\bm{x})(\widetilde{\eta}(\bm{x})-1/2)<0\big)
\Big(P_{\widetilde{\bm{X}}}(\bm{x})-P_{\bm{X}}(\bm{x})\Big)d\bm{x} \Big|.
\end{align*}

For $A_2(g)$, the first upper bound can be developed as
\begin{align*}
A_2(g) =&\Big|
\int_{\mathcal{X}} |2\widetilde{\eta}(\bm{x})-1|I\big(f(\bm{x})(\widetilde{\eta}(\bm{x})-1/2)<0\big)
P_{\bm{X}}(\bm{x})d\bm{x}\\
&- 
\int_{\mathcal{X}}
|2\eta(\bm{x})-1|I\big(f(\bm{x})(\eta(\bm{x})-1/2)<0\big)
P_{\bm{X}}(\bm{x})d\bm{x} \Big| \\
\leq &
\int_{\mathcal{X}}2 |\widetilde{\eta}(\bm{x})-\eta(\bm{x})|I\big(f(\bm{x})(\widetilde{\eta}(\bm{x})-1/2)<0\big)
P_{\bm{X}}(\bm{x})d\bm{x}+\\
&\Big|
\int_{\mathcal{X}}
|2\eta(\bm{x})-1|\Big[
I\big(f(\bm{x})(\widetilde{\eta}(\bm{x})-1/2)<0\big)-
I\big(f(\bm{x})(\eta(\bm{x})-1/2)<0\big) \Big]
P_{\bm{X}}(\bm{x})d\bm{x} \Big| \\
\leq &
2 \Vert \widetilde{\eta}-\eta\Vert_{L^2(\mathbb{P}_{X})}
\widetilde{C}(f)+
\Phi(\widetilde{f}^\star),
\end{align*}
where $\widetilde{C}(f)=\sqrt{\mathbb{P}\big(f(\bm{x})(\widetilde{\eta}(\bm{X})-1/2)<0\big)}$ and the last inequality follows from the Cauchy-Schwarz inequality. The second type of upper bound can be bounded as
\begin{align*}
    A_2(g) =&\Big|
\int_{\mathcal{X}} |2\widetilde{\eta}(\bm{x})-1|I\big(f(\bm{x})(\widetilde{\eta}(\bm{x})-1/2)<0\big)
P_{\bm{X}}(\bm{x})d\bm{x}\\
&- 
\int_{\mathcal{X}}
|2\eta(\bm{x})-1|I\big(f(\bm{x})(\eta(\bm{x})-1/2)<0\big)
P_{\bm{X}}(\bm{x})d\bm{x} \Big| \\
\leq &
\Big|
\int_{\mathcal{X}} |2\widetilde{\eta}(\bm{x})-1|
\left[I\big(f(\bm{x})(\widetilde{\eta}(\bm{x})-1/2)<0\big)
-I\big(f(\bm{x})(\eta(\bm{x})-1/2)<0\big)\right]
P_{\bm{X}}(\bm{x})d\bm{x}\Big| \\
&+2 \int_{\mathcal{X}}
|\eta(\bm{x})-\widetilde{\eta}(\bm{x})|I\big(f(\bm{x})(\eta(\bm{x})-1/2)<0\big)
P_{\bm{X}}(\bm{x})d\bm{x} \Big|. 
\end{align*}
Note that $I\big(f(\bm{x})(\widetilde{\eta}(\bm{x})-1/2)<0\big)
-I\big(f(\bm{x})(\eta(\bm{x})-1/2)<0\big) \leq I(f^\star(\bm{x}) \neq \widetilde{f}^\star(\bm{x}))$, we further have
\begin{align*}
    A_2(g)  \leq&  2\Big|
\int_{\mathcal{X}} |\widetilde{\eta}(\bm{x})-\eta(\bm{x})|
I(f^\star(\bm{x}) \neq \widetilde{f}^\star(\bm{x}))
P_{\bm{X}}(\bm{x})d\bm{x}\Big| \\
&+
2 \int_{\mathcal{X}}
|\eta(\bm{x})-\widetilde{\eta}(\bm{x})|I\big(f(\bm{x})(\eta(\bm{x})-1/2)<0\big)
P_{\bm{X}}(\bm{x})d\bm{x} \Big| \\
\leq & 2 \mathbb{E}_{\bm{X}\sim \mathbb{P}_{\bm{X}}}
\left\{
I \left(
\bm{X} \in \mathcal{A}^c(f^{\star},\widetilde{f}^{\star})
\right)
\left|\widetilde{\eta}(\bm{X})-\eta(\bm{X})\right|
\right\} +
2 \Vert \widetilde{\eta}-\eta\Vert_{L^2(\mathbb{P}_{X})}
C(f),
\end{align*}
where $C(f)=\sqrt{\mathbb{P}\big(f(\bm{x})(\eta(\bm{X})-1/2)<0\big)}$.

 Next, we turn to bound $\big| \Phi(\widetilde{f}_{\mathcal{F}}^\star) - \Phi(f_{\mathcal{F}}^\star) \big|$. By the fact that $f_{\mathcal{F}}^\star$ and $\widetilde{f}_{\mathcal{F}}^\star$ minimizes $\Phi(f)$ and $\widetilde{\Phi}(f)$, respectively, it follows that 
\begin{align*}
\Big|\Phi(\widetilde{f}_{\mathcal{F}}^\star) - \Phi(f_{\mathcal{F}}^\star) \Big| =&
\Phi(\widetilde{f}_{\mathcal{F}}^\star) - \Phi(f_{\mathcal{F}}^\star)   
\leq  
\big( \Phi(\widetilde{f}_{\mathcal{F}}^\star)- \widetilde{\Phi}(\widetilde{f}_{\mathcal{F}}^\star) \big)-
\big(  \Phi(f_{\mathcal{F}}^\star) -  \widetilde{\Phi}(f_{\mathcal{F}}^\star)  \big)\\
\leq &
\big| \Phi(\widetilde{f}_{\mathcal{F}}^\star)- \widetilde{\Phi}(\widetilde{f}_{\mathcal{F}}^\star)  \big|+\big| \Phi(f_{\mathcal{F}}^\star) -  \widetilde{\Phi}(f_{\mathcal{F}}^\star)   \big|.
\end{align*}
Applying a similar argument, we can conclude that
\begin{align*}
&U(\tf,\f) \leq \big| \Phi(\widetilde{f}_{\mathcal{F}}^\star)- \widetilde{\Phi}(\widetilde{f}_{\mathcal{F}}^\star)  \big|+\big| \Phi(f_{\mathcal{F}}^\star) -  \widetilde{\Phi}(f_{\mathcal{F}}^\star)   \big| \\
\leq &
2D_{\mathcal{K}}(\mathbb{P}_{\bm{X}},\mathbb{P}_{\widetilde{\bm{X}}})
+ 2 \Vert \widetilde{\eta}-\eta\Vert_{L^2(\mathbb{P}_{X})}
\left(\widetilde{C}(\tf)+C(\f)\right)+
4 \mathbb{E}
\left\{
I \left(
\bm{X} \in \mathcal{A}^c(f^{\star},\widetilde{f}^{\star})
\right)
\left|\widetilde{\eta}(\bm{X})-\eta(\bm{X})\right|
\right\},
\end{align*}
where $\mathcal{K} = \left\{k(\bm{x})=|2\widetilde{\eta}(\bm{x})-1|I\big(f(\bm{x})(\widetilde{\eta}(\bm{x})-1/2)<0\big), g \in \{f_{\mathcal{F}}^\star,\widetilde{f}^\star_{\mathcal{F}}\} \right\}$. This completes the proof. \qed \\

\noindent
\textbf{Proof of Theorem \ref{Thm:WUB}.} By Theorem \ref{Thm:UTBmore}, we have the following result for any $\mathcal{F}$ and $\mathbb{P}_{\widetilde{\bm{X}},\widetilde{Y}}$:
\begin{align*}
U(\tf,\f) 
\leq &2D_{\mathcal{K}(\mathcal{F})}(\mathbb{P}_{\bm{X}},\mathbb{P}_{\widetilde{\bm{X}}})
+ 4 \Lambda(\mathcal{F}) \Vert \widetilde{\eta}-\eta\Vert_{L^2(\mathbb{P}_{X})}+
4 \Upsilon(\widetilde{\eta}),
\end{align*}
where $\mathcal{K}(\mathcal{F}) = \left\{k(\bm{x})=|2\widetilde{\eta}(\bm{x})-1|I\big(f(\bm{x}) \neq \widetilde{f}^\star(\bm{x})\big), f \in \{f_{\mathcal{F}}^\star,\widetilde{f}^\star_{\mathcal{F}}\} \right\}$. It is worth noting that the functions in $\mathcal{K}(\mathcal{F})$ is always bounded by one for any $\mathcal{F}$. Therefore, we have
\begin{align*}
\sup_{\mathcal{F}}    D_{\mathcal{K}(\mathcal{F})}(\mathbb{P}_{\bm{X}},\mathbb{P}_{\widetilde{\bm{X}}})
\leq 
D_{\mathcal{H}_{ind}}(\mathbb{P}_{\bm{X}},\mathbb{P}_{\widetilde{\bm{X}}})
\triangleq \mathrm{TV}(\mathbb{P}_{\widetilde{\bm{X}}},\mathbb{P}_{\bm{X}}),
\end{align*}
where $\mathcal{H}_{ind}$ represents the set of all indicator functions. Next, using the fact that $\Lambda(\mathcal{F})\leq 1$ for any $\mathcal{F}$ and Lemma \ref{Lemma:CompLemma}, we have
\begin{align*}
   & 4 \Lambda(\mathcal{F}) \Vert \widetilde{\eta}-\eta\Vert_{L^2(\mathbb{P}_{X})}+
4 \Upsilon(\widetilde{\eta}) \\
\leq &4\Vert \widetilde{\eta}-\eta\Vert_{L^2(\mathbb{P}_{X})}+
\max\left\{
4 C_0^{\frac{1}{\gamma+2}}\left(\Vert \eta - \widetilde{\eta} \Vert_{L^2(\mathbb{P}_{\bm{X}})}^2\right)^{\frac{\gamma+1}{\gamma+2}},
4\widetilde{C}_0^{\frac{1}{\widetilde{\gamma}+2}}\left(\Vert \eta - \widetilde{\eta} \Vert_{L^2(\mathbb{P}_{\bm{X}})}^2\right)^{\frac{\widetilde{\gamma}+1}{\widetilde{\gamma}+2}}
\right\} \\
\leq & C_1 \left(\Vert \widetilde{\eta}-\eta\Vert_{L^2(\mathbb{P}_{X})}+\Vert \widetilde{\eta}-\eta  \Vert_{L^2(\mathbb{P}_{\bm{X}})}^{\frac{2\gamma^\prime+2}{\gamma^\prime+2}}\right),
\end{align*}
for some positive constant $C$.
To sum up, we have
\begin{align*}
\overline{U}(\mathbb{P}_{\widetilde{\bm{X}},\widetilde{Y}})=\sup_{\mathcal{F}} U(\tf,\f) \leq
2\mathrm{TV}(\mathbb{P}_{\bm{X}},\mathbb{P}_{\widetilde{\bm{X}}})
+ C_1 \left(\Vert \widetilde{\eta}-\eta\Vert_{L^2(\mathbb{P}_{X})}+\Vert \widetilde{\eta}-\eta  \Vert_{L^2(\mathbb{P}_{\bm{X}})}^{\frac{2\gamma^\prime+2}{\gamma^\prime+2}}\right).
\end{align*}
This completes the proof. \qed \\

\noindent
\textbf{Proof of Theorem \ref{Thm:ConditGene}.} By Theorem \ref{Thm:UTBmoreFinal}, we have
\begin{align*}
    \mathbb{E}_{\mathcal{D}}\left\{\overline{U}(\mathbb{P}_{\widetilde{\bm{X}}, \widetilde{Y}})
    \right\}
\lesssim \mathbb{E}_{\mathcal{D}}\left\{\mathrm{TV}(\mathbb{P}_{\bm{X}},\mathbb{P}_{\widetilde{\bm{X}}})
\right\}
+ \mathbb{E}_{\mathcal{D}}\left\{\Vert \widetilde{\eta}-\eta\Vert_{L^2(\mathbb{P}_{X})}
\right\}.
\end{align*}

Next, we provide upper bounds for $\mathbb{E}_{\mathcal{D}}\left\{\mathrm{TV}(\mathbb{P}_{\bm{X}},\mathbb{P}_{\widetilde{\bm{X}}})
\right\}$ and $\mathbb{E}_{\mathcal{D}}\left\{\Vert \widetilde{\eta}-\eta\Vert_{L^2(\mathbb{P}_{X})}
\right\}$, respectively. First, the total variation distance between $\mathbb{P}_{\bm{X}}$ and $\mathbb{P}_{\widetilde{\bm{X}}}$ can be upper bounded as follows \citep{sriperumbudur2012empirical}:
\begin{align*}
\mathrm{TV}(\mathbb{P}_{\widetilde{\bm{X}}},\mathbb{P}_{\bm{X}}) 
    \leq &
\sup_{h \in \mathcal{H}_{1}}
\left|
\mathbb{E}_{\widetilde{\bm{X}} \sim \mathbb{P}_{\widetilde{\bm{X}}}}\Big[
h(\widetilde{\bm{X}})\Big]-
\mathbb{E}_{\bm{X}\sim \mathbb{P}_{\bm{X}}}\Big[
h(\bm{X})
\Big]
\right| \\
= &
\sup_{h \in \mathcal{H}_{1}}
\left\{
\mathbb{E}_{\widetilde{\bm{X}} \sim \mathbb{P}_{\widetilde{\bm{X}}}}\Big[
h(\widetilde{\bm{X}})
\Big]-
\mathbb{E}_{\bm{X}\sim \mathbb{P}_{\bm{X}}}\Big[
h(\bm{X})
\Big]\right\}
\triangleq
D_{\mathcal{H}_1}(\mathbb{P}_{\widetilde{\bm{X}}},\mathbb{P}_{\bm{X}})
,
\end{align*}
where $\mathcal{H}_1 = \{h(\bm{x}):\Vert h \Vert_{\infty} \leq 1\}$ is the class of all measurable functions bounded by 1, and the equality follows from the fact that $\mathcal{H}_1=-\mathcal{H}_1$.

Next, using the triangle inequality, $D_{\mathcal{H}_1}(\mathbb{P}_{\widetilde{\bm{X}}},\mathbb{P}_{\bm{X}})
$ can be bounded as
\begin{align}
\label{Gan:Eq1}
D_{\mathcal{H}_1}(\mathbb{P}_{\widetilde{\bm{X}}},\mathbb{P}_{\bm{X}})
    \leq 
D_{\mathcal{H}_1}(\mathbb{P}_{\widetilde{\bm{X}}},\widehat{\mathbb{P}}_{\bm{X}})+
D_{\mathcal{H}_1}(\widehat{\mathbb{P}}_{\bm{X}},\mathbb{P}_{\bm{X}})
\leq 
D_{\mathcal{H}_1}(\mathbb{P}_{\widetilde{\bm{X}}}^\star,\widehat{\mathbb{P}}_{\bm{X}})+
D_{\mathcal{H}_1}(\widehat{\mathbb{P}}_{\bm{X}},\mathbb{P}_{\bm{X}}),
\end{align}
where the second inequality follows from the fact that $\mathbb{P}_{\widetilde{\bm{X}}}$ is derived from the optimal neural generator $g_{\widehat{\bm{\theta}}}$ as defined in (\ref{GAN:generator}) and $\mathbb{P}_{\widetilde{\bm{X}}}^\star$ denotes the output distribution of $g_{\bm{\theta}^\star}$, where $g_{\bm{\theta}^\star} = \arg\min_{g \in \mathcal{G}} D_{\mathcal{H}_1}(\mathbb{P}_{g_{\bm{\theta}}(\bm{Z})}, \mathbb{P}_{\bm{X}})$. This indicates that among all distributions generated by $g_{\bm{\theta}}$ within $\mathcal{G}$, $\mathbb{P}_{\widetilde{\bm{X}}}^\star$ has the smallest distance to $\mathbb{P}_{\bm{X}}$ according to the metric $D_{\mathcal{H}_1}$. Applying the triangle inequality to (\ref{Gan:Eq1}), it follows that
\begin{align*}
D_{\mathcal{H}_1}(\mathbb{P}_{\widetilde{\bm{X}}},\mathbb{P}_{\bm{X}})
\leq 
D_{\mathcal{H}_1}(\mathbb{P}_{\widetilde{\bm{X}}}^\star,\mathbb{P}_{\bm{X}})+
2D_{\mathcal{H}_1}(\widehat{\mathbb{P}}_{\bm{X}},\mathbb{P}_{\bm{X}}).
\end{align*}
Therefore, we further have
\begin{align*}
    \mathbb{E}_{\mathcal{D}}\left\{ D_{\mathcal{H}_1}(\mathbb{P}_{\widetilde{\bm{X}}},\mathbb{P}_{\bm{X}}) \right\}
    \leq 
    D_{\mathcal{H}_1}(\mathbb{P}_{\widetilde{\bm{X}}}^\star,\mathbb{P}_{\bm{X}})+
    2\mathbb{E}_{\mathcal{D}}\left\{ D_{\mathcal{H}_1}(\widehat{\mathbb{P}}_{\bm{X}},\mathbb{P}_{\bm{X}})
    \right\}.
\end{align*}

\noindent
\textbf{Step 1: Bounding $\mathbb{E}\left\{ D_{\mathcal{H}_1}(\widehat{\mathbb{P}}_{\bm{X}},\mathbb{P}_{\bm{X}})
    \right\}$.} The proof of this part is mainly based on that of Theorem 4 of \citet{liang2021well}. Given that $P_{\bm{X}}$ is supported on $\mathcal{X}=[0,1]^p$, by Assumption \ref{Ass:FeaSmoo}, $P_{\bm{X}}$ is squared-integrable and hence admits the Fourier trigonometric series decomposition:
\begin{align*}
    P(\bm{x}) = \sum_{\bm{\xi} \in \mathbb{N}^p} \theta_{\bm \xi}\psi_{\bm \xi}(\bm{x}),
\end{align*}
where $\mathbb{N}$ denotes the set of all natural numbers, $\bm{\xi}$ is a $p$-dimensional integer vector representing a multi-index, and $\psi_{\bm \xi}(\bm{x})=\prod_{i=1}^p \psi_{\xi_i}(x_i)$ represents a tensorized orthogonal basis.

Next, we construct a smooth version of the empirical distribution of $(\bm{x}_1,\ldots,\bm{x}_n)$, which is given as
\begin{align*}
    \widehat{P}(\bm{x})=
   \sum_{\bm{\xi} \in \mathbb{N}^p}
   \widehat{\theta}_{\bm{\xi}} \psi_{\bm{\xi}}(\bm{x}),
\end{align*}
where $\widehat{\theta}_{\bm{\xi}}$ is defined as
\begin{align*}
    \widehat{\theta}_{\bm{\xi}}=
    \begin{cases}
        \frac{1}{n}\sum_{i=1}^n \prod_{j=1}^p
        \psi_{\xi_j}(x_{ij}), &\mbox{ for $\bm{\xi}$ satisfies $\Vert\bm{\xi}\Vert_{\infty} \leq M$}, \\
        0, &\mbox{ otherwise}.
    \end{cases}
\end{align*}
According to \citet{liang2021well}, $\widehat{P}$ filters out all the high frequency components, when the multi-index $\max_{i \in [p]}\xi_i$ is upper bounded by $M$. Next, for any function $ h \in \mathcal{H}_1$, note that $ h $ is defined on $\mathcal{X}$ and is bounded by 1. Therefore, $ h^2(\bm{x}) $ is integrable on $[0,1]^p$. Similarly, we can decompose $ h $ as $h(\bm{x})=\sum_{\bm{\xi} \in \mathbb{N}^p}\theta_{\bm{\xi}}(h)\psi_{\bm{\xi}}(\bm{x})$. Subsequently, we have
\begin{align*}
    \mathbb{E}
    \left(
    D_{\mathcal{H}_1}(\widehat{\mathbb{P}}_{\bm{X}},\mathbb{P}_{\bm{X}})
    \right)
   & =
    \mathbb{E}
    \left(
\sup_{h \in \mathcal{H}_1}
\int_{\mathcal{X}}h(\bm{x})
\left(
P_{\bm{X}}(\bm{x})-
\widehat{P}_{\bm{X}}(\bm{x})
\right)d\bm{x}
    \right) \\
    &=
        \mathbb{E}
    \left(
\sup_{h \in \mathcal{H}_1}
\int_{\mathcal{X}}h(\bm{x})
\left(
   \sum_{\bm{\xi} \in \mathbb{N}^p}
   \theta_{\bm{\xi}} \psi_{\bm{\xi}}(\bm{x})-
   \sum_{\bm{\xi} \in \mathbb{N}^p}
   \widehat{\theta}_{\bm{\xi}} \psi_{\bm{\xi}}(\bm{x})
\right)d\bm{x}
    \right) \\
 & = 
    \mathbb{E}
    \left(
\sup_{h \in \mathcal{H}_1}
\int_{\mathcal{X}}h(\bm{x})
\left(
   \sum_{\bm{\xi} \in \mathbb{N}^p}
   \theta_{\bm{\xi}} \psi_{\bm{\xi}}(\bm{x})-
   \sum_{\bm{\xi} \in \mathbb{N}^p}
   \widehat{\theta}_{\bm{\xi}} \psi_{\bm{\xi}}(\bm{x})
\right)d\bm{x}
    \right) \\
     & = 
    \mathbb{E}
    \left(
\sup_{h \in \mathcal{H}_1}
   \sum_{\bm{\xi} \in \mathbb{N}^p}
\int_{\mathcal{X}}
h(\bm{x})\psi_{\bm{\xi}}(\bm{x})
\left(
   \theta_{\bm{\xi}} -
   \widehat{\theta}_{\bm{\xi}} 
\right)d\bm{x}
    \right).
\end{align*}
Note that $\int_{\mathcal{X}}
h(\bm{x})\psi_{\bm{\xi}}(\bm{x})
d\bm{x} = \theta_{\bm{\xi}}(h)$, it follows that
\begin{align*}
    &    \mathbb{E}
    \left(
    D_{\mathcal{H}_1}(\widehat{\mathbb{P}}_{\bm{X}},\mathbb{P}_{\bm{X}})
    \right) = 
    \mathbb{E}
    \left(
\sup_{h \in \mathcal{H}_1}
   \sum_{\bm{\xi} \in \mathbb{N}^p}
\theta_{\bm{\xi}}(h)
\left(
   \theta_{\bm{\xi}} -
   \widehat{\theta}_{\bm{\xi}} 
\right)
    \right) \\
    \leq &
    \mathbb{E}
    \left(
\sup_{h \in \mathcal{H}_1}
   \sum_{\bm{\xi} \in [M]^p}
\theta_{\bm{\xi}}(h)
\left(
   \theta_{\bm{\xi}} -
   \widehat{\theta}_{\bm{\xi}} 
\right)
    \right)+
        \mathbb{E}
    \left(
\sup_{h \in \mathcal{H}_1}
   \sum_{\bm{\xi} \in \mathbb{N}^p\setminus [M]^p}
\theta_{\bm{\xi}}(h)
\left(
   \theta_{\bm{\xi}} -
   \widehat{\theta}_{\bm{\xi}} 
\right)
    \right)  \\
= &      \mathbb{E}
    \left(
\sup_{h \in \mathcal{H}_1}
   \sum_{\bm{\xi} \in [M]^p}
\theta_{\bm{\xi}}(h)
\left(
   \theta_{\bm{\xi}} -
   \widehat{\theta}_{\bm{\xi}} 
\right)
    \right)+
        \mathbb{E}
    \left(
\sup_{h \in \mathcal{H}_1}
   \sum_{\bm{\xi} \in \mathbb{N}^p\setminus [M]^p}
\theta_{\bm{\xi}}(h)
   \theta_{\bm{\xi}} 
    \right)   \triangleq E_1 + E_2,
\end{align*}
where the last equality follows from the fact that $\widehat{\theta}_{\bm{\xi}} =0$ for $\bm{\xi} \in \mathbb{N}^p\setminus [M]^p$. Next, we turn to bound $E_1$ and $E_2$ separately. For $E_1$, by Cauchy-Schwarz inequality, we have
\begin{align*}
    E_1 = &\mathbb{E}
    \left(
\sup_{h \in \mathcal{H}_1}
   \sum_{\bm{\xi} \in [M]^p}
\theta_{\bm{\xi}}(h)
\left(
   \theta_{\bm{\xi}} -
   \widehat{\theta}_{\bm{\xi}} 
\right)
    \right) \leq 
    \mathbb{E}
    \left(
\sup_{h \in \mathcal{H}_1}
  \sqrt{\sum_{\bm{\xi} \in [M]^p}
\theta_{\bm{\xi}}^2(h)}\sqrt{
  \sum_{\bm{\xi} \in [M]^p}
\left(
   \theta_{\bm{\xi}} -
   \widehat{\theta}_{\bm{\xi}} 
\right)^2}
    \right) \\
    \leq &
    \mathbb{E}
    \left(
\sqrt{
  \sum_{\bm{\xi} \in [M]^p}
\left(
   \theta_{\bm{\xi}} -
   \widehat{\theta}_{\bm{\xi}} 
\right)^2}
    \right)  \leq     
\sqrt{
  \sum_{\bm{\xi} \in [M]^p}
  \mathbb{E}
    \left(
   \theta_{\bm{\xi}} -
   \widehat{\theta}_{\bm{\xi}} 
\right)^2 } \leq \sqrt{\frac{M^p}{n}},
\end{align*}
where the second last inequality follows from the Jensen's inequality, the third last inequality follows from the fact that $\sup_{h \in \mathcal{H}_1}\sum_{\bm{\xi} \in [M]^p}\theta_{\bm{\xi}}^2(h)\leq 1$ (the Parseval's Theorem;  \citealp{folland2009fourier}), and the last inequality holds by the fact that
\begin{align*}
    \mathbb{E}
    \left(
    \widehat{\theta}_{\bm{\xi}} -
   \theta_{\bm{\xi}} 
\right)^2 = \mathbb{E}\left(
\frac{1}{n}\sum_{i=1}^n \prod_{j=1}^p
        \psi_{\xi_j}(x_{ij})
- \int_{\mathcal{X}}P_{\bm{X}}(\bm{x})\psi_{\bm{\xi}}(\bm{x})d\bm{x}
        \right)^2 \leq
        \frac{1}{n} \mathbb{E}_{\bm{X}\sim \mathbb{P}_{\bm{X}}}(\psi^2_{\bm{\xi}}(\bm{X}))\leq \frac{1}{n},
\end{align*}
where the last inequality follows by the fact that $\psi_{\bm{\xi}}(\bm{x})$ is a product of trigonometric functions.

In what follows, we proceed to bound $E_2$.

\begin{align*}
     E_2 =  &      \mathbb{E}
    \left(
\sup_{h \in \mathcal{H}_1}
   \sum_{\bm{\xi} \in \mathbb{N}^p\setminus [M]^p}
\theta_{\bm{\xi}}(h)
   \theta_{\bm{\xi}} 
    \right)  \leq 
          \mathbb{E}
    \left(
\sup_{h \in \mathcal{H}_1}
\sqrt{
   \sum_{\bm{\xi} \in \mathbb{N}^p\setminus [M]^p}
\theta_{\bm{\xi}}^2(h)
}    \right) \cdot
\sqrt{
   \sum_{\bm{\xi} \in \mathbb{N}^p\setminus [M]^p}
   \theta_{\bm{\xi}}^2} \\
   =&
   \sup_{h \in \mathcal{H}_1}
\sqrt{
   \sum_{\bm{\xi} \in \mathbb{N}^p\setminus [M]^p}
\theta_{\bm{\xi}}^2(h)
}  \cdot \sqrt{
   \sum_{\bm{\xi} \in \mathbb{N}^p\setminus [M]^p}
   \theta_{\bm{\xi}}^2} \leq  \sqrt{   \frac{1}{(1+M^2)^{\alpha}}
   \sum_{\bm{\xi} \in \mathbb{N}^p\setminus [M]^p}
   (1+\Vert \bm{\xi}\Vert_2^2)^{\alpha}
   \theta_{\bm{\xi}}^2},
\end{align*}
where the last inequality follows from the fact that $\sup_{h \in \mathcal{H}_1}
\sqrt{
   \sum_{\bm{\xi} \in \mathbb{N}^p\setminus [M]^p}
\theta_{\bm{\xi}}^2(h)
} \leq 1$. Note that, by Assumption \ref{Ass:FeaSmoo}, we have
\begin{align}
\label{Ineq1}
    \sqrt{  
   \sum_{\bm{\xi} \in \mathbb{N}^p\setminus [M]^p}
   (1+\Vert \bm{\xi}\Vert_2^2)^{\alpha}
   \theta_{\bm{\xi}}^2} \lesssim r.
\end{align}
Here (\ref{Ineq1}) can be derived from the relationship between the decay rate of Fourier coefficients and the smoothness of Sobolev space \citep{bogachev2020fourier}. Plugging (\ref{Ineq1}) to the upper bound for $E_2$, it holds that
\begin{align*}
    E_1 + E_2 \lesssim \sqrt{\frac{M^p}{n}} + \frac{r}{(1+M^2)^{\alpha/2}}
    \lesssim \frac{M^{p/2}}{\sqrt{n}} + \frac{r}{M^{\alpha}}.
\end{align*}
Choosing $M \asymp (n r^2)^{\frac{1}{2\alpha+p}}$, we have
\begin{align}
\label{BoundStep1}
    \mathbb{E}_{\mathcal{D}}
    \left(
    D_{\mathcal{H}_1}(\widehat{\mathbb{P}}_{\bm{X}},\mathbb{P}_{\bm{X}})
    \right) \lesssim n^{-\frac{\alpha}{2\alpha+p}},
\end{align}
    with the constant depending on $r$ being omitted. This completes the proof of Step 1.

\noindent
\textbf{Step 2: Bounding $\Vert \eta - \widetilde{\eta}\Vert_{L^2(\mathbb{P}_{\bm{X}})}$.} We first consider the decomposition of $\Vert \eta - \widetilde{\eta}\Vert_{L^2(\mathbb{P}_{\bm{X}})}$ into an approximation error term and a estimation error term.
\begin{align}
\label{ErrorDecomp}
    \Vert \eta - \widetilde{\eta}\Vert_{L^2(\mathbb{P}_{\bm{X}})}
    \leq 
    \underbrace{\Vert \eta - L_{\bm{\omega}^\star}\Vert_{L^2(\mathbb{P}_{\bm{X}})}}_{\text{Approximation Error}}
   +  \underbrace{\Vert L_{\widehat{\bm{\omega}}} - L_{\bm{\omega}^\star}\Vert_{L^2(\mathbb{P}_{\bm{X}})}}_{\text{Estimation Error}},
\end{align}
where $L_{\bm{\omega}^\star}= \argmin\limits_{L_{\bm{\omega}}\in \mathcal{N}} \Vert \eta - L_{\bm{\omega}}\Vert_{L^2(\mathbb{P}_{\bm{X}})}$ denotes the optimal neural network within $\mathcal{N}(W^\prime,L^\prime)$ for approximating $\eta$. 

The approximation capacity of fully-connected neural networks has been widely studied in the literature \citep{nakada2020adaptive, siegel2023optimal, NNregres, yang2024optimal}, demonstrating that the approximation error decreases as the number of parameters in the neural networks increases. By Assumption \ref{Ass:FeaSmoo}, we can verify that $\eta$ belongs to the Sobolev space with smoothness parameter $\beta$. By Theorem 2.1 of \citet{yang2024optimal}, it holds that
\begin{align}
\label{ApproError}
    \Vert \eta - L_{\bm{\omega}^\star}\Vert_{L^{\infty}(\mathbb{P}_{\bm{X}})}
    \leq C_0 (WL)^{-\frac{2\beta}{p}},
\end{align}
for some positive constants $C_0$. This result reduces to the result in Theorem 1 of \citet{siegel2023optimal} if the width $W$ is fixed. Additionally, (\ref{ApproError}) implies that $\Vert L_{\bm{\omega}^\star} \Vert_{L^{\infty}(\mathbb{P}_{\bm{X}})} \leq 1+C_0$ since $WL)^{-\frac{2\beta}{p}} \leq 1$.

Next, we provide a large deviation inequality for $\Vert \eta - \widetilde{\eta}\Vert_{L^2(\mathbb{P}_{\bm{X}})}$. Notice that $\widetilde{\eta}(\bm{x})=\min\{1,\max\{L_{\widehat{\bm{\omega}}}(\bm{x}),0\}\}$, we have
\begin{align}
\label{ClipIneq}
    \frac{1}{n}\sum_{i=1}^n (\widetilde{\eta}(\bm{x}_i)-y_i^\prime)^2 \leq \frac{1}{n}\sum_{i=1}^n (L_{\widehat{\bm{\omega}}}-y_i^\prime)^2.
\end{align}
Here (\ref{ClipIneq}) holds due to the fact that $y_i^\prime \in \{0,1\}$. If $L_{\widehat{\bm{\omega}}}$ is outside $[0,1]$, the clipping operation makes $\widetilde{\eta}$ closer to $y_i^\prime$. Therefore, we define a class of clipped neural networks as
\begin{align*}
    \mathcal{N}^c = \left\{L^c_{\bm{\omega}}(\bm{x})=\min\{1,\max\{L_{\bm{\omega}}(\bm{x}),0\}\}:L_{\bm{\omega}} \in \mathcal{N}  \right\}.
\end{align*}
For any $L_{\bm{\omega}} \in \mathcal{N}$, it holds that
\begin{align*}
    \mathbb{E}\left(L^c_{\bm{\omega}}(\bm{x}) - Y^\prime\right)^2
    \leq \mathbb{E}\left(L_{\bm{\omega}}(\bm{x}) - Y^\prime\right)^2
    \mbox{ and }
    \Vert \eta - L^c_{\bm{\omega}}\Vert_{L^2(\mathbb{P}_{\bm{X}})}
    \leq \Vert \eta - L_{\bm{\omega}}\Vert_{L^2(\mathbb{P}_{\bm{X}})},
\end{align*}
where $Y^\prime = (Y+1)/2$. Let $\mathcal{N}_{\delta} = \{ L^c_{\bm{\omega}} \in \mathcal{N}^c:\Vert\eta-L^c_{\bm{\omega}}\Vert_{L^2(\mathbb{P}_{\bm{X}})}^2 \geq \delta  \}$. For any $\delta>0$,
\begin{align*}
\mathbb{P}\Big(
\Vert
\eta-\widetilde{\eta}\Vert_{L^2(\mathbb{P}_{\bm{X}})}^2 \geq \delta \Big) \leq 
\mathbb{P}\Big( \sup_{L^c_{\bm{\omega}} \in \mathcal{N}_{\delta}}
\frac{1}{n}\sum_{i=1}^n (L_{\bm{\omega}^\star}(\bm{x}_i)-y_i^\prime)^2 - \frac{1}{n}\sum_{i=1}^n (L_{\bm{\omega}}^c(\bm{x}_i)-y_i^\prime)^2\geq 0 \Big).
\end{align*}
For ease of notation, we denote that $U_n(\bm{\omega}) = \frac{1}{n}\sum_{i=1}^n (L^c_{\bm{\omega}}(\bm{x}_i)-y_i^\prime)^2$ and $U(\bm{\omega}) = \mathbb{E}(L^c_{\bm{\omega}}(\bm{X})-Y^\prime)^2$. Here it should be noted that for any $\bm{\omega}$,
\begin{align*}
    U(\bm{\omega}) = & \mathbb{E}(L^c_{\bm{\omega}}(\bm{X})-Y^\prime)^2=
    \mathbb{E}(L^c_{\bm{\omega}}(\bm{X})-\eta(\bm{X})+\eta(\bm{X})-Y^\prime)^2 \\
    =&\mathbb{E}(L^c_{\bm{\omega}}(\bm{X})-\eta(\bm{X}))^2 +
    \mathbb{E}(\eta(\bm{X})-Y^\prime)^2 = 
   \Vert\eta-L^c_{\bm{\omega}}\Vert_{L^2(\mathbb{P}_{\bm{X}})}^2 + U(\eta).
\end{align*}
Particularly, for $\bm{\omega}^\star$, with a slight abuse of notation, we consider
\begin{align*}
 U_n(\bm{\omega}^\star) = \frac{1}{n}\sum_{i=1}^n (L_{\bm{\omega}^\star}(\bm{x}_i)-y_i^\prime)^2 \mbox{ and } U(\bm{\omega}^\star) = \Vert\eta-L_{\bm{\omega}^\star}\Vert_{L^2(\mathbb{P}_{\bm{X}})}^2 + U(\eta).
\end{align*}
Here for $\bm{\omega}^\star$, we use the non-clipped neural network. Then, we have
\begin{align*}
\mathbb{P}\Big(
\Vert
\eta-\widetilde{\eta}\Vert_{L^2(\mathbb{P}_{\bm{X}})}^2 \geq \delta \Big)
\leq 
\mathbb{P}\Big(
\sup_{L^c_{\bm{\omega}} \in \mathcal{N}_{\delta}}
U_n(\bm{\omega}^\star) -  U_n(\bm{\omega}) \geq 0
\Big).
\end{align*}
Notice that $\mathcal{N}_{\delta}$ admits the decomposition as $\mathcal{N}_{\delta} = \cup_{i=1}^{n_0} \mathcal{H}_i$, where $n_0$ is the largest integer such that $2^{n_0}\delta \leq 1$ and $\mathcal{H}_i$ is defined as
$$
\mathcal{H}_{i} = \left\{ L^c_{\bm{\omega}} \in \mathcal{N}_{\delta}:2^{i-1}\delta \leq \Vert \eta-L^c_{\bm{\omega}}\Vert_{L^2(\mathbb{P}_{\bm{X}})}^2 
\leq \Vert \eta-L^c_{\bm{\omega}}\Vert_{L^{\infty}(\mathbb{P}_{\bm{X}})}^2 
\leq 2^i \delta\right\}.
$$ 
Therefore, we further have
\begin{align*}
\mathbb{P}\Big(
\Vert
\eta-\widetilde{\eta}\Vert_{L^2(\mathbb{P}_{\bm{X}})}^2 \geq \delta \Big)
\leq 
\sum_{i=1}^n \mathbb{P}\Big(
\sup_{L^c_{\bm{\omega}} \in \mathcal{H}_i}
U_n(\bm{\omega}^\star) -  U_n(\bm{\omega}) \geq 0
\Big).
\end{align*}
Clearly, it suffices to bound $\mathbb{P}\Big(
\sup_{L^c_{\bm{\omega}} \in \mathcal{H}_i}
U_n(\bm{\omega}^\star) -  U_n(\bm{\omega}) \geq 0
\Big)$ for upper bounding $\mathbb{P}\Big(
\Vert \eta-\widetilde{\eta}\Vert_{L^2(\mathbb{P}_{\bm{X}})}^2 \geq \delta \Big)$. For $i \geq 1$, 
\begin{align*}
&
\mathbb{P}\Big(
\sup_{L^c_{\bm{\omega}} \in \mathcal{H}_i}
U_n(\bm{\omega}^\star) -  U_n(\bm{\omega}) \geq 0
\Big)  \\
\leq &
\mathbb{P}\Big(
\sup_{L^c_{\bm{\omega}} \in \mathcal{H}_i}
\big[U_n(\bm{\omega}^\star)- U(\bm{\omega}^\star)\big] -  \big[U_n(\bm{\omega}) - U(\bm{\omega})\big] \geq \inf_{L^c_{\bm{\omega}} \in \mathcal{H}_i} U(\bm{\omega}) - U(\bm{\omega}^\star)
\Big) \\
= &
\mathbb{P}\Big(
\sup_{L^c_{\bm{\omega}} \in \mathcal{H}_i}
\big[U_n(\bm{\omega}^\star)- U(\bm{\omega}^\star)\big] -  \big[U_n(\bm{\omega}) - U(\bm{\omega})\big] \geq \inf_{L^c_{\bm{\omega}} \in \mathcal{H}_i} U(\bm{\omega}) -U(\eta)+ U(\eta)- U(\bm{\omega}^\star)
\Big) \\
\leq &
\mathbb{P}\Big(
\sup_{L^c_{\bm{\omega}} \in \mathcal{H}_i}
\big[U_n(\bm{\omega}^\star)- U(\bm{\omega}^\star)\big] -  \big[U_n(\bm{\omega}) - U(\bm{\omega})\big] \geq 2^{i-1}\delta -  \Vert \eta- L_{\bm{\omega}^\star}\Vert_{L^2(\mathbb{P}_{\bm{X}})}^2
\Big).
\end{align*}
Here we let $\Vert \eta- L_{\bm{\omega}^\star}\Vert_{L^2(\mathbb{P}_{\bm{X}})}^2 \leq \delta/4$, we have
\begin{align*}
    \mathbb{P}\Big(
\sup_{L^c_{\bm{\omega}} \in \mathcal{H}_i}
U_n(\bm{\omega}^\star) -  U_n(\bm{\omega}) \geq 0
\Big) \leq 
\mathbb{P}\Big(
\sup_{L^c_{\bm{\omega}} \in \mathcal{H}_i}
\big[U_n(\bm{\omega}^\star)- U(\bm{\omega}^\star)\big] -  \big[U_n(\bm{\omega}) - U(\bm{\omega})\big] \geq 2^{i-2}\delta
\Big).
\end{align*}
Define $V_i(\bm{\omega})$ as
$$
V_i(\bm{\omega}) =(L_{\bm{\omega}^\star}(\bm{x}_i)-y_i^\prime)^2- (L^c_{\bm{\omega}}(\bm{x}_i)-y_i^\prime)^2
-U(\bm{\omega}^\star)+U(\bm{\omega}).
$$ 
Here it can be verified that $|V_i(\bm{\omega})|\leq (C_0+2)^2$. Then we further have
\begin{align}
\label{LCE}
     & \mathbb{P}\left(
\sup_{L^c_{\bm{\omega}} \in \mathcal{H}_i}
U_n(\bm{\omega}^\star) -  U_n(\bm{\omega}) \geq 0
\right) \notag \\
\leq &   \mathbb{P}\left\{
\sup_{L^c_{\bm{\omega}} \in \mathcal{H}_i}
\frac{1}{n}
\sum_{i=1}^n V_i(\bm{\omega})
-
\frac{1}{n}
\mathbb{E}\left(
\sup_{L^c_{\bm{\omega}} \in \mathcal{H}_i}
\sum_{i=1}^n V_i (\bm{\omega})
\right)
\geq 2^{i-2}\delta-
\mathbb{E}\left(
\sup_{L^c_{\bm{\omega}} \in \mathcal{H}_i}
\frac{1}{n}
\sum_{i=1}^n V_i (\bm{\omega})
\right)
\right\}.
\end{align}

Next, we turn to bound $\mathbb{E}\left(
\sup_{L^c_{\bm{\omega}} \in \mathcal{H}_i}
\sum_{i=1}^n V_i (\bm{\omega})
\right)$ based on the Rademacher complexity. By the symmetrization argument for the Rademacher complexity (Lemma 4 of \citet{bousquet2003introduction}), we have
\begin{align*}
\mathbb{E}_{\mathcal{D}}\left(
\sup_{L^c_{\bm{\omega}} \in \mathcal{H}_i}
\sum_{i=1}^n V_i (\bm{\omega})
\right) \leq &2
\mathbb{E}_{\mathcal{D},\bm{\xi}}\left(
\sup_{L^c_{\bm{\omega}} \in \mathcal{H}_i}
\sum_{i=1}^n \xi_i 
\left[
(L_{\bm{\omega}^\star}(\bm{x}_i)-y_i^\prime)^2- (L^c_{\bm{\omega}}(\bm{x}_i)-y_i^\prime)^2
\right]
\right) \\
=&2
\mathbb{E}_{\mathcal{D},\bm{\xi}}\left(
\sup_{L^c_{\bm{\omega}} \in \mathcal{H}_i}
\sum_{i=1}^n \xi_i 
Q_{\bm{\omega}}(\bm{x}_i,y_i^\prime)
\right),
\end{align*}
where $\xi_i$'s are independent Rademacher random variables and $Q_{\bm{\omega}}(\bm{x}_i,y_i^\prime)=
(L^c_{\bm{\omega}^\star}(\bm{x}_i)-y_i^\prime)^2- (L^c_{\bm{\omega}}(\bm{x}_i)-y_i^\prime)^2$. Here $\frac{1}{\sqrt{n}}\sum_{i=1}^n \xi_i 
Q_{\bm{\omega}}(\bm{x}_i,y_i^\prime)$ is a sub-Gaussian process with metric $\rho_{\mathcal{D}}$
\begin{align*}
    \rho_{\mathcal{D}}(Q_{\bm{\omega}_1},Q_{\bm{\omega}_2})=
    \sqrt{\frac{1}{n}\sum_{i=1}^n \left(Q_{\bm{\omega}_1}(\bm{x}_i,y_i^\prime)-Q_{\bm{\omega_2}}(\bm{x}_i,y_i^\prime)\right)^2}.
\end{align*}
Let $\bar{\rho}_{\mathcal{D}}=\sup_{L^c_{\bm{\omega}_1},L^c_{\bm{\omega}_2} \in \mathcal{H}_i}\rho_{\mathcal{D}}(Q_{\bm{\omega}_1},Q_{\bm{\omega}_2})$ be the diameter of $\mathcal{H}_i$. Note that, for $L^c_{\bm{\omega}_1},L^c_{\bm{\omega}_2}\in \mathcal{H}_i$, it holds that
\begin{align*}
&\left(Q_{\bm{\omega}_1}(\bm{x}_i,y_i^\prime)-Q_{\bm{\omega_2}}(\bm{x}_i,y_i^\prime)\right)^2 \\
=&  \left\{ \left[
(L^c_{\bm{\omega}^\star}(\bm{x}_i)-y_i^\prime)^2- (L^c_{\bm{\omega}_1}(\bm{x}_i)-y_i^\prime)^2
\right]-\left[
(L^c_{\bm{\omega}^\star}(\bm{x}_i)-y_i^\prime)^2- (L^c_{\bm{\omega}_2}(\bm{x}_i)-y_i^\prime)^2
\right] \right\}^2\\
=&\left[(L^c_{\bm{\omega}_1}(\bm{x}_i)-y_i^\prime)^2
-(L^c_{\bm{\omega}_2}(\bm{x}_i)-y_i^\prime)^2
\right]^2  \leq 4\left(L^c_{\bm{\omega}_1}(\bm{x}_i)-L^c_{\bm{\omega}_2}(\bm{x}_i)
\right)^2  \\
\leq & 4
\Vert L^c_{\bm{\omega}_1}-L^c_{\bm{\omega}_2}\Vert_{L^{\infty}(\mathbb{P}_{\bm{X}})}^2,
\end{align*}
where the second last inequality follows from the fact that $\Vert L^c_{\bm{\omega} }\Vert_{L^{\infty}(\mathbb{P}_{\bm{X}})}\leq 1$. Therefore, we have
\begin{align*}
    \mathbb{E}(\bar{\rho}_{\mathcal{D}})
    \leq &
\sqrt{\mathbb{E}\left(\sup_{L^c_{\bm{\omega}_1},L^c_{\bm{\omega}_2} \in \mathcal{H}_i} \frac{1}{n}\sum_{i=1}^n \left(Q_{\bm{\omega}_1}(\bm{x}_i,y_i^\prime)-Q_{\bm{\omega_2}}(\bm{x}_i,y_i^\prime)\right)^2
    \right)
    }  \\
    \leq  &
        2\sqrt{\Vert L^c_{\bm{\omega}_1}-L^c_{\bm{\omega}_2}\Vert_{L^{\infty}(\mathbb{P}_{\bm{X}})}^2 
    } 
    \leq  2^{i+2} \sqrt{\delta}.
\end{align*}
By the Dudley's entropy integral, we have
\begin{align*}
 &  \mathbb{E}_{\mathcal{D}}\left(
\sup_{L^c_{\bm{\omega}} \in \mathcal{H}_i}
\sum_{i=1}^n V_i (\bm{\omega})
\right)
\leq  \mathbb{E}_{\mathcal{D}}\left( \sqrt{n}\int_{0}^{\bar{\rho}_{\mathcal{D}}}
\sqrt{\log C(\mathcal{H}_i,\rho_{\mathcal{D}},\epsilon)} d\epsilon
\right) \\
\leq &
\sqrt{n}\int_{0}^{\mathbb{E}_{\mathcal{D}}(\bar{\rho}_{\mathcal{D}})}
\sqrt{\log C(\mathcal{N}_{\delta},\Vert \cdot \Vert_{L^{\infty}(\mathbb{P}_{\bm{X}})},\epsilon)} d\epsilon
\leq 
\sqrt{n}\int_{0}^{2^{i+2} \sqrt{\delta}}
\sqrt{\log C(\mathcal{N}_{\delta},\Vert \cdot \Vert_{L^{\infty}(\mathbb{P}_{\bm{X}})},\epsilon)} d\epsilon,
\end{align*}
where $C(\mathcal{H}_i,\rho,\epsilon)$ denotes the $\epsilon$-covering number of $\mathcal{H}_i$ with respect to the metric $\rho$, and the second last inequality follows from the Jensen's inequality.

For any $C_2 \leq 1$ and $C_1 \geq 1$, it holds that
\begin{align*}
\int_{0}^{C_2 } \sqrt{\log(C_1/t)}dt =
C_1 \int_{C_1C_2^{-1} }^{\infty} \frac{\sqrt{\log(s)}}{s^2}ds 
\leq 
\frac{C_1}{\sqrt{\log(C_1C_2^{-1})}} \int_{C_1C_2^{-1} }^{\infty} \frac{\log(s)}{s^2}ds \lesssim
C_2 \sqrt{\log(C_1/C_2)},
\end{align*}
where the last inequality follows from that $\int_{a}^{\infty} \log(x)/x^2 dx = (\log(a)+1)/a$. By Lemma 5 of \citet{NNregres}, it holds that
\begin{align*}
  & \sqrt{n}\int_{0}^{2^{i+2} \sqrt{\delta}}
\sqrt{\log C(\mathcal{N}_{\delta},\Vert \cdot \Vert_{L^{\infty}(\mathbb{P}_{\bm{X}})},\epsilon)} d\epsilon \\
\lesssim  &\sqrt{n}\int_{0}^{2^{i+2} \sqrt{\delta}}(WL)\sqrt{\log\left(\frac{4(LW)^3}{\epsilon}\right)}d\epsilon
\lesssim 2^{i+2} \sqrt{n\delta}WL \log((WL)^3/(2^{i}\delta)).
\end{align*}

To sum up, we have
\begin{align*}
  \frac{1}{n}  \mathbb{E}\left(
\sup_{L^c_{\bm{\omega}} \in \mathcal{H}_i}
\frac{1}{n}
\sum_{i=1}^n V_i (\bm{\omega})
\right)
\lesssim 
 \frac{2^{i+2}\sqrt{\delta} WL}{\sqrt{n}} \log((WL)^3/(2^{i}\delta)).
\end{align*}
If we require $\frac{2^{i+2}\sqrt{\delta} WL}{\sqrt{n}} \log((WL)^3/(2^{i}\delta)) \leq 2^{i-3}\delta$, then we have $\sqrt{\delta} \gtrsim \frac{WL}{\sqrt{n}}\log(WLn)$. Note that
\begin{align*}
    \mathbb{E}
    \left(
V_i(\bm{\omega})    \right)^2 =&
    \mathbb{E}
    \left[
(L_{\bm{\omega}^\star}(\bm{x}_i)-y_i^\prime)^2- (L^c_{\bm{\omega}}(\bm{x}_i)-y_i^\prime)^2
\right]^2 \leq (C_0+1)^2
    \mathbb{E}
    \left[
(L_{\bm{\omega}^\star}(\bm{x}_i)-L^c_{\bm{\omega}}(\bm{x}_i)
\right]^2 \\
\leq &2(C_0+2)^2 \Vert L^c_{\bm{\omega}} - \eta \Vert_{L^2(\mathbb{P}_{\bm{X}})}^2+
2(C_0+2)^2\Vert L_{\bm{\omega}^\star} - \eta \Vert_{L^2(\mathbb{P}_{\bm{X}})}^2.
\end{align*}
Let $C_1 = (C_0+2)^2$. Therefore, we have
\begin{align*}
   \sup_{L_{\bm{\omega}}\in \mathcal{H}_i} \sum_{i=1}^n 
   \mathbb{E}
    \left(
V_i(\bm{\omega})    \right)^2 \leq 2^{i+1}C_1 n\delta  +2C_1 n\delta
\leq 2^{i+2}C_1n\delta.
\end{align*}
Combining this with (\ref{LCE}) yields that
\begin{align*}
\mathbb{P}\left(
\sup_{L^c_{\bm{\omega}} \in \mathcal{H}_i}
U_n(\bm{\omega}^\star) -  U_n(\bm{\omega}) \geq 0
\right)\leq 
    \mathbb{P}\left\{
\sup_{L^c_{\bm{\omega}} \in \mathcal{H}_i}
\frac{1}{n}
\sum_{i=1}^n V_i(\bm{\omega})
-
\frac{1}{n}
\mathbb{E}\left(
\sup_{L^c_{\bm{\omega}} \in \mathcal{H}_i}
\sum_{i=1}^n V_i (\bm{\omega})
\right)
\geq 2^{i-3}\delta
\right\}.
\end{align*}
Applying Theorem 1.1 of \citet{klein2005concentration}, we have
\begin{align*}
   \mathbb{P}\left(
\sup_{L^c_{\bm{\omega}} \in \mathcal{H}_i}
U_n(\bm{\omega}^\star) -  U_n(\bm{\omega}) \geq 0
\right)  \leq \exp\left(
-\frac{n^2 2^{2i-6}\delta^2/C_1}{n2^{i+4}\delta+n2^{i}\delta}
\right) \leq \exp(-C n i\delta),
\end{align*}
for some positive constant $C$. 
\begin{align*}
    \mathbb{P}\Big(
\Vert
\eta-\widetilde{\eta}\Vert_{L^2(\mathbb{P}_{\bm{X}})}^2 \geq \delta \Big) \leq \sum_{i=1}^\infty\exp(-C n i\delta)
=\frac{\exp(-Cn\delta)}{1-\exp(-Cn\delta)}
\leq 2\exp(-Cn\delta),
\end{align*}
where the last inequality holds when $n\delta \rightarrow \infty$. Next, we combine the following facts about $\delta$
\begin{align*}
    \delta  \gtrsim \frac{(WL)^2}{n}\log^2(WLn)
    \mbox{ and } (WL)^{-\frac{4\beta}{p}} \leq \delta/4.
\end{align*}
Choosing $WL \asymp n^{\frac{p}{4\beta+2p}}$ and $\delta \asymp n^{-\frac{2\beta}{2\beta+p}}\log^2(n)$, we have
\begin{align}
\label{BoundStep2}
\mathbb{E}_{\mathcal{D}}
\left(
\Vert
\eta-\widetilde{\eta}\Vert_{L^2(\mathbb{P}_{\bm{X}})}^2 
\right)
\lesssim n^{-\frac{2\beta}{2\beta+p}} \log^2(n).
\end{align}
This completes the proof of Step 2. Combining (\ref{BoundStep1}) and (\ref{BoundStep2}) yields that
\begin{align*}
    \mathbb{E}_{\mathcal{D}}\left\{\overline{U}(\mathbb{P}_{\widetilde{\bm{X}}, \widetilde{Y}})
    \right\}
\lesssim  &
D_{\mathcal{H}_1}(\mathbb{P}_{\widetilde{\bm{X}}}^\star,\mathbb{P}_{\bm{X}})+
n^{-\frac{2\beta}{2\beta+p}} \log^2(n)+
n^{-\frac{\alpha}{2\alpha+p}} \\
= &
\min_{g_{\bm{\theta}} \in \mathcal{G}}
    \mathrm{TV}(\mathbb{P}_{\bm{X}},\mathbb{P}_{g_{\bm{\theta}}(\bm{Z})}) +
    n^{-\frac{2\beta}{2\beta+p}} \log^2(n)+
n^{-\frac{\alpha}{2\alpha+p}}.
\end{align*}
This completes the proof. \qed \\

\noindent
\textbf{Proof of Theorem \ref{Thm:Consis}.} We first denote that $f_{\mathcal{F}_1}^\star = \argmin_{f \in \mathcal{F}_1}R(f)$ and $f_{\mathcal{F}_2}^\star = \argmin_{f \in \mathcal{F}_2}R(f)$. It can be verified that a sufficient condition for $R(\widetilde{f}_{\mathcal{F}_1}^\star)>R(\widetilde{f}_{\mathcal{F}_2}^\star)$ is $R(f_{\mathcal{F}_1}^\star) > R(\widetilde{f}_{\mathcal{F}_2}^\star)$, or equivalently $\Phi(f_{\mathcal{F}_1}^\star) > \Phi(\widetilde{f}_{\mathcal{F}_2}^\star)$. Recall that $f^\star(\bm{x})=\sign(\eta(\bm{x}) - 1/2)$ and $\widetilde{f}^\star(\bm{x})=\widetilde{f}^\star(\bm{x})$
\begin{align*}
&\Phi(\widetilde{f}_{\mathcal{F}_2}^\star) =
\int_{\mathcal{X}} 
I\left(
\widetilde{f}_{\mathcal{F}_2}^\star(\bm{x}) )
\neq 
f^\star(\bm{x})
\right)|2\eta(\bm{x})-1| P_{\bm{X}}(\bm{x}) d\bm{x} \\
\leq &
\int_{\mathcal{X}} 
I\left(
\widetilde{f}_{\mathcal{F}_2}^\star(\bm{x}) 
\neq 
\widetilde{f}^\star(\bm{x})
\right)|2\eta(\bm{x})-1|  P_{\bm{X}}(\bm{x}) d\bm{x} +
\int_{\mathcal{X}} 
I\left(
f^\star(\bm{x})
\neq 
\widetilde{f}^\star(\bm{x})
\right)|2\eta(\bm{x})-1| P_{\bm{X}}(\bm{x}) d\bm{x}\\
\leq &\underbrace{\int_{\mathcal{X}} 
I\left(
\widetilde{f}_{\mathcal{F}_2}^\star(\bm{x}) 
\neq 
\widetilde{f}^\star(\bm{x})
\right)|2\eta(\bm{x})-1|  P_{\bm{X}}(\bm{x}) d\bm{x}}_{A_1}+
2 \mathbb{E}
\left[I\left(\bm{X}\in \mathcal{A}^c(f^\star,\widetilde{f}^\star)\right) \cdot |\widetilde{\eta}(\bm{X})-\eta(\bm{X})|
\right]
,
\end{align*}
where the second inequality follows from the fact that $\widetilde{f}_{\mathcal{F}_2}^\star(\bm{x}) 
\neq 
\widetilde{f}^\star(\bm{x})$ implies $|2\eta(\bm{x})-1|\leq 2|\widetilde{\eta}(\bm{X})-\eta(\bm{X})|$, and the first inequality follows from the fact that 
\begin{align*}
I\left(
\widetilde{f}_{\mathcal{F}_2}^\star(\bm{x}) 
\neq 
f^\star(\bm{x})
\right) \leq I\left(
\widetilde{f}_{\mathcal{F}_2}^\star(\bm{x}) 
\neq 
\widetilde{f}^\star(\bm{x})
\right)+I\left(
f^\star(\bm{x})
\neq 
\widetilde{f}^\star(\bm{x})
\right).
\end{align*}
Next, $A_1$ can be further bounded as
\begin{align*}
    A_1 \leq &  \int_{\mathcal{X}} 
I\left(
\widetilde{f}_{\mathcal{F}_2}^\star(\bm{x}) 
\neq 
\widetilde{f}^\star(\bm{x})
\right)|2\widetilde{\eta}(\bm{x})-1|  P_{\bm{X}}(\bm{x}) d\bm{x}\\
&+
2\int_{\mathcal{X}} 
I\left(
\widetilde{f}_{\mathcal{F}_2}^\star(\bm{x}) 
\neq 
\widetilde{f}^\star(\bm{x})
\right)|\widetilde{\eta}(\bm{x})-\eta(\bm{x})|  P_{\bm{X}}(\bm{x}) d\bm{x} \\
= &
\underbrace{
\int_{\mathcal{X}} 
I\left(
\widetilde{f}_{\mathcal{F}_2}^\star(\bm{x}) 
\neq 
\widetilde{f}^\star(\bm{x})
\right)|2\widetilde{\eta}(\bm{x})-1|  P_{\bm{X}}(\bm{x}) d\bm{x}}_{A_2}+
2 \mathbb{E}
\left[I\left(\bm{X}\in \mathcal{A}^c(\widetilde{f}_{\mathcal{F}_2}^\star,\widetilde{f}^\star)\right) \cdot |\widetilde{\eta}(\bm{X})-\eta(\bm{X})|
\right].
\end{align*}

Next, we turn to bound $A_2$ via the decomposition of $\mathcal{X}$ into two parts.
\begin{align*}
\mathcal{X} = 
\left\{\bm{x} \in \mathcal{X}: \frac{P_{\bm{X}}(\bm{x})}{P_{\widetilde{\bm{X}}}(\bm{x})} \leq C
\right\} \cup \left\{\bm{x} \in \mathcal{X}: \frac{P_{\bm{X}}(\bm{x})}{P_{\widetilde{\bm{X}}}(\bm{x}) }\geq C
\right\}  \triangleq \mathcal{X}_1 \cup \mathcal{X}_2.
\end{align*}
Then, $A_2$ can be written as
\begin{align*}
A_2  = &
\int_{\mathcal{X}_1} 
I\left(
\widetilde{f}_{\mathcal{F}_2}^\star(\bm{x}) 
\neq 
\widetilde{f}^\star(\bm{x})
\right)|2\widetilde{\eta}(\bm{x})-1|  P_{\bm{X}}(\bm{x}) d\bm{x}+
\int_{\mathcal{X}_2} 
I\left(
\widetilde{f}_{\mathcal{F}_2}^\star(\bm{x}) 
\neq 
\widetilde{f}^\star(\bm{x})
\right)|2\widetilde{\eta}(\bm{x})-1|  P_{\bm{X}}(\bm{x}) d\bm{x}\\
\leq & C
\int_{\mathcal{X}_1} 
I\left(
\widetilde{f}_{\mathcal{F}_2}^\star(\bm{x}) 
\neq 
\widetilde{f}^\star(\bm{x})
\right)|2\widetilde{\eta}(\bm{x})-1|  P_{\widetilde{\bm{X}}}(\bm{x}) d\bm{x}+
V \cdot C^{-d} \leq 
C \widetilde{\Phi}(\widetilde{f}_{\mathcal{F}_2}^\star)+V \cdot C^{-d}.
\end{align*}
By the fact that $f(x)=ax+bx^{-d}$ attains its minimum at $x=(db/a)^{1/(d+1)}$, we set $C=(dV/\widetilde{\Phi}(\widetilde{f}_{\mathcal{F}_2}^\star))^{\frac{1}{d+1}}$, which yields that
\begin{align}
\label{UBG0}
A_2
\leq   ( d^{\frac{1}{d+1}}+d^{-\frac{d}{d+1}})V^{\frac{1}{d+1}}
\left(\widetilde{\Phi}(\widetilde{f}_{\mathcal{F}_2}^\star)\right)^{\frac{d}{d+1}}
\leq 
( d^{\frac{1}{d+1}}+d^{-\frac{d}{d+1}})
V^{\frac{1}{d+1}}
\left(\widetilde{\Phi}(f_{\mathcal{F}_2}^\star)\right)^{\frac{d}{d+1}},
\end{align}
where the second inequality follows from the fact that $\widetilde{\Phi}(\widetilde{f}_{\mathcal{F}_2}^\star) \leq \widetilde{\Phi}(f_{\mathcal{F}_2}^\star)$.

Next, we turn to establish relation between $\widetilde{\Phi}(f_{\mathcal{F}_2}^\star)$ and $\Phi(f_{\mathcal{F}_2}^\star)$. We first apply a similar treatment to $\mathcal{X}$ by the decomposition as
\begin{align*}
\mathcal{X} = 
\left\{\bm{x} \in \mathcal{X}: \frac{P_{\widetilde{\bm{X}}}(\bm{x})}{P_{\bm{X}}(\bm{x})} \leq C
\right\} \cup \left\{\bm{x} \in \mathcal{X}: \frac{P_{\widetilde{\bm{X}}}(\bm{x}) }{P_{\bm{X}}(\bm{x})}\geq C
\right\}  \triangleq \mathcal{X}_3 \cup \mathcal{X}_4,
\end{align*}
for some positive constants $C$. With this, it follows that
\begin{align*}
\widetilde{\Phi}(f_{\mathcal{F}_2}^\star) = &
\int_{\mathcal{X}} 
I\left(
f_{\mathcal{F}_2}^\star(\bm{x}) 
\neq 
\widetilde{f}^\star(\bm{x})
\right)|2\widetilde{\eta}(\bm{x})-1|  P_{\widetilde{\bm{X}}}(\bm{x}) d\bm{x} \\
\leq &
\int_{\mathcal{X}_3} 
I\left(
f_{\mathcal{F}_2}^\star(\bm{x}) 
\neq 
\widetilde{f}^\star(\bm{x})
\right))|2\widetilde{\eta}(\bm{x})-1|  P_{\widetilde{\bm{X}}}(\bm{x}) d\bm{x} \\
+&
\int_{\mathcal{X}_4} 
I\left(
f_{\mathcal{F}_2}^\star(\bm{x}) 
\neq 
\widetilde{f}^\star(\bm{x})
\right)|2\widetilde{\eta}(\bm{x})-1|  P_{\widetilde{\bm{X}}}(\bm{x}) d\bm{x} \\
\leq & C
\int_{\mathcal{X}} 
I\left(
f_{\mathcal{F}_2}^\star(\bm{x}) 
\neq 
\widetilde{f}^\star(\bm{x})
\right)|2\widetilde{\eta}(\bm{x})-1|  P_{\bm{X}}(\bm{x}) d\bm{x} + V  \cdot C^{-d}.
\end{align*}
Similarly, taking $C = \left(Vd/\int_{\mathcal{X}} 
I\left(
f_{\mathcal{F}_2}^\star(\bm{x}) 
\neq 
\widetilde{f}^\star(\bm{x})
\right)|2\widetilde{\eta}(\bm{x})-1|  P_{\bm{X}}(\bm{x}) d\bm{x}\right)^{\frac{1}{d+1}}$ yields that
\begin{align}
\label{Bound_G21}
\widetilde{\Phi}(f_{\mathcal{F}_2}^\star) \leq (d^{\frac{1}{d+1}}+d^{-\frac{d}{d+1}})
V^{\frac{1}{d+1}} 
\Big(
\underbrace{\int_{\mathcal{X}} 
I\left(
f_{\mathcal{F}_2}^\star(\bm{x}) 
\neq 
\widetilde{f}^\star(\bm{x})
\right)|2\widetilde{\eta}(\bm{x})-1|  P_{\bm{X}}(\bm{x}) d\bm{x}}_{A_3}
\Big)^{\frac{d}{d+1}}.
\end{align}
Next, we proceed to bound $A_3$.
\begin{align}
\label{Bound_G2}
A_3 = &\int_{\mathcal{X}} 
I\left(
f_{\mathcal{F}_2}^\star(\bm{x}) 
\neq 
\widetilde{f}^\star(\bm{x})
\right)|2\widetilde{\eta}(\bm{x})-1|  P_{\bm{X}}(\bm{x}) d\bm{x} \notag \\
\leq &
\int_{\mathcal{X}} 
I\left(
f_{\mathcal{F}_2}^\star(\bm{x}) 
\neq 
f^\star(\bm{x})
\right)|2\widetilde{\eta}(\bm{x})-1|  P_{\bm{X}}(\bm{x}) d\bm{x}
+
\int_{\mathcal{X}} 
I\left(
\widetilde{f}^\star(\bm{x}) 
\neq 
f^\star(\bm{x})
\right)|2\widetilde{\eta}(\bm{x})-1|  P_{\bm{X}}(\bm{x}) d\bm{x} \notag \\
\leq &2
\mathbb{E}_{\bm{X}\sim \mathbb{P}_{\bm{X}}}
\left[I\left(\bm{X}\in \mathcal{A}^c(f_{\mathcal{F}_2}^\star,f^\star)\right) |\widetilde{\eta}(\bm{X})-\eta(\bm{X})|
\right]+ \Phi(f_{\mathcal{F}_2}^\star) \notag \\
+ &
2
\mathbb{E}_{\bm{X}\sim \mathbb{P}_{\bm{X}}}
\left[I\left(\bm{X}\in \mathcal{A}^c(\widetilde{f}^\star,f^\star)\right) |\widetilde{\eta}(\bm{X})-\eta(\bm{X})|
\right].
\end{align}
For each $\mathcal{F}$, we define $\Lambda_\mathcal{F} = \mathcal{A}^c(\widetilde{f}^\star,f^\star)  \cup \mathcal{A}^c(f_{\mathcal{F}}^\star,f^\star) \cup \mathcal{A}^c(\widetilde{f}_{\mathcal{F}}^\star,\widetilde{f}^\star)$ and $\Upsilon_{\mathcal{F}}$ as
\begin{align*}
    \Upsilon_{\mathcal{F}} = 
    \mathbb{E}_{\bm{X}\sim \mathbb{P}_{\bm{X}}}
\left[I\left(\bm{X}\in \Upsilon_{\mathcal{F}} \right) |\widetilde{\eta}(\bm{X})-\eta(\bm{X})|
\right].
\end{align*}
Combining (\ref{Bound_G21}) and (\ref{Bound_G2}), we get
\begin{align}
\label{U3_Bound3}
\widetilde{\Phi}(f_{\mathcal{F}_2}^\star) 
\leq (d^{\frac{1}{d+1}}+d^{-\frac{d}{d+1}})
V^{\frac{1}{d+1}} 
\left(
\Phi(f_{\mathcal{F}_2}^\star)+
4\Upsilon_{\mathcal{F}_2}
\right)^{\frac{d}{d+1}}.
\end{align}
Plugging (\ref{U3_Bound3}) into (\ref{UBG0})
\begin{align*}
A_2
\leq 
( d^{\frac{1}{d+1}}+d^{-\frac{d}{d+1}})^{\frac{2d+1}{d+1}} 
V^{\frac{2d+1}{(d+1)^2}} 
\left(
\Phi(f_{\mathcal{F}_2}^\star)+
4\Upsilon_{\mathcal{F}_2}
\right)^{\frac{d^2}{(d+1)^2}}.
\end{align*}
To sum up, we get
\begin{align*}
\Phi(\widetilde{f}_{\mathcal{F}_2}^\star) \leq &
( d^{\frac{1}{d+1}}+d^{-\frac{d}{d+1}})^{\frac{2d+1}{d+1}} 
V^{\frac{2d+1}{(d+1)^2}} 
\left(
\Phi(f_{\mathcal{F}_2}^\star)+
4\Upsilon_{\mathcal{F}_2}
\right)^{\frac{d^2}{(d+1)^2}}+
4\Upsilon_{\mathcal{F}_2} \\
\leq &
K_{d,V}\left(
\Phi(f_{\mathcal{F}_2}^\star)
\right)^{\frac{d^2}{(d+1)^2}}+K_{d,V}
\left(4\Upsilon_{\mathcal{F}_2}\right)^{\frac{d^2}{(d+1)^2}}+
4\Upsilon_{\mathcal{F}_2},
\end{align*}
where $K_{d,V} = ( d^{\frac{1}{d+1}}+d^{-\frac{d}{d+1}})^{\frac{2d+1}{d+1}} 
V^{\frac{2d+1}{(d+1)^2}}$. Clearly, we have $\Phi(\widetilde{f}_{\mathcal{F}_2}^\star) < \Phi(\widetilde{f}_{\mathcal{F}_1}^\star)$ if the following inequality holds true.
\begin{align*}
K_{d,V}\left(
\Phi(f_{\mathcal{F}_2}^\star)
\right)^{\frac{d^2}{(d+1)^2}}+K_{d,V}
\left(4\Upsilon_{\mathcal{F}_2}\right)^{\frac{d^2}{(d+1)^2}}+
4\Upsilon_{\mathcal{F}_2} < \Phi(f_{\mathcal{F}_1}^\star).
\end{align*}
This completes the proof of Theorem \ref{Thm:Consis}. \qed \\

\noindent
\textbf{Proof of Theorem \ref{Thm:Consis_2}.}
We first note that $\widetilde{\eta}(\bm{x}),\eta(\bm{x}) \in [0,1]$. Therefore, $\Vert \widetilde{\eta} - \eta \Vert_{L^2(\mathbb{P}_{\bm{X}})} \leq \Vert \widetilde{\eta} - \eta \Vert_{L^2(\mathbb{P}_{\bm{X}})}^{\frac{d^2}{(d+1)^2}}$. This indicates a sufficient condition to achieve condition in Theorem \ref{Thm:Consis} is
\begin{align*}
    4(K_{d,V}+1)
\Vert \widetilde{\eta} - \eta \Vert_{L^2(\mathbb{P}_{\bm{X}})}^{\frac{d^2}{(d+1)^2}}< \Phi(f_{\mathcal{F}_1}^\star)-
K_{d,V}\left(
\Phi(f_{\mathcal{F}_2}^\star)
\right)^{\frac{d^2}{(d+1)^2}}.
\end{align*}
Then for any $(\mathcal{F}_1,\mathcal{F}_2) \in \mathcal{H}(v)$, we have
\begin{align*}
&\mathbb{P}
\left(
\big(R(f_{\mathcal{F}_2}^\star)-R(f_{\mathcal{F}_1}^\star)\big)\big(R(\widetilde{f}_{\mathcal{F}_2}^\star)-R(\widetilde{f}_{\mathcal{F}_1}^\star)\big)>0
\right) \\
\geq & 
\mathbb{P}
\left(
4(K_{d,V}+1)
\Vert \widetilde{\eta} - \eta \Vert_{L^2(\mathbb{P}_{\bm{X}})}^{\frac{d^2}{(d+1)^2}}
\leq v
\right) \geq 1 - C_1\exp\left(-C_2 a_n v^{\frac{(d+1)^2}{d^2}}(4K_{d,V}+4)^{-\frac{(d+1)^2}{d^2}}\right).
\end{align*}
This completes the proof. \qed \\

\section{Proof of Corollaries}
\label{Sec:POC}
In this Appendix, we provide proofs for Corollaries \ref{Cro:Down}-\ref{Coro:Consis}. \\

\noindent 
\textbf{Proof of Corollary \ref{Cro:Down}.} First, Theorem \ref{Thm:UTBmore} shows that
\begin{align}
\label{BounFinal}
U(\tf,\f) 
\leq &2D_{\mathcal{K}}(\mathbb{P}_{\bm{X}},\mathbb{P}_{\widetilde{\bm{X}}})
+ 4 \Lambda(\mathcal{F}) \Vert \widetilde{\eta}-\eta\Vert_{L^2(\mathbb{P}_{X})}+
4 \Upsilon(\widetilde{\eta}).
\end{align}
Given that $f^\star = \widetilde{f}^\star$, (\ref{BounFinal}) becomes
\begin{align*}
    U(\tf,\f) 
\leq &2D_{\mathcal{K}}(\mathbb{P}_{\bm{X}},\mathbb{P}_{\widetilde{\bm{X}}})
+ 4 \Lambda(\mathcal{F}) \Vert \widetilde{\eta}-\eta\Vert_{L^2(\mathbb{P}_{X})}.
\end{align*}
Furthermore, $f^\star = \widetilde{f}^\star$ and $f^\star \in \mathcal{F}$ implies that
\begin{align*}
&D_{\mathcal{K}}(\mathbb{P}_{\bm{X}},\mathbb{P}_{\widetilde{\bm{X}}})\\
= &
\left|
\mathbb{E}_{\bm{X}}\Big[
|2\widetilde{\eta}(\bm{X})-1|I\big(f_{\mathcal{F}}^\star(\bm{X}) \neq f^\star(\bm{X})\big)
\Big]-
\mathbb{E}_{\widetilde{\bm{X}}}\Big[
|2\widetilde{\eta}(\widetilde{\bm{X}})-1|I\big(\widetilde{f}^\star_{\mathcal{F}}(\widetilde{\bm{X}}) \neq \widetilde{f}^\star(\widetilde{\bm{X}})\big)
\Big]
\right|=0.
\end{align*}
This completes the proof. \qed \\

\noindent 
\textbf{Proof of Corollary \ref{Coro:Consis}.} The proof of Corollary \ref{Coro:Consis} is similar to that of Theorem \ref{Thm:Consis_2}. Note that when $\Phi(f_{\mathcal{F}_2}^\star)=0$, $\mathcal{W}(\mathcal{F}_1,\mathcal{F}_2) = \Phi(f_{\mathcal{F}_1}^\star)$. Then applying similar steps as follows:
\begin{align*}
&\mathbb{P}
\left(
\big(R(f_{\mathcal{F}_2}^\star)-R(f_{\mathcal{F}_1}^\star)\big)\big(R(\widetilde{f}_{\mathcal{F}_2}^\star)-R(\widetilde{f}_{\mathcal{F}_1}^\star)\big)>0
\right) \\
\geq & 
\mathbb{P}
\left(
4(K_{d,V}+1)
\Vert \widetilde{\eta} - \eta \Vert_{L^2(\mathbb{P}_{\bm{X}})}^{\frac{d^2}{(d+1)^2}}
\leq \Phi(f_{\mathcal{F}_1}^\star)
\right) \\
\geq & 1 - C_1\exp\left(-C_2 a_n \left(\Phi(f_{\mathcal{F}_1}^\star)\right)^{\frac{(d+1)^2}{d^2}}(4K_{d,V}+4)^{-\frac{(d+1)^2}{d^2}}\right).
\end{align*}
This completes the proof. \qed \\
\renewcommand\refname{References}
\bibliography{Utility_Framework_Revision}

\end{document}